\documentclass[runningheads]{llncs}

\usepackage{accv}

\usepackage{accvabbrv}

\usepackage{graphicx}
\usepackage{booktabs}
\usepackage{wrapfig}
\usepackage{xcolor}
\usepackage{multicol}
\usepackage{multirow}
\usepackage{pdflscape}
\usepackage{float}
\usepackage{xfrac}

\usepackage[accsupp]{axessibility}  %

\usepackage{hyperref}

\usepackage{orcidlink}

\begin{document}

\newcommand{\np}{Naïve PAINE}
\newcommand{\scaleboxratio}{0.8}
\newcommand{\blue}[1]{\textcolor{blue}{#1}}
\newcommand{\red}[1]{\textcolor{red}{#1}}

\title{\np:\\Lightweight Text-to-Image Generation Improvement with %
Prompt Evaluation} 

\titlerunning{\np}

\author{Joong Ho Kim\inst{1} \and
Nicholas Thai\inst{1} \and
Souhardya Saha Dip\inst{1} \and \\
Dong Lao\inst{2} \and
Keith G. Mills\inst{1}}

\authorrunning{J. H. Kim et al.}

\institute{LSU ATHENA Lab, Baton Rouge, LA, U.S.A. \and
LSU Vision Lab, Baton Rouge, LA, U.S.A.}

\maketitle

\begin{abstract}
  Text-to-Image (T2I) generation is primarily driven by Diffusion Models (DM) which %
rely on random Gaussian noise. Thus, like playing the slots at a casino, a DM will produce different results given the same user-defined inputs. 
This imposes a gambler's burden: To perform multiple generation cycles to obtain a satisfactory result. However, even though DMs use stochastic sampling to seed generation, the distribution of generated content quality highly depends on the %
prompt and the generative ability %
of a DM with respect to it.

To account for this, 
we propose \np\ %
for improving the generative quality of Diffusion Models by leveraging %
T2I preference benchmarks. We %
directly 
predict the numerical quality of an image from the initial noise %
and %
given prompt. %
\np\ %
then selects a handful of quality noises and %
forwards them to the DM for generation. Further, \np\ provides feedback on the DM generative quality given the prompt %
and is lightweight enough to seamlessly fit into existing DM pipelines. %
Experimental results demonstrate that \np\ outperforms existing %
approaches on several prompt corpus benchmarks.

\end{abstract}

\section{Introduction}
\label{sec:intro}

Using computers to realize %
artwork from human thoughts has been a long-standing goal of Artificial Intelligence (AI)~\cite{cohen2016harold, nake1971nocart, nees1969generative}. Once delegated to science fiction, this task is %
being realized by AI generated content (AIGC) models~\cite{tang2024hart, goodfellow2014generative, ramesh2021zero}, among which %
latent Diffusion Models (DM)~\cite{Rombach_2022_CVPR, podell2023sdxl, esser2024scaling, chen2024pixartAlpha, chen2024pixartsigma, bfl2024Flux, xie2024sana, li2024hunyuandit, cai2025z} %
are arguably the most influential class  %
in terms of %
open-source impact~\cite{Civitai, civitaiJuggernautRagnarok_by_RunDiffusion, civitaiIllustriousV20}. %

However, interacting with a diffusion model can feel surprisingly similar to playing the slots in a casino. The user makes some initial choices, %
such as providing a textual prompt or picking which machine to sit in front of. Then, once the lever is pulled, 
the sampler begins its stochastic process~\cite{song2020denoising, song2023consistency, ho2020denoising}, and one counts the timesteps. %
If the result is unsatisfactory, the instinctual choice is to pull the lever again~\cite{ferrari2022behavioral, delfabbro2023complex}, yet %
each attempt consumes resources: casino tokens in Las Vegas, computational energy on a GPU, and in both cases, time. %
The governing principle is the same: every outcome begins from randomness, and every additional trial carries a cost.

This cost is incurred because DMs, like slot machines, have %
inherent %
stochasticity %
that imposes a non-negligible impact on generative performance~\cite{venkatraman2025outsourced, kalaivanan2025ess, wang2025source, om2025posterior} %
in terms of qualitative image perception~\cite{xu2025good} and quantitative metrics~\cite{ghosh2023geneval}. However, just as gamblers, arguably some of the most persistent data engineers, have historically attempted to reverse engineer random number generators to make slot machines predictable~\cite{schneier2017slots, koerner2017wired}, 
there are recent developments %
which can improve the output quality of DMs via initial noise optimization. %
Such approaches may mutate the initial noise~\cite{wang2025silent, zhou2025golden, tang2024inference} or adopt %
prompt manipulation~\cite{jiang2024frap, jiang2025pixelman, jiang2026raise} and image editing~\cite{zhang2023adding} ideas to alter %
generation%
~\cite{guo2024initno, bai2024zigzag, chefer2023attend, chen2024find}. %

However, a key limitation %
of current literature is that DM %
generative performance %
strongly depends on the %
prompt, as we show in Section~\ref{sec:prompt_influence}. Specifically, while there are several prompt-based T2I benchmarks~\cite{wu2023human, ma2025hpsv3, kirstain2023pick, ghosh2023geneval} which aim to quantify a human's preference for a given image with respect to the prompt, %
these are finite compared to the scope of human imagination. Moreover, 
modeling human preference is an inherently noisy task%
~\cite{mance2022noise, jaramillo2022radiologists,  xu2023imagereward}, %
benchmark performance is not all you need. 
Rather, before a gambler reverse engineers %
a casino slot machine, they should first determine \textit{which} slot machine yields the best jackpot.

We cast this issue as a research question: \emph{Given an initial noise, can one estimate how good the output is likely to be without executing the entire denoising process?} We answer this question with \np, 
or Naïve Prompt-Aware Initial Noise Evaluator, a lightweight, %
plug-and-play method that achieves initial noise optimization whilst also providing %
valuable and interpretable feedback on how well a DM can %
generate desired content. Specifically, \np\ short-circuits %
DM generation %
to estimate the performance of a would-be AI image from the initial stochastic noise that would seed generation. %
Further, we design our method to provide interpretable feedback on how hard a given prompt would be to generate. Our detailed contributions are as follows:

First, we cast the problem of initial noise optimization as a scalar prediction regression task. %
We construct PAINE as a predictor model that estimates the performance of DM-generated images from the prompts and initial noise tensors that are inputs to the process. Unlike more intensive~\cite{li2025noisear, eyring2024reno, eyring2025noise} or guided~\cite{guo2024initno, tang2024inference, chefer2023attend} approaches, PAINE is DM-agnostic, adapts to the usage of multiple prompt text encoders, and does not alter DM generation or require fine-tuning.

Second, we conduct a study on the influence that the prompt, noise and DM itself have on benchmark metric distributions in Section~\ref{sec:prompt_influence}. We find that the prompt %
significantly determines %
the distribution of scores, i.e., some prompts are more difficult to generate than others. Further, although sampling a noise serves to sample from the preference metric distribution, the most/least optimal noise varies with each prompt. Different from existing literature~\cite{tang2024inference, zhou2025golden, wang2025silent} which may mutate the noise to maximize performance for the prompt, we take inspiration from Bayes' Theorem~\cite{provost2013data, bishop2006pattern} and design our method to also provide a \textit{prior} measure of how well a DM can generate an image for a given %
prompt regardless of what noise is given, giving rise to \np.

Third, \np\ is both computationally lightweight and %
imposes a smaller inference latency overhead than existing methods~\cite{zhou2025golden} on various hardware. Further, it is easy to deploy  %
in various DM pipelines and workflows, whether as directly part of a codebase such as Diffusers~\cite{von-platen-etal-2022-diffusers} or easily integrated into a more complex GUI-based framework like ComfyUI~\cite{comfyui_contributors2025, xu2025comfyui} and others~\cite{autoamtic1111-stable-diffusion-webui}. 

Through extensive experiments we demonstrate that \np\ achieves 
competitive 
initial noise optimization in terms of quantitative benchmarks and qualitative generation results, 
whilst also providing accurate and interpretable prompt-performance feedback. To prove the generalizability of our method we test on several popular T2I DMs including SDXL~\cite{podell2023sdxl}, DreamShaper~\cite{lykon2023dreamshaper}, Hunyuan~\cite{li2024hunyuandit} and PixArt-$\Sigma$~\cite{chen2024pixartsigma}, outperforming other contemporary lightweight approaches~\cite{zhou2025golden} in terms of prompt-adherence and %
preference metrics~\cite{wu2023human, ma2025hpsv3}. %

\section{Background \& Related Work}
\label{sec:related}

\noindent\textbf{Diffusion Models (DM)} generate content by initially sampling a Gaussian noise $X_T\in\mathbb{R}^{H \times W \times C} \sim \mathcal{N}(0, 1)$ which is gradually refined using the reverse diffusion process (RDP)~\cite{song2020denoising}. Most Text-to-Image (T2I) DMs~\cite{podell2023sdxl, chen2024pixartsigma, li2024hunyuandit, esser2024scaling, xie2024sana} operate in a latent embedding space~\cite{Rombach_2022_CVPR} of a VAE~\cite{kingma2019introduction} encoder/decoder pair where $H$, $W$ and $C$ correspond to VAE height, width and channel embedding dimensions, respectively. The RDP is iterative, where at every timestep $t$, a scheduler computes $X_{t-1}$ from $X_t$, i.e., in the case of DDIM~\cite{song2020denoising},

\begin{equation}
    \centering
    \label{eq:rdp}
    X_{t-1} = \alpha_{t-1} (\dfrac{X_t - \sigma_t\epsilon_\theta (X_t, t, c)}{\alpha_t}) + \sigma_{t-1}\epsilon_{\theta}(X_t, t, c), 
\end{equation}
where $\epsilon_{\theta}$ is the \textit{denoiser} neural network such as U-Net~\cite{Rombach_2022_CVPR, podell2023sdxl} or Diffusion Transformer (DiT)~\cite{peebles2023scalable, chen2024pixartsigma, li2024hunyuandit} and $c$ is user-defined conditioning. In the case of T2I generation 
$c \in \mathbb{R}^{tok\times d_{tok}}$ %
is the embedding of a textual prompt, e.g., output of a CLIP~\cite{radford2021learning, hessel2021clipscore} or LLM such as  T5~\cite{raffel2020exploring} where $tok$ and $d_{tok}$ refer to the number of tokens and token embedding dimension, respectively. The objective of $\epsilon_\theta$ is to facilitate adequate reverse diffusion of $X_t$ with respect to the user-provided conditioning. For example, if $c_1$ embeds `A green dog' and $c_2$ embeds `A red cat', the resultant $X_{0, c_1}$ and $X_{0,c_2}$ should yield substantially different images. 

However, $X_T$ is randomly sampled, so different initializations %
will yield qualitatively and quantitatively different results, as prior literature shows~\cite{xu2025good, chen2024find, tang2024inference, venkatraman2025outsourced}. %
This issue is compounded as $\epsilon_{\theta}$ is a large deep neural network %
that performs iterative inference. Ultimately, this means that while DMs have revolutionized T2I generation, many images will %
be subpar, leading to wasteful computation.

\subsection{Noise-Aware Optimization}
\label{sec:dm_noise_optim_rw}

Noise-aware optimization is a class of methods which take into account the random stochasticity of $X_T$ to improve DM generation quality in a variety of tasks such as protein synthesis~\cite{kalaivanan2025ess, smith2025calibrating} among others~\cite{om2025posterior, wang2025source, venkatraman2025outsourced}. In the specific context of visual content generation, i.e., T2I DMs, existing literature can generally be subdivided depending on whether fine-tuning of the denoiser DNN $\epsilon_{\theta}$ is involved - analogous to the distinction between Post-Training Quantization (PTQ)~\cite{li2023q, he2024ptqd, li2021brecq} and Quantization-Aware Training (QAT)~\cite{sui2024bitsfusion, ma2024era} methods for DNN bit-precision compression. \np\ is more like PTQ. 

Post-training methodologies apply %
mutation either to the noise or within the denoiser $\epsilon_\theta$ itself. For example, InitNO~\cite{guo2024initno} utilizes attention map edits~\cite{chefer2023attend} during DM inference to alter the generation trajectory and facilitate prompt-adherence~\cite{jiang2024pixelman}. DNO~\cite{tang2024inference} cast $X_T$ as a learnable variable and use reinforcement learning (RL) to train a policy for optimizing it. %
NoiseQuery~\cite{wang2025silent} collects a finite dataset of high-quality noise samples in an offline fashion and then matches the user prompt to the best noise at inference-time rather than sampling $X_T$. Finally, Golden Noise~\cite{zhou2025golden} train a Noise Prompt Network (NPNet) that mutates the initial noise $X_T$ into an optimal `golden noise' $X_T'$ based on the prompt. Although these methods do not require costly fine-tuning of $\epsilon_\theta$, they implement a mapping where a single prompt is mapped to a single optimal noise. However, as we show in Section~\ref{sec:prompt_influence}, %
the generative performance of a given DM is a statistical distribution with moments conditionally dependent on the user prompt. Thus, in contrast to the existing literature, our proposed \np\ utilizes %
sampling 
from the distribution of generative quality that is conditioned on the prompt.

Finally, in addition to post-training methodologies, there are a number of training~\cite{li2024immiscible} or fine-tuning optimization approaches. ReNO~\cite{eyring2024reno} and NoiseAR~\cite{li2025noisear} utilize reinforcement learning fine-tuning to optimize the initial noise while HyperNoise~\cite{eyring2025noise} combines step-distillation~\cite{sauer2023adversarial} with LLM test-time scaling~\cite{muennighoff2025s1}.  %
These approaches %
require GPU-intensive fine-tuning the denoiser $\epsilon_{\theta}$ and can be restricted in application, i.e., both ReNO and HyperNoise are applicable to step-distilled DMs like SD-Turbo~\cite{sauer2023adversarial} or SANA-Sprint~\cite{chen2025sana}. In contrast, our proposed \np\ is a model agnostic with plug-and-play integration into existing pipelines for %
hardware ranging from industrial GPUs to Nvidia DGX Spark units. %

\subsection{Human Preference Metrics}
\label{sec:human_pref}
The past several years have seen the development of a number of benchmarks and tools designed to capture the noisy and subjective human judgment~\cite{mance2022noise, jaramillo2022radiologists,  xu2023imagereward} of AI image content with respect to the textual prompt utilized to condition %
generation. These benchmarks typically utilize some kind of predictor to map a $(prompt, image)$ pair to a scalar score value.

The progenitor model for most of these approaches is CLIPScore~\cite{hessel2021clipscore}, a common DM evaluation metric where prompt-image pairs are scored by how well %
the image adheres to the prompt. %
More recent metrics such as HPSv2~\cite{wu2023human}, HPSv3~\cite{ma2025hpsv3}, ImageReward~\cite{xu2023imagereward} and PickScore~\cite{kirstain2023pick} %
train a predictor (or fine-tune from CLIPScore) to explicitly model the preference a human judge would express for a given image with respect to the prompt. %

Let $p$ be a textual prompt, i.e., `A red cat', and $I$ be an image. Human preference metric predictors %
first collect a carefully curated dataset of images and the human preference rankings for those images. We refer interested readers to specific %
publications for the exact details behind each metric. Second, these methods train a predictor model $\phi$ that learns to map prompt-image pairs $(p, I)$ to a \textit{scalar} score $S_{p, I} \in \mathbb{R}$, the %
numerical representation of the human preference for the image given the prompt. %
We formally represent this task as follows: %

\begin{equation}
    \centering
    \label{eq:scoring}
    S_{p, I} = \phi(p, I). %
\end{equation}
Typically, higher $S_{p, I}$ is better. In an ideal world, we could use a single prompt $p$ to generate several images $\mathcal{I}_p=\{I_1, I_2,...,I_N\}$, measure their preference scores $\{S_{p, I_1}, S_{p, I_2},...S_{p, I_N}\}$ and determine the most preferential image as $I_{best}$ by taking the \textit{argmax} over each image. However, note that $\phi$ is a predictor that provides accurate, yet imperfect estimations, so $I_{best}$ is not guaranteed to truly be the most preferred image. Rather, a more realistic assumption is that scores lie on some distribution $\mathcal{N}(\mu_{S_p}, \sigma_{S_p})$ and that $I_{best}$ resides in its upper tail. %

\begin{figure}[t!]
    \centering
    \includegraphics[width=\linewidth]{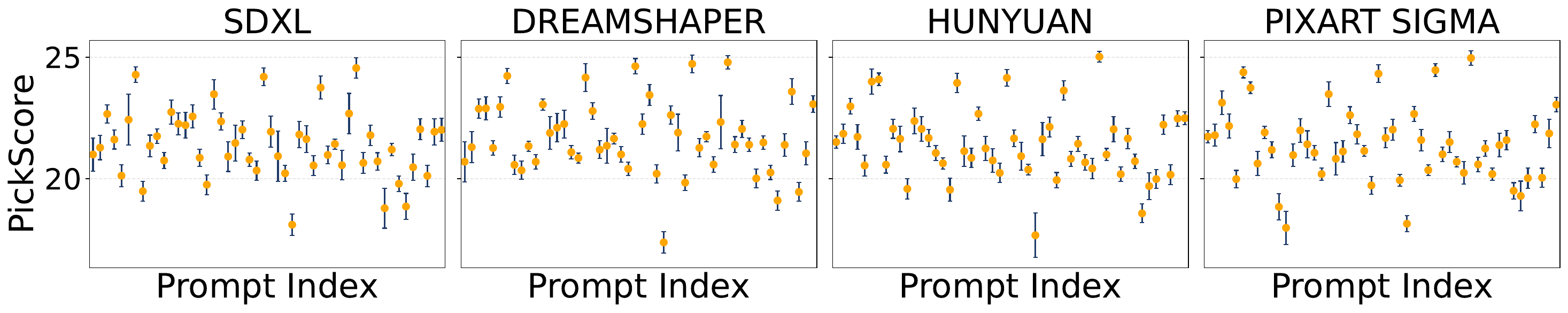} 
    \caption{Measuring the PickScore~\cite{kirstain2023pick} distribution across several prompts and DMs~\cite{podell2023sdxl, lykon2023dreamshaper, li2024hunyuandit, chen2024pixartsigma}. Specifically, we select the first 50 training prompts from Golden Noise~\cite{zhou2025golden} and generate 20 images per prompt, resetting the random seed prior to the first generation. Then, we plot the PickScore score mean and standard deviation (error bar).} %
    \label{fig:score_dists}
    \vspace{-6mm}
\end{figure}

\begin{wrapfigure}{rt}{0.45\textwidth}
    \centering
    \includegraphics[width=2in]{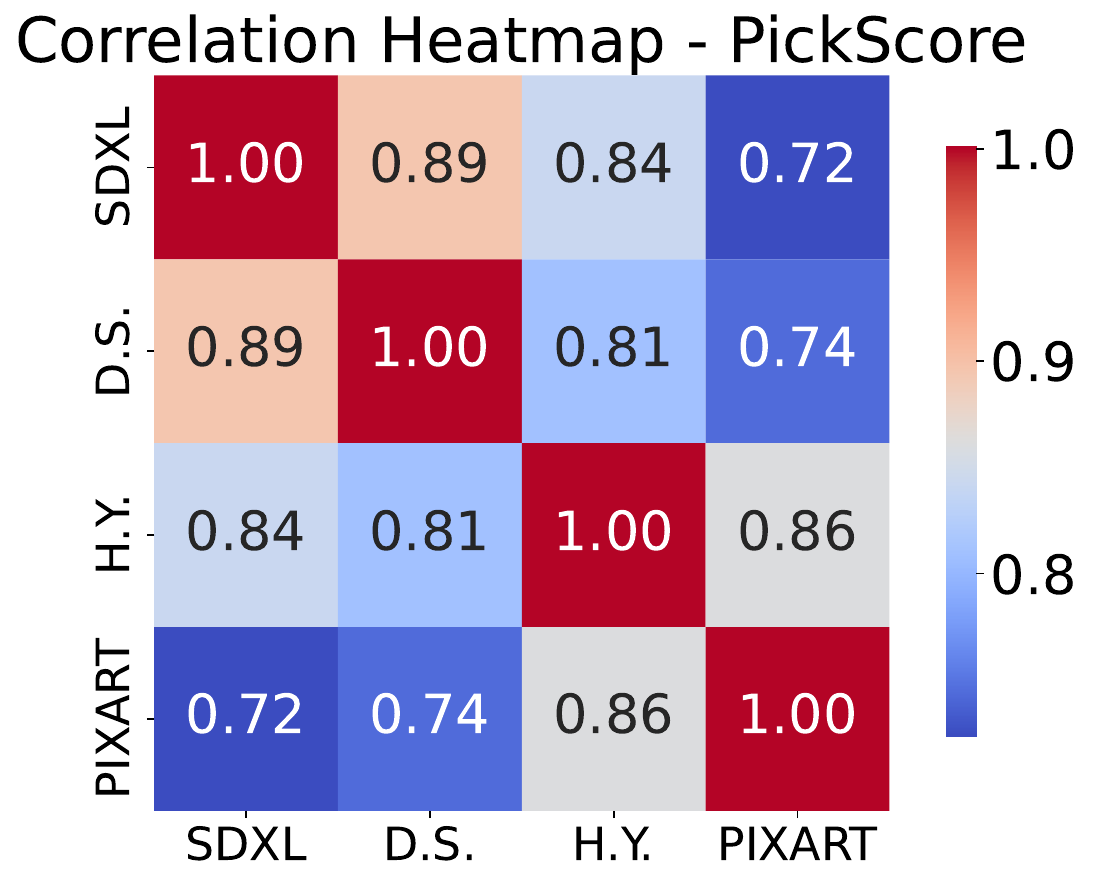} 
    \caption{Pearson Correlation Coefficient (PCC) matrix comparing the mean PickScore for each DM. `D.S.' and `H.Y.' shorthand for \cite{lykon2023dreamshaper} and \cite{li2024hunyuandit}.} %
    \label{fig:pcc_dm}
    \vspace{-6mm}
\end{wrapfigure}

\subsection{Human Preferences and Prompt Influence}
\label{sec:prompt_influence}

We conduct a simple experiment to determine what factors influence the moments, $\mu_{S_p}$ and $\sigma_{S_p}$, of this distribution. %
Specifically, we %
sample 50 prompts from the Pickapic training set~\cite{kirstain2023pick} and consider 
a number of different DMs~\cite{chen2024pixartAlpha, chen2024pixartsigma, li2024hunyuandit, bfl2024Flux, xie2024sana}. For each prompt on each DM, we fix the random seed and then generate 20 images. %
We then estimate the preference score, and %
compute the moments of the score population. For each DM, we then plot the mean and standard deviation of the scores based on prompt index.

\begin{wrapfigure}{rt}{0.45\textwidth}
    \centering
    \includegraphics[width=2in]{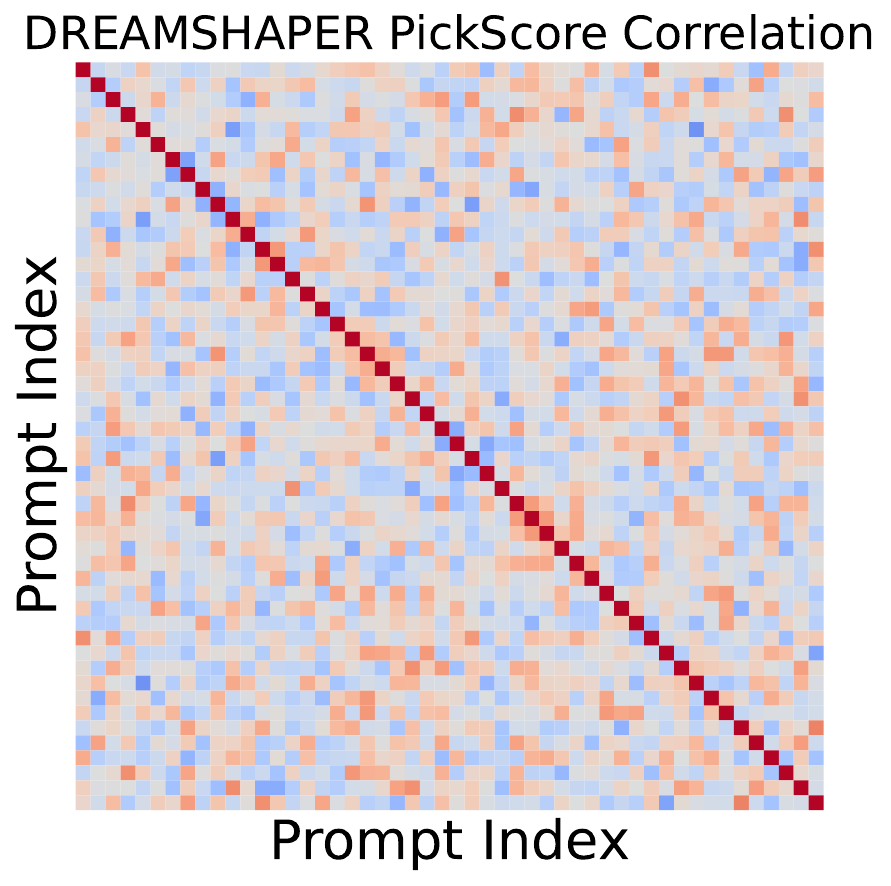} 
    \caption{Pearson Correlation Coefficient (PCC) matrix comparing the mean PickScore across prompts for \cite{lykon2023dreamshaper}.} 
    \label{fig:pcc_prompt}%
    \vspace{-7mm}
\end{wrapfigure}
Figure~\ref{fig:score_dists} illustrates our findings. The mean $\mu_{S_p}$ %
of the score distribution highly depends on the prompt $p$ and %
varies greatly with the prompt index. It is also evident that the standard deviation $\sigma_{S_p}$ depends on the prompt and noise as the error bars are not consistent across prompts, despite fixing the random seed.

Further, we compare the score distributions across DMs and note that while there are some similarities, i.e., some prompts generally obtain higher or lower preference score, the pattern is noisy. In fact, Figure~\ref{fig:pcc_dm} quantifies the degree to which different score means $\mu_{S_p}$ are linearly correlated. This result corroborates~\cite{zhou2025golden, chen2024find} and shows that different DMs will excel or produce substandard generation for different subsets of prompts.

Next, we measure to what extent a random initial noise $X_T$ that produces a high score on one prompt $p_i$ will also produce a high score on a different prompt $p_j$. To this end we compute the Pearson correlation coefficient between two series of scores $\{S_{p_i, I_1},S_{p_i, I_2},...S_{p_i, I_N}\}$ and $\{S_{p_j, I_1},S_{p_j, I_2},...S_{p_j, I_N}\}$. Figure~\ref{fig:pcc_prompt} illustrates our findings, which show that there is very little correlation between choice of noise and preference score across prompts for a given DM.

Overall, the primary insights we draw from these initial background experiments are that the choice of prompt greatly influences the range and distribution of achievable human preference scores more so than the actual DM utilized. Further, the choice of initial noise $X_T$ controls where on the distribution a given image will sit, but the choice of optimal noise must be done on a per-prompt basis. Next, we incorporate these insights into the design of our predictor.

\section{Methodology}
\label{sec:method}

\begin{figure}[t!]
    \centering
    \includegraphics[width=\linewidth]{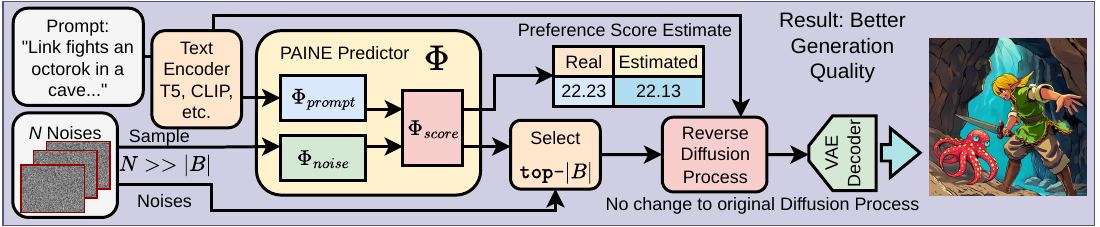}
    \caption{We augment a DM pipeline %
    with \np\ %
    which receives the prompt embedding $c$ and $N$ %
    noises as input to estimate the human preference score metric as if each noise were processed into an image. We sort and rank the scores of the $N$ noises and %
    forward the top-$|B|$ noises %
    to the DM for normal generation. Further, score estimations %
    provide feedback to the user on how well the DM can perform on a given prompt. Image from Hunyuan. Scores from our dataset and method on PickScore~\cite{kirstain2023pick}.}
    \label{fig:diagram}
    \vspace{-6mm}
\end{figure}

In this section we introduce \np\ in detail. Specifically, we describe the %
PAINE architecture, %
and 
how we practically deploy it in a plug-and-play fashion. Finally, we describe how \np\ %
provides feedback on prompt generation difficulty. Figure~\ref{fig:diagram} provides a high-level overview of our method.

\subsection{Prompt Evaluation Prior to Generation}

The PAINE predictor %
learns to estimate the scalar human preference of a prompt-conditioned image, $S_{p, I}$, from the initial, randomly sampled noise $X_T$ utilized to seed the DM RDP \textit{prior} to committing to costly image generation. Specifically, in the context of DM T2I generation, we first extend Equation~\ref{eq:scoring} to explicitly incorporate the generation process mentioned in Section~\ref{sec:related} as follows:

\begin{equation}
    \centering
    \label{eq:scoring_dm}
    S_{p, I} = \phi(p, \texttt{VAE}_{D}(\texttt{RDP}_{\epsilon_\theta}(X_T, \texttt{ENCODE}(p)))),
\end{equation}
where $\texttt{VAE}_D$ and $\texttt{RDP}_{\epsilon_\theta}$ represent the VAE decoder~\cite{kingma2019introduction, Rombach_2022_CVPR} utilized to translate denoised latents into image space and the iterative reverse diffusion process from Eq.~\ref{eq:rdp}, respectively, while $\texttt{ENCODE}(p)$ is the DM-specific way of encoding the textual prompt $p$ into its embedding $c$; $c=\texttt{ENCODE}(p)$. 

Equation~\ref{eq:scoring_dm} has two inputs: the prompt $p$ and the stochastically sampled initial latent $X_T$. Per Section~\ref{sec:prompt_influence} the user-provided prompt controls the overall score distribution $\mathcal{N}(\mu_{S_p}, \sigma_{S_p})$ while the initial noise determines \textit{where} on the distribution a generated image $I$ will sit. However, we note a key distinction between the \textit{generated image} $I$ and its human preference score $S_{p,I}$. Specifically, an image is a 3D matrix consisting of many elements, i.e., a $1024^2$ in RGB space  contains over 3M elements, necessitating large computation to \textit{generate}, such as the RDP and compute-intensive denoiser $\epsilon_\theta$. In contrast, the score is \textit{just} a single scalar value drawn from a distribution $S_{p, I}\sim \mathcal{N}(\mu_{S_p}, \sigma_{S_p})$ which necessitates smaller computation to \textit{predict}~\cite{white2021powerful, wen2020neural, canalys2024Smartphone, lu2023pinat}. Thus, we recast the score calculation Eq.~\ref{eq:scoring_dm} in a simpler format, 

\begin{equation}
    \centering
    \label{eq:paine}
    S_{p, I} = \Phi(c, X_T) = \Phi_{score}(\Phi_{prompt}(\texttt{ENCODE}(p)), \Phi_{noise}(X_T)),
\end{equation}
where $\Phi$ is the PAINE predictor that consists of a prompt-conditioning encoder $\Phi_{prompt}$, noise encoder $\Phi_{noise}$ and score predictor $\Phi_{score}$. Whereas existing open-weight human preference predictors $\phi$ receive the prompt $p$ and generated image $I$ as input, the PAINE predictor receives the DM-specific encoded prompt conditioning $c$ and stochastically-sampled initial latent $X_T$, $\texttt{size}(X_T) < \texttt{size}(I)$ as input to directly predict the scalar preference score.

The intuition and use-case of the PAINE predictor is simple: Given a user-provided prompt $p$, and desired number of output images $|B|$, the PAINE predictor will first sample $N >> |B|$ candidate initial noises. Each of the $N$ noises will be fed into the PAINE predictor alongside the prompt encoding $c$\footnote{This incurs no compute overhead since the DM encodes $p$ into $c$ anyway.}, estimating a series of scalars $\mathcal{S}_p =\{S_{p, I_1}, S_{p, I_2},...,S_{p, I_N}\}$ which represents the score each noise would receive \textit{if} the costly RDP and denoiser $\epsilon_\theta$ were invoked to generate an image using that noise. We then sort and rank the candidate noises using their corresponding score estimations in $\mathcal{S}_p$ and proceed to forward the top-$|B|$ candidates for full DM generation into images $\{I_{p, 1}, I_{p, 2}...,I_{p, |B|}\}$. 

This approach differs from existing literature in several ways: \textit{First, } we are considering the scenario where the user may want to generate more than one image per prompt, whereas the frameworks of prior literature~\cite{zhou2025golden, wang2025silent, guo2024initno, eyring2025noise, tang2024inference, li2025noisear} target academic benchmark numbers by generally focusing on finding a single optimal noise per prompt. Second, by structuring our predictor around the score distribution $\mathcal{N}(\mu_{S_p}, \sigma_{S_p})$, we more robustly account for noise and subjectivity in human preferences in images~\cite{mance2022noise, jaramillo2022radiologists, xu2023imagereward}. Further, we are able to manipulate the PAINE predictor components in Eq.~\ref{eq:paine} to provide additional functionality in a Naïve Bayesian~\cite{provost2013data, bishop2006pattern} manner to provide insight on how well a given DM \textit{can} perform on a given prompt, which we next describe. 

\subsection{\np: Prompt Generation Performance Estimation}
\label{sec:method_prompt_estimation}

Recall from Section~\ref{sec:prompt_influence} that the prompt $p$ primarily influences the score distribution, such that the most/least optimal noise is not consistent across prompts, but rather plays a role in sampling from the score distribution. We train PAINE on a dataset consisting of many $(p, X_T, S_{p, I})$ tuples to predict $S_{p, I}$. In this dataset $p$ is not unique and appears in many tuples, but is matched with different $X_T$ tensors and $S_{p, I}$ scalars. Further details are provided in Section~\ref{sec:setup}.

Given this framing, we can further exploit the PAINE architecture to provide additional functionality by predicting $\mu_{S_p}$, the general performance of a given DM denoiser $\epsilon_\theta$ on the user-provided prompt $p$. Specifically, we %
mask $\Phi_{noise}$ by %
zeroing the output \textit{prior} to combining it with the output of $\Phi_{prompt}$ and feeding it into $\Phi_{score}$ to predict $S_{p, I}$. The intuition %
is that in the absence of noise information $\Phi_{noise}(X_T)$, $\Phi_{score}$ will instead try to approximate $\mu_{S_p}$. 

This structure is loosely inspired by classical Naïve Bayesian decomposition $p(H|E) \propto p(E|H)p(H)$~\cite{russell2010artificial, poole2010artificial, provost2013data, bishop2006pattern}. %
Specifically, in the absence of noise information $\Phi_{noise}(X_T)$, PAINE estimates a \textit{prior} $p(H)=\mu_{S_p}$. %
In this framing the 
noise information $\Phi_{noise}(X_T)$ %
acts as the likelihood $p(E|H)=\zeta\sigma_{p, I}$; $\zeta \sim \mathcal{N}(0, 1)$ necessary for us to model the full posterior $p(H|E)=S_{p, I}=\mu_{S_p}+\zeta\sigma_{p, I}$; again note that this is an analogy and we are not claiming a formal derivation. Rather, we articulate the layman explanation of this analogy as `What is the average human preference score for this noise given how well the DM can generate given the provided prompt.' 
Thus, when embedded into an existing DM pipeline~\cite{von-platen-etal-2022-diffusers}, \np\ not only enables us to perform initial noise optimization, but to manually tweak our prompt %
to maximize the quality of generated images prior to generation. %

\section{Results}
\label{sec:results}

In this section we evaluate \np\ on a suite of T2I DMs. Specifically, we enumerate predictor training and pipeline integration in %
Section~\ref{sec:setup}. %
Next, Sections~\ref{sec:eval} and \ref{sec:eval_prompt} sweep our evaluation compared to contemporary literature as well as our ability to model preference metric means, respectively. 
Finally, we provide ablation results in Section~\ref{sec:ablation}.

\subsection{Predictor Architecture, Datasets, Training and Evaluation}
\label{sec:setup}

We train and evaluate \np\ on four T2I DMs: Hunyuan-DiT~\cite{li2024hunyuandit}, PixArt-$\Sigma$~\cite{chen2024pixartsigma}, DreamShaper-XL-v2-Turbo~\cite{lykon2023dreamshaper} and SDXL~\cite{podell2023sdxl} by generating $1024^2$ RGB images. For all experiments in this paper, we use the Diffusers~\cite{von-platen-etal-2022-diffusers} DM pipelines unless otherwise indicated. 

To train our predictor, we assemble a training set of prompt, noise and score $(p, X_T, S_{p, I})$ tuples. Specifically, we sample 5k random prompts from the Pick-a-Pic~\cite{kirstain2023pick} training dataset. For each DM, we loop over the prompt set, generating 20 images per prompt with different initial noises. For each prompt-image pair we measure several human preference metrics~\cite{wu2023human, xu2023imagereward, hessel2021clipscore} but primarily rely on PickScore~\cite{kirstain2023pick} as the target $S_{p, I}$. 

The \np\ predictor architecture $\Phi$ consists of three modules. First, the prompt encoder $\Phi_{prompt}$ receives the prompt encoding $c = \texttt{ENCODE}(p)$ and processes it into a vector. To do this, $\Phi_{prompt}$ consists of a single transformer layer where we append an additional, learnable token to the end of the sequence prior to the self-attention mechanism. This additional token is passed through the remainder of the transformer block while the original sequence is discarded. Further, we instantiate one distinct $\Phi_{prompt}$ per DM text encoder, i.e., one for PixArt-$\Sigma$ which relies solely on a T5 LLM~\cite{raffel2020exploring}, and two for Hunyuan/SDXL/DreamShaper which all use dual text encoders~\cite{radford2021learning}. 

The noise encoder $\Phi_{noise}$ consists of a simple 4-stage ResNet~\cite{he2016deep} encoder which gradually downsamples the height and width of the latent $X_T$ while increasing the number of channels. The final layer of $\Phi_{noise}$ is an adaptive max pooling layer which compresses the remaining height and width down to 1 for %
a vector output. Finally, the score encoder $\Phi_{score}$ simply concatenates the output of the noise and prompt encoder(s) and then feeds them through an MLP to output a scalar prediction. %

For training, we split our data 80\%/10\%/10\%  into distinct training, validation, and test sets. These splits are conducted over the prompts so all tuples corresponding to a given prompt $p$ will only appear in one partition. The overall loss function consists of the Mean Absolute Error (MAE) regression loss and the differentiable Spearman's Rank Correlation Coefficient (SRCC) from Blondel et al., 2020~\cite{blondel2020fast}. Specifically, the differentiable SRCC loss is computed by measuring the soft ranking correlation of ground-truth targets and predictions in a given batch. \np\ trains for 50 epochs with grouped batches of $k=6$ prompts (120 samples per batch) using the AdamW~\cite{loshchilov19AdamW} optimizer. %
More details can be found in the supplementary. %

\begin{table}[!t]
\begin{center}
\caption{T2I Quantitative Comparison on several human preference benchmarks and prompt datasets. Best and second best results in \textbf{bold} and \textit{italics}, respectively.}

\scalebox{\scaleboxratio}{
\begin{tabular}{ccccccc}
\toprule
Model & Dataset & Method & HPSv2~(↑) & HPSv3~(↑) & ImageReward~(↑) & PickScore~(↑) \\ \midrule

\multirow{12}{*}{PixArt-$\Sigma$}
&  \multirow{3}{*}{Pick-a-Pic} & Standard     & \textit{29.83} & \textit{8.60} & \textit{81.24} & \textbf{21.90} \\
&                              & Golden Noise & 28.08 & 7.91 & 68.13 & 21.63  \\
&                              & \np          & \textbf{30.82} & \textbf{8.73} & \textbf{86.02} & \textit{21.82} \\ \cline{2-7} 

&  \multirow{3}{*}{HPDv2}      & Standard     & \textit{30.91} & \textit{10.53} & \textbf{106.45} & \textit{22.98} \\
&                              & Golden Noise & 28.97 & 10.09 & 98.39 & 22.82  \\
&                              & \np          & \textbf{32.71} & \textbf{10.58} & \textit{106.11} & \textbf{23.01} \\ \cline{2-7} 

&  \multirow{3}{*}{HPSv2}      & Standard     & \textit{31.70} & \textit{11.93} & \textit{103.21} & \textit{22.68} \\
&                              & Golden Noise & 28.95 & 11.56 & 97.96 & 22.52  \\
&                              & \np          & \textbf{31.79} & \textbf{11.96} & \textbf{103.23} & \textbf{22.70} \\ \cline{2-7} 

&  \multirow{3}{*}{DrawBench}  & Standard     & 28.54 & \textbf{9.60} & \textit{73.41} & \textit{22.33} \\
&                              & Golden Noise & \textit{28.64} & 8.34 & 57.95 & 22.09  \\
&                              & \np          & \textbf{30.03} & \textit{9.57} & \textbf{75.60} & \textbf{22.34} \\ \midrule 

\multirow{12}{*}{Hunyuan}
&  \multirow{3}{*}{Pick-a-Pic} & Standard & 29.48 & 8.06 & 85.47 & 21.81 \\
&                              & Golden Noise & \textit{29.94} & \textit{8.09} & \textit{87.00} & \textit{21.85} \\
&                              & \np     & \textbf{30.77} & \textbf{8.24} & \textbf{89.52} & \textbf{21.86} \\ \cline{2-7} 

&  \multirow{3}{*}{HPDv2}      & Standard & 31.04 & 10.01 & \textit{109.27} & 22.88 \\
&                              & Golden Noise & \textit{31.20 }& \textbf{10.12} & 108.29 & \textit{22.89}  \\
&                              & \np     & \textbf{32.79} & \textit{10.02} & \textbf{114.78} & \textbf{23.00} \\ \cline{2-7} 

&  \multirow{3}{*}{HPSv2}      & Standard & \textit{30.06} & 11.42 & 106.41 & \textit{22.58} \\
&                              & Golden Noise & 29.00 & \textbf{11.70} & \textit{108.25} & \textbf{22.61} \\
&                              & \np     & \textbf{31.70} & \textit{11.43} & \textit{108.42} & \textit{22.58} \\ \cline{2-7} 

&  \multirow{3}{*}{DrawBench}  & Standard & 28.49 & \textit{9.18} & 84.45 & 22.31 \\
&                              & Golden Noise & \textit{29.02} & \textbf{9.66} & \textit{87.96} & \textbf{22.40} \\
&                              & \np     & \textbf{30.16} & 9.10 & \textbf{89.88} & \textit{22.33}\\ \midrule

\multirow{12}{*}{DreamShaper}
&  \multirow{3}{*}{Pick-a-Pic} & Standard     & 32.51 & \textit{10.33} & 98.06 & 22.60 \\
&                              & Golden Noise & \textit{32.69} & \textbf{10.36} & \textit{103.91} & \textbf{22.69}  \\
&                              & \np          & \textbf{33.27} & 10.29 & \textbf{106.60} & \textit{22.64} \\ \cline{2-7} 

&  \multirow{3}{*}{HPDv2}      & Standard     & \textit{33.95} & \textit{12.38} & \textit{129.51} & \textbf{23.88} \\
&                              & Golden Noise & 29.85 & \textbf{12.78} & \textbf{133.04} & \textit{23.62}  \\
&                              & \np          & \textbf{34.85} & 12.01 & 119.49 & 23.59\\ \cline{2-7} 

&  \multirow{3}{*}{HPSv2}      & Standard     & \textit{34.39} & 13.90 & 121.91 & \textit{23.42} \\
&                              & Golden Noise & 29.70 & \textbf{14.19} & \textbf{123.66} & 23.39  \\
&                              & \np          & \textit{34.44} & \textit{13.94} & \textit{123.24} & \textbf{23.44} \\ \cline{2-7} 

&  \multirow{3}{*}{DrawBench}  & Standard     & \textit{30.17} & 11.57 & 96.37 & 22.92 \\
&                              & Golden Noise & 29.51 & \textbf{12.17} & \textbf{97.88} & \textbf{23.03}  \\
&                              & \np          & \textbf{31.58} & \textit{11.70} & \textit{97.28} & \textit{22.93} \\ \midrule 

\multirow{12}{*}{SDXL}
&  \multirow{3}{*}{Pick-a-Pic} & Standard     & 28.40 & 7.10 & 57.94 & 21.65 \\
&                              & Golden Noise & \textit{28.68} & \textit{7.20} & \textit{65.01} & \textbf{21.85}  \\
&                              & \np          & \textbf{28.85} & \textbf{7.90} & \textbf{80.86} & \textit{21.82} \\ \cline{2-7} 

&  \multirow{3}{*}{HPDv2}      & Standard     & 29.66 & 8.73 & 82.66 & 22.83 \\
&                              & Golden Noise & \textit{29.88} & \textbf{9.07} & \textit{98.81} & \textbf{22.94}  \\
&                              & \np          & \textbf{31.70} & \textit{8.95} & \textbf{100.78} & \textit{22.93} \\ \cline{2-7} 

&  \multirow{3}{*}{HPSv2}      & Standard     & \textit{28.69} & \textit{10.24} & 85.06 & 22.52 \\
&                              & Golden Noise & \textbf{28.74} & \textbf{10.78} & \textit{91.07} & \textit{22.60}  \\ 
&                              & \np          & 28.67 & 10.21 & \textbf{92.37} & \textbf{23.01} \\ \cline{2-7} 

&  \multirow{3}{*}{DrawBench}  & Standard     & 26.83 & 7.88 & 60.08 & 22.28 \\
&                              & Golden Noise & \textit{27.14} & \textbf{8.52} & \textbf{67.92} & \textbf{22.38}  \\
&                              & \np          & \textbf{28.66} & \textit{8.16} & \textit{66.00} & \textit{22.31} \\ 
\bottomrule
\end{tabular}
}
\vspace{-5mm}
\label{table:main_exp_full}
\end{center}
\end{table}

Once trained, our \np\ predictors integrate into existing DM pipelines as follows: In addition to the prompt $p$, number of images to generate $|B|$ and other hyperparameters, the user provides the number of initial noises $N$ to sample. After the pipeline encodes the prompt, $c = \texttt{ENCODE}(p)$, we sample $N=100$ noises and estimate their scores $S_{p, I}$. %
We sort and rank the noises, forwarding only the top-$|B|$ for full generation. Alternatively, given prompt encoding $c$, \np\ can also simply estimate the mean preference score $\mu_{S_p}$.

\subsection{T2I Evaluation}
\label{sec:eval}

We evaluate \np\ on a variety of benchmark prompt sets~\cite{kirstain2023pick, wu2023human, saharia2022photorealistic}, human preference predictors $\phi$ such as HPSv2~\cite{wu2023human}, HPSv3~\cite{ma2025hpsv3}, ImageReward~\cite{xu2023imagereward} and PickScore~\cite{kirstain2023pick}. %
In terms of baselines we primarily compare to Golden Noise~\cite{zhou2025golden} who mutate the initial noise $X_T$ given the prompt and a standard baseline where no initial noise optimization is performed. We also provide some comparisons to NoiseAR~\cite{li2025noisear} who use RL fine-tuning to autoregressively modify the DM generation process. Please note that while these methods also perform initial noise optimization, like \np, they have no equivalent to the candidate pool size parameter $N$; $N=100$ in our comparisons. %

\noindent\textbf{Human preference benchmarks.} Table~\ref{table:main_exp_full} reports our findings for 64 comparisons across all DM, prompt set and human preference metric $\phi$ combinations. %
\np\ achieves the best performance on more than 30 of these comparisons and the second best performance on more than 20. Further, even though we primarily train our predictor on PickScore, we find that the performance generalizes across other preference metrics, specifically HPSv2 and v3, as well. In terms of performance on specific DMs, we find that our method is generally stronger on newer DiT models like Hunyuan and PixArt-$\Sigma$, but is also very competitive on older U-Nets like SDXL and its fine-tune, DreamShaper. Overall, considering the inherent noisiness of human preference metrics~\cite{mance2022noise, jaramillo2022radiologists,  xu2023imagereward}, these results demonstrate the generative competitiveness of our proposed method. 

Also, both our training target and evaluation metrics are learned preference predictors rather than direct human judgments, and may share correlated biases from similar training data or architectures. A large-scale human study would more directly validate these gains, but is costly and beyond this work's scope; we instead rely on generalization across four independent, widely-adopted predictors as a practical proxy for robustness. Additional quantitative results, including on the GenEval~\cite{ghosh2023geneval} benchmark, can be found in the supplementary.

\begin{figure}[t!]
    \centering
    \setlength{\tabcolsep}{1.2pt}
    \renewcommand{\arraystretch}{1.2}
    \footnotesize
    \scalebox{\scaleboxratio}{
    \begin{tabular}{c c c c}
        \begin{tabular}{c}Standard\end{tabular} &
        \begin{tabular}{c}Golden Noise\end{tabular} &
        \begin{tabular}{c}NoiseAR\end{tabular} &
        \begin{tabular}{c}\np\end{tabular} \\

        \includegraphics[width=0.3\textwidth]{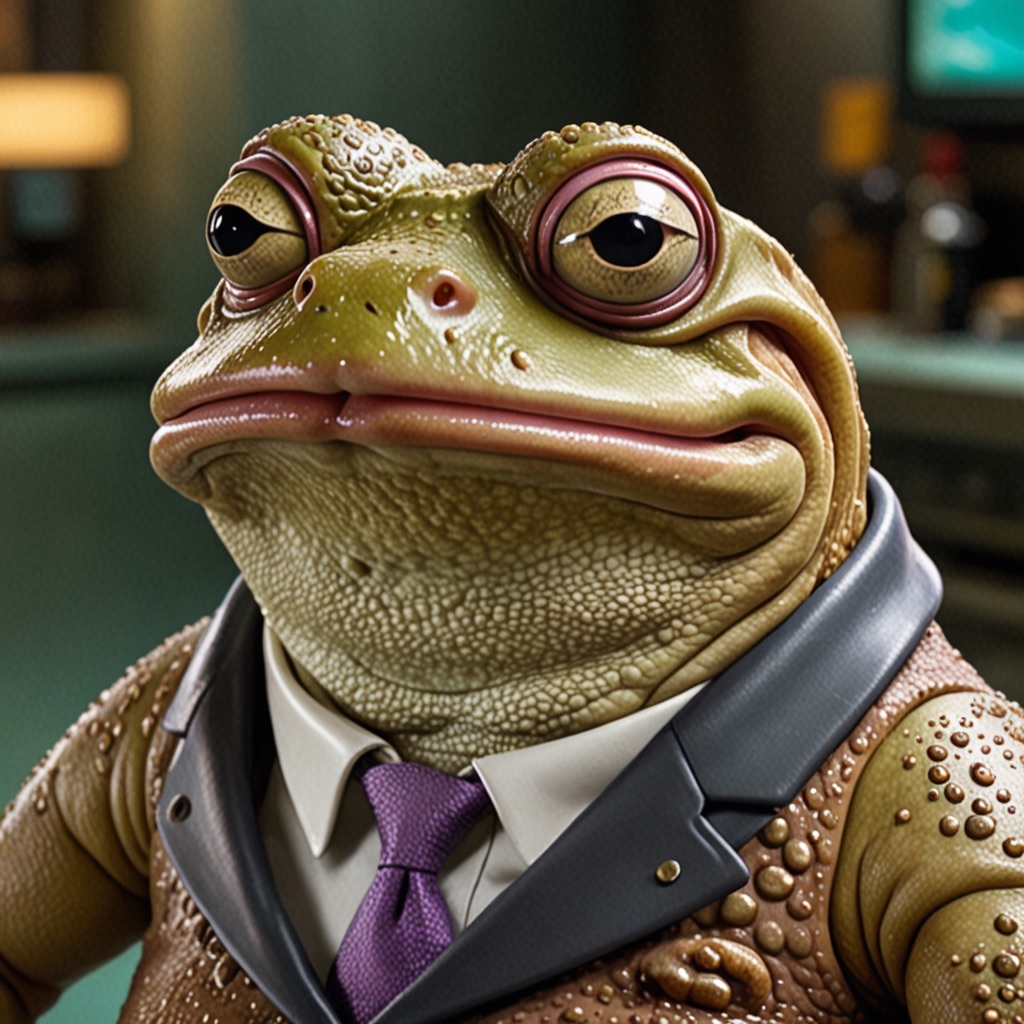} &
        \includegraphics[width=0.3\textwidth]{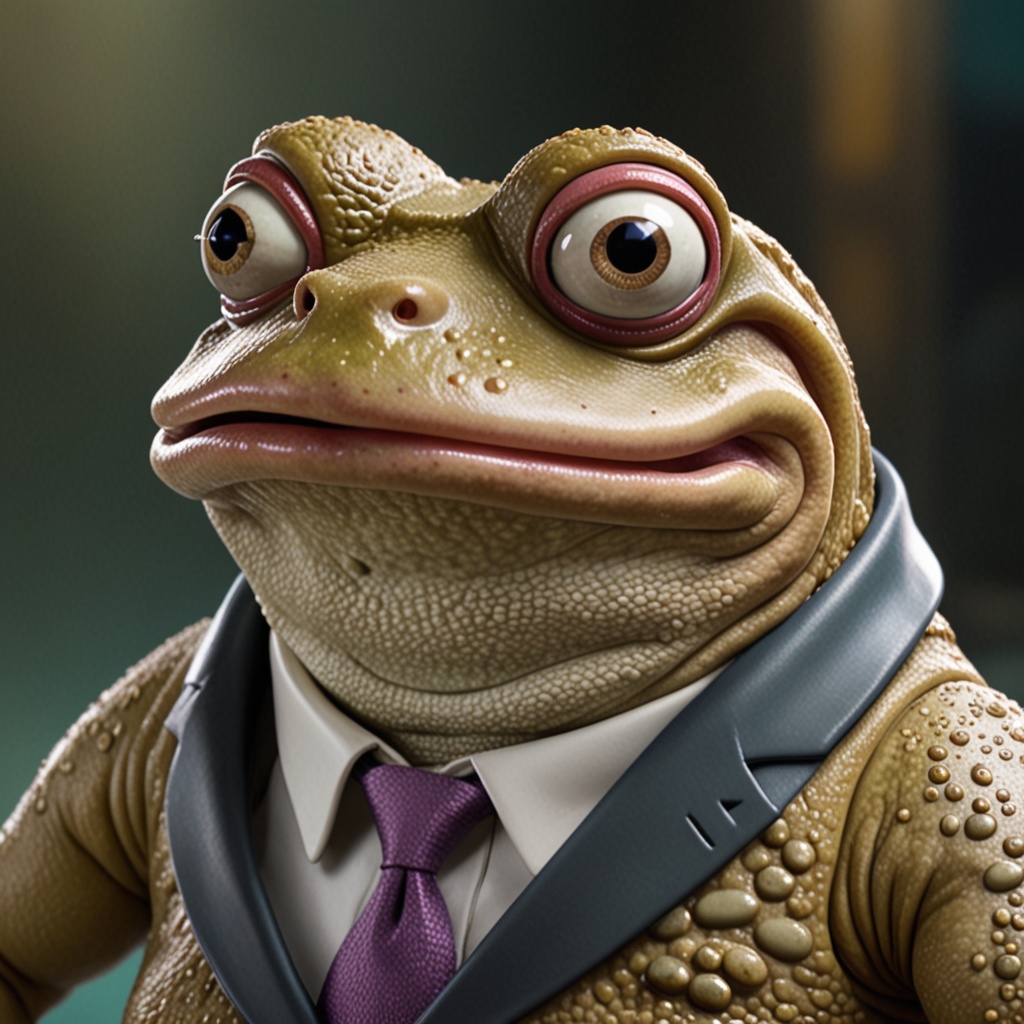} &
        \includegraphics[width=0.3\textwidth]{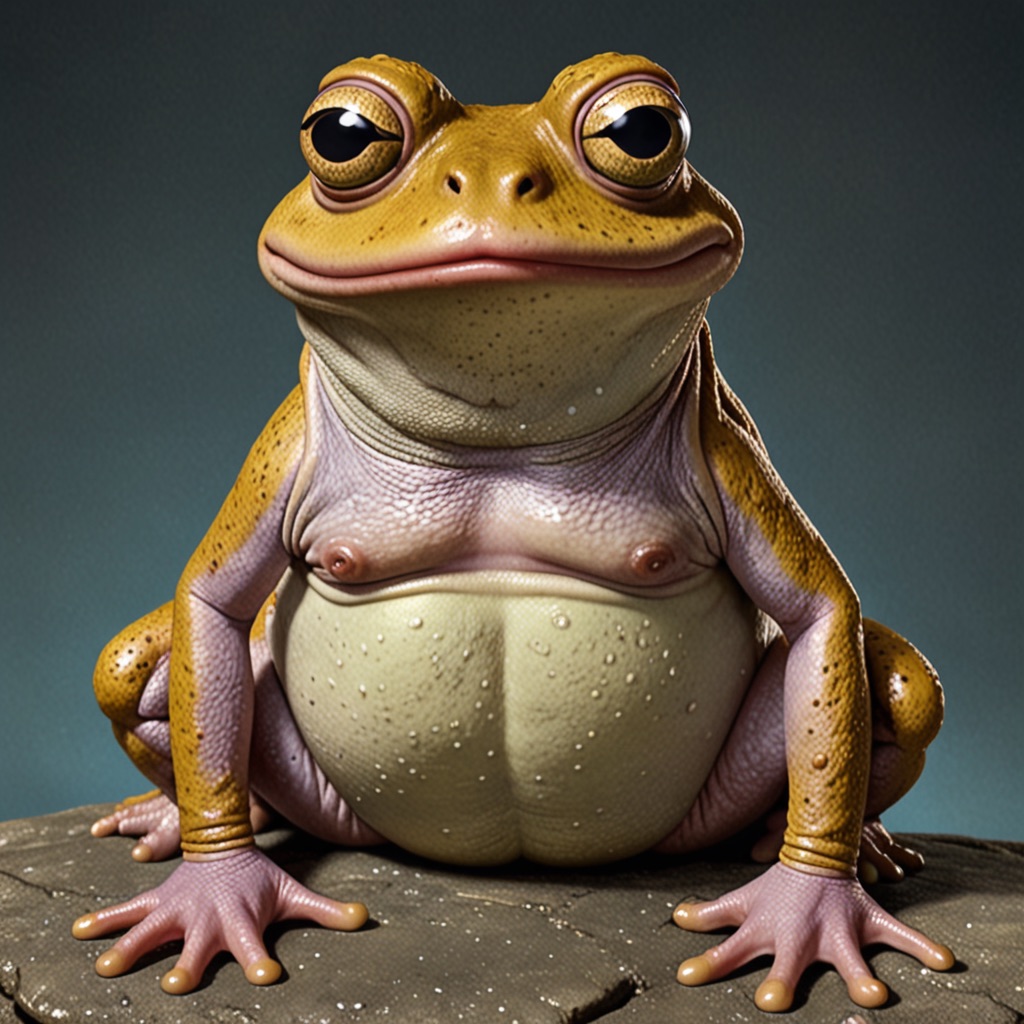} &
        \includegraphics[width=0.3\textwidth]{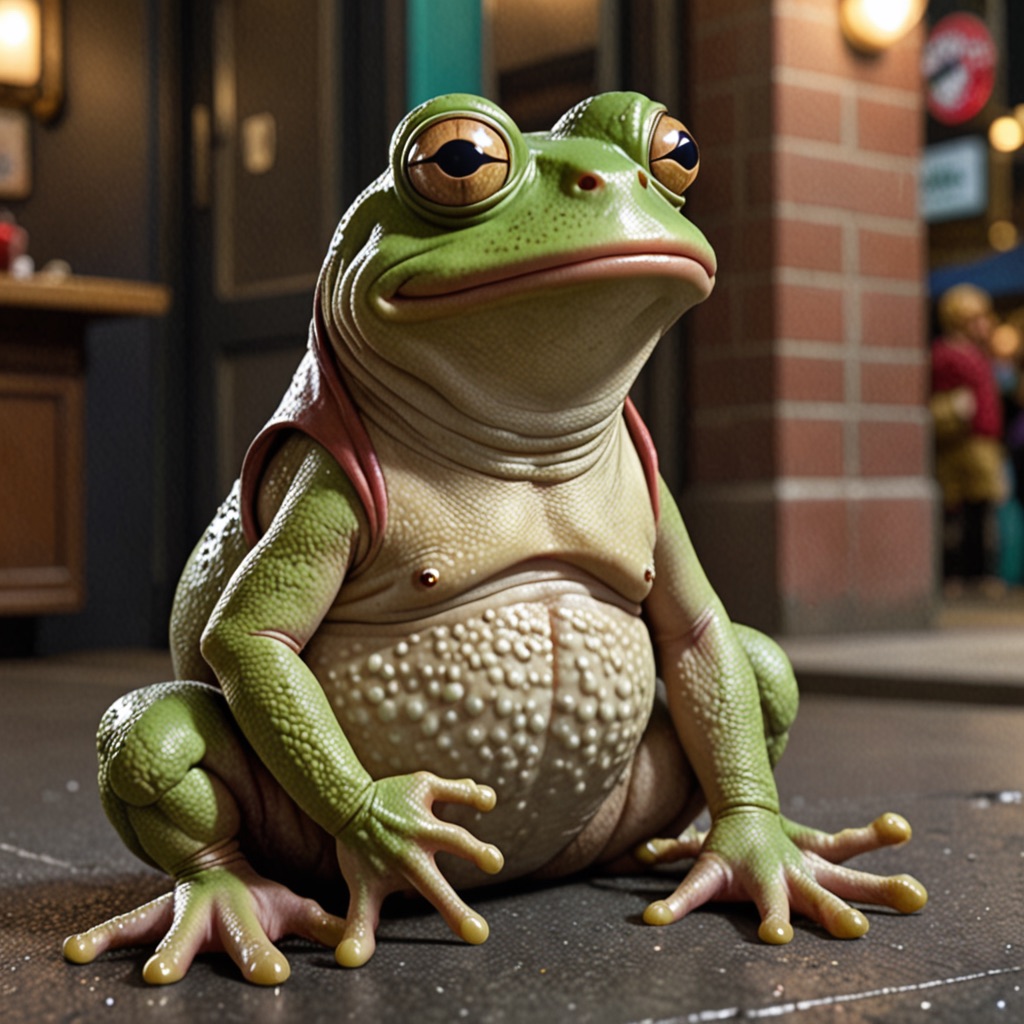} \\
        \multicolumn{4}{l}{\begin{tabular}{l}DreamShaper $\times$ Pick-a-pic Prompt: ``A hyper-realistic representation of the hypnotoad from Futurama''\end{tabular}} \\

        \includegraphics[width=0.3\textwidth]{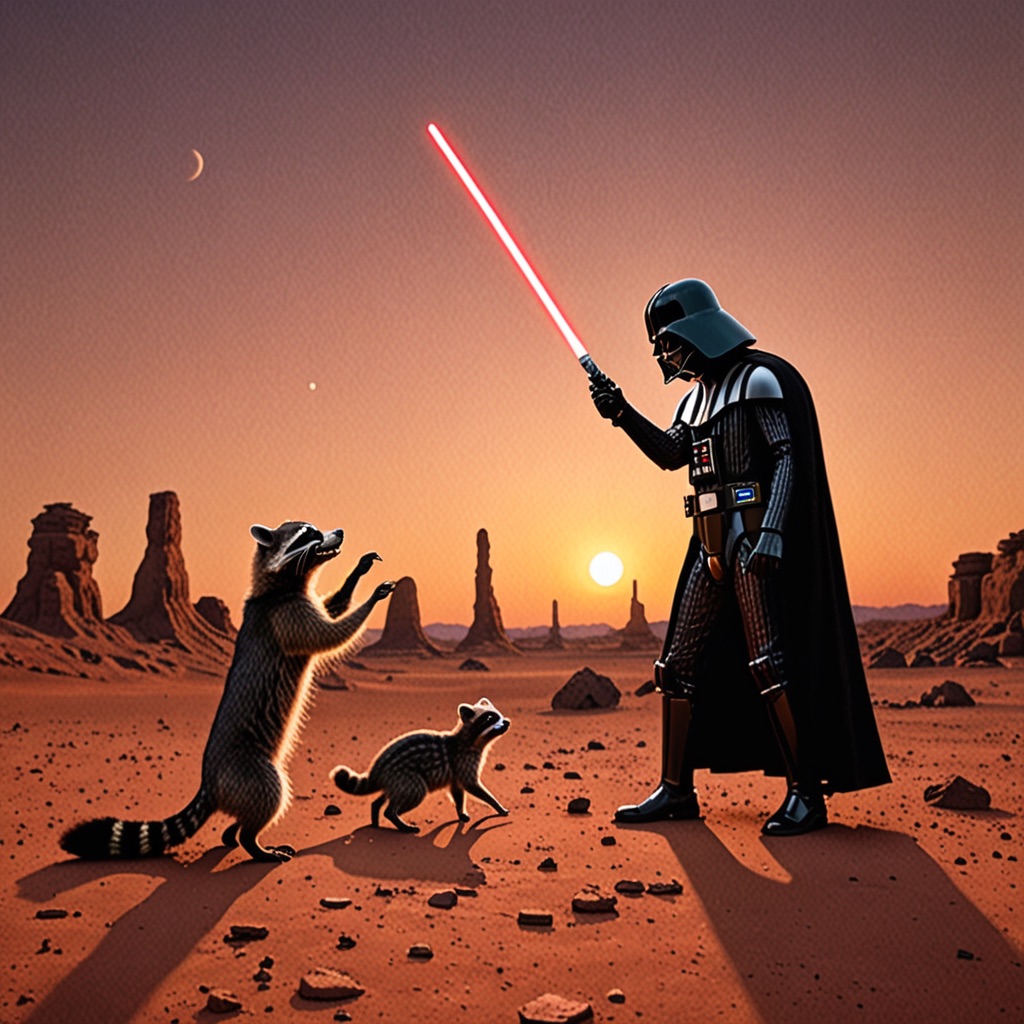} &
        \includegraphics[width=0.3\textwidth]{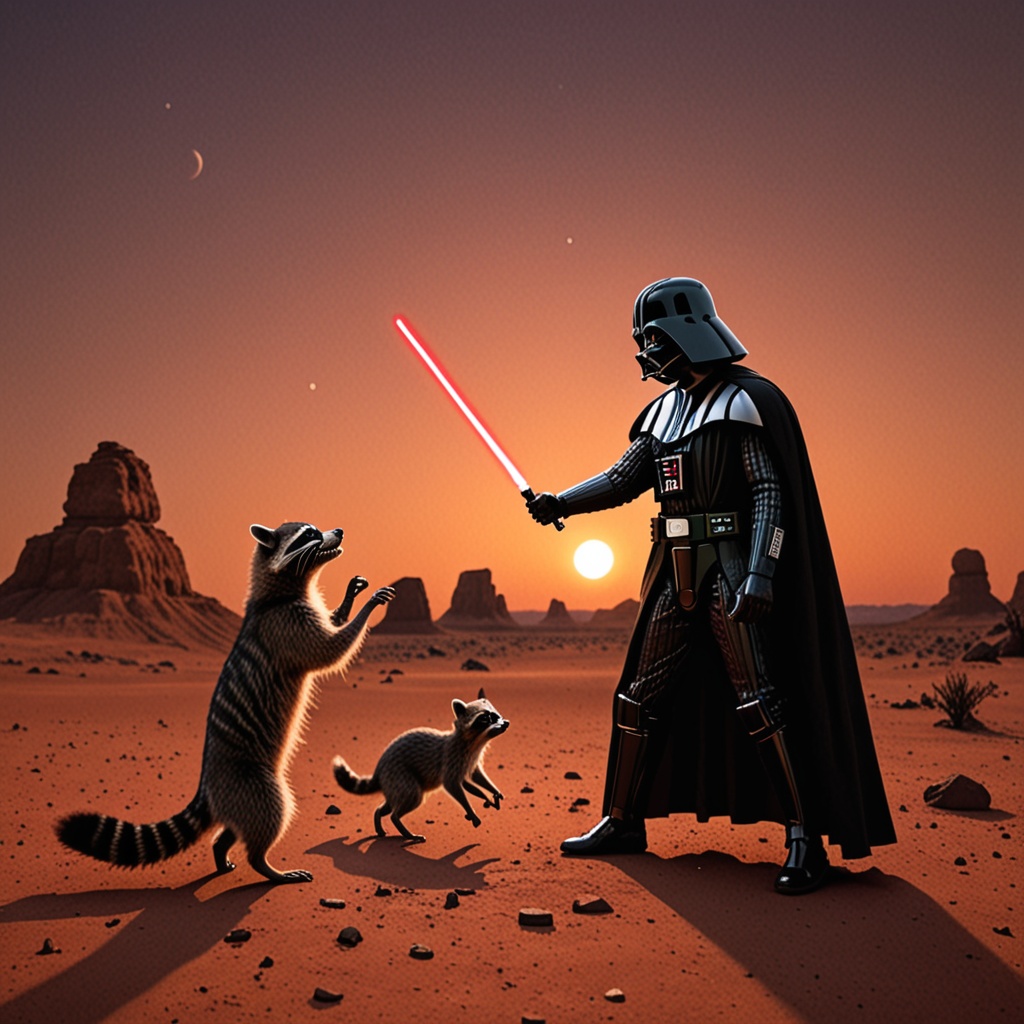} &
        \includegraphics[width=0.3\textwidth]{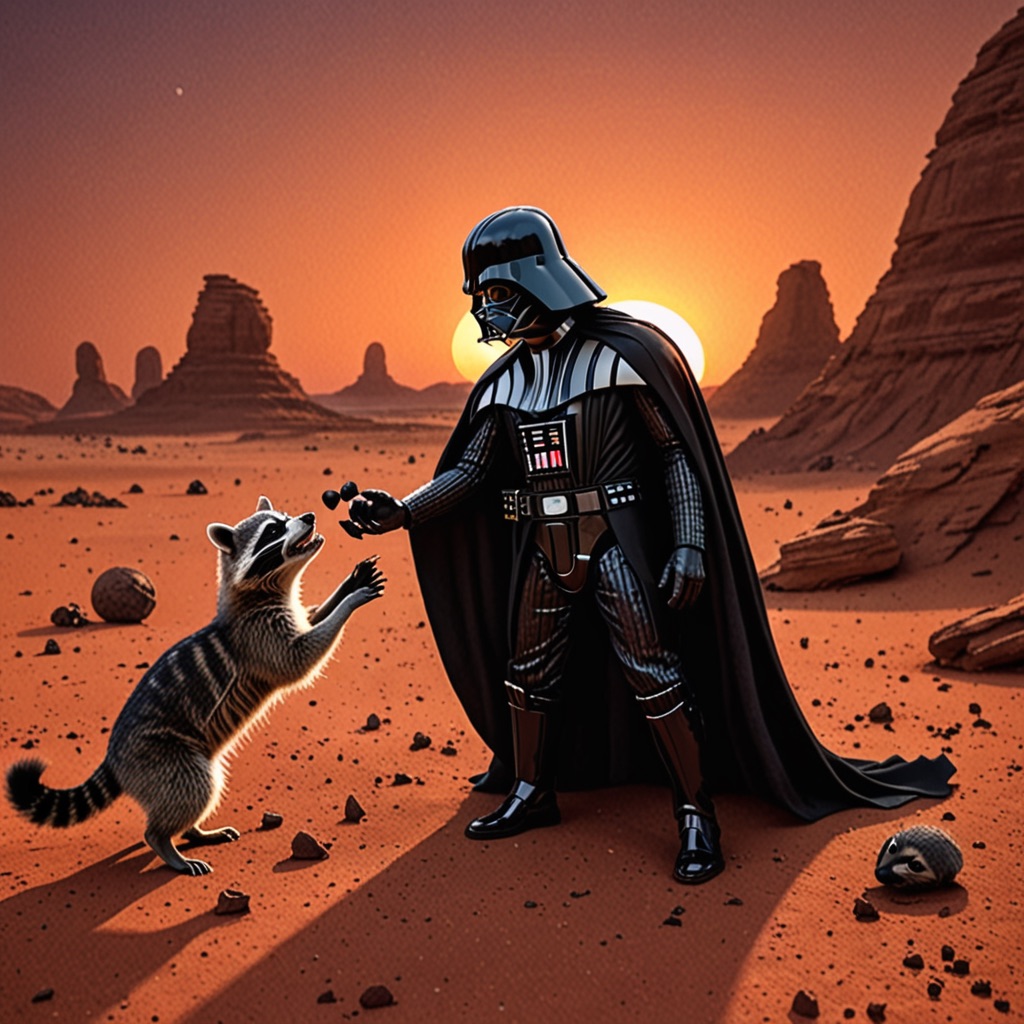} &
        \includegraphics[width=0.3\textwidth]{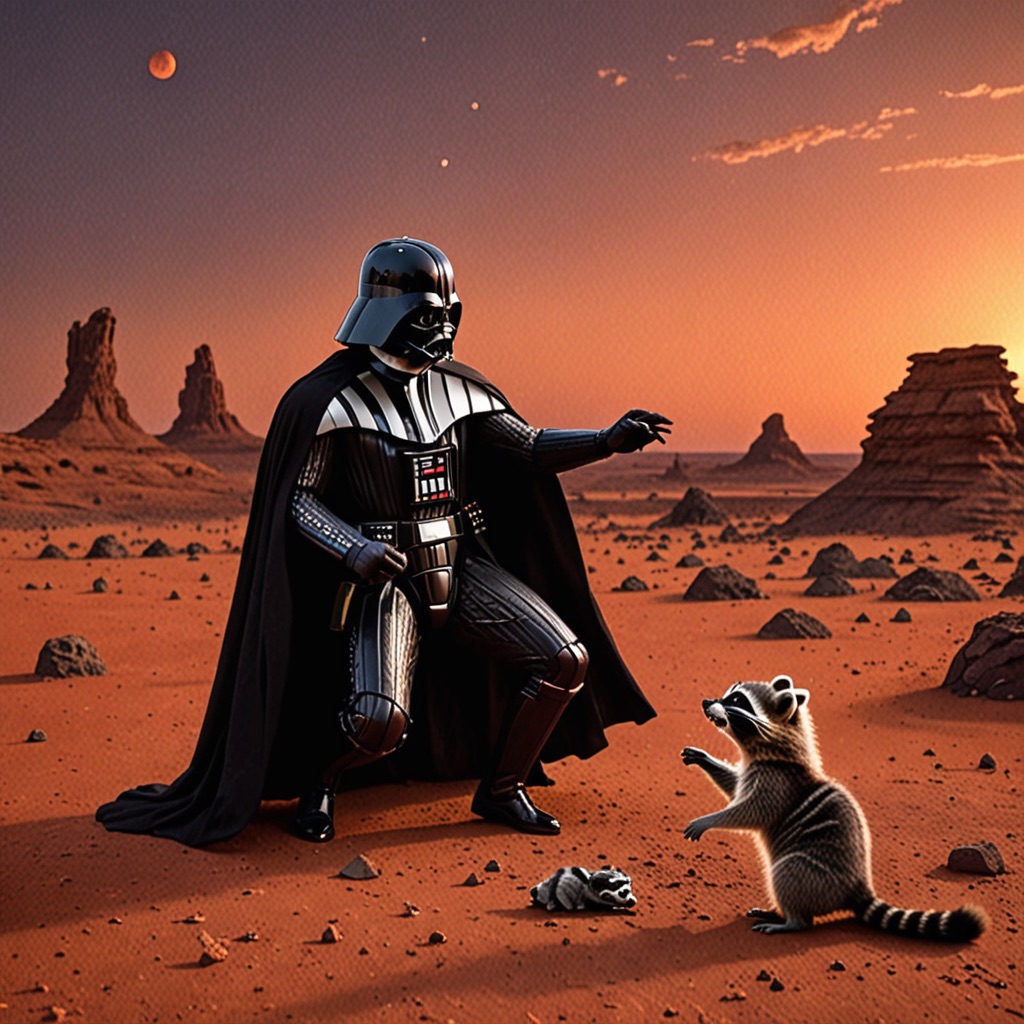} \\
        \multicolumn{4}{l}{\begin{tabular}{l}DreamShaper $\times$ DrawBench Prompt: ``Darth Vader playing with raccoon in Mars during sunset''\end{tabular}} \\

        \includegraphics[width=0.3\textwidth]{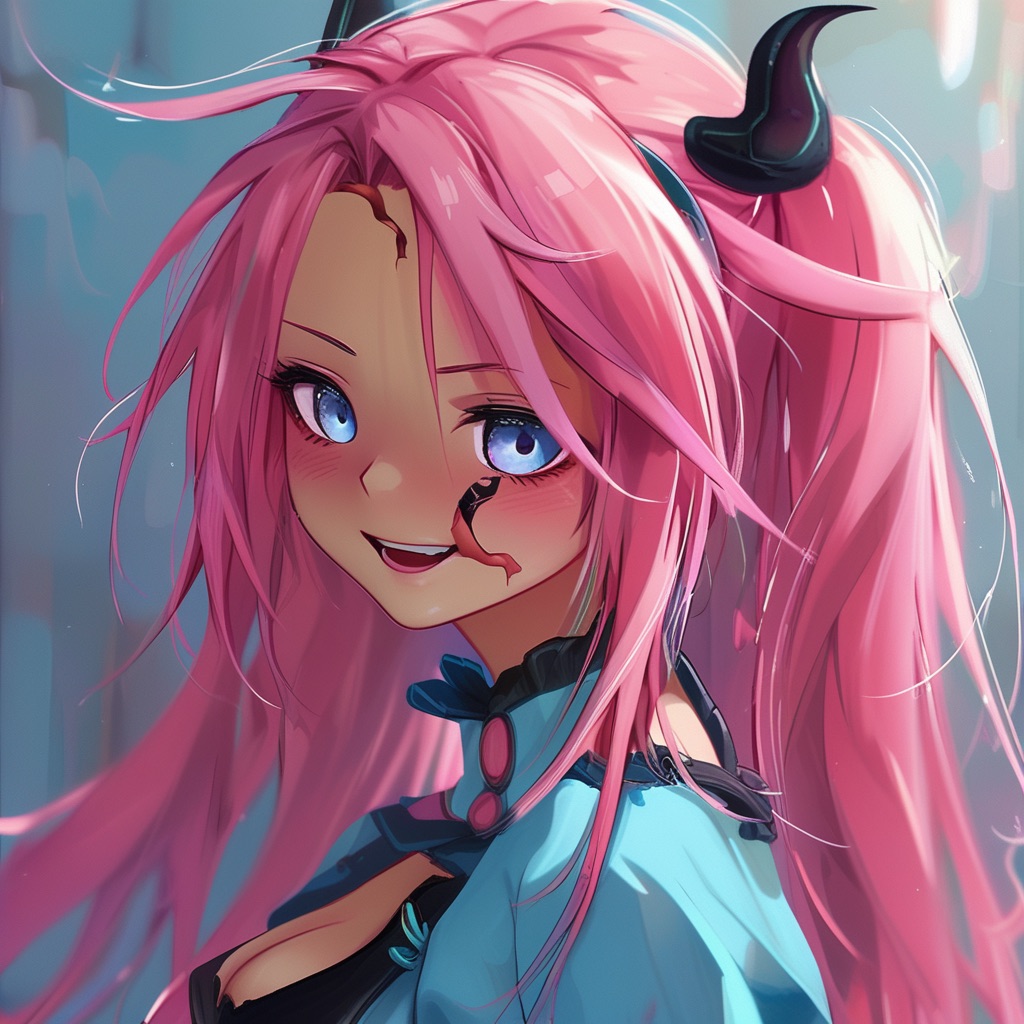} &
        \includegraphics[width=0.3\textwidth]{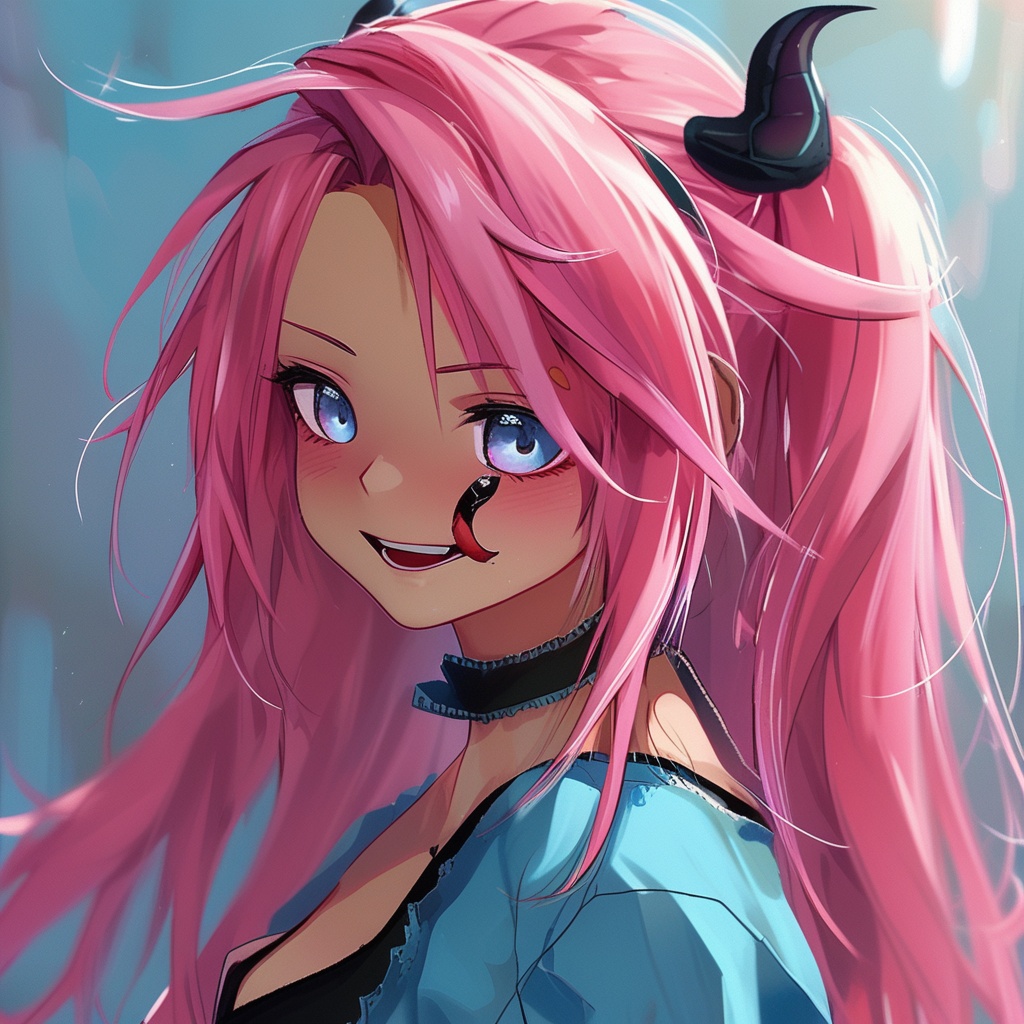} &
        \includegraphics[width=0.3\textwidth]{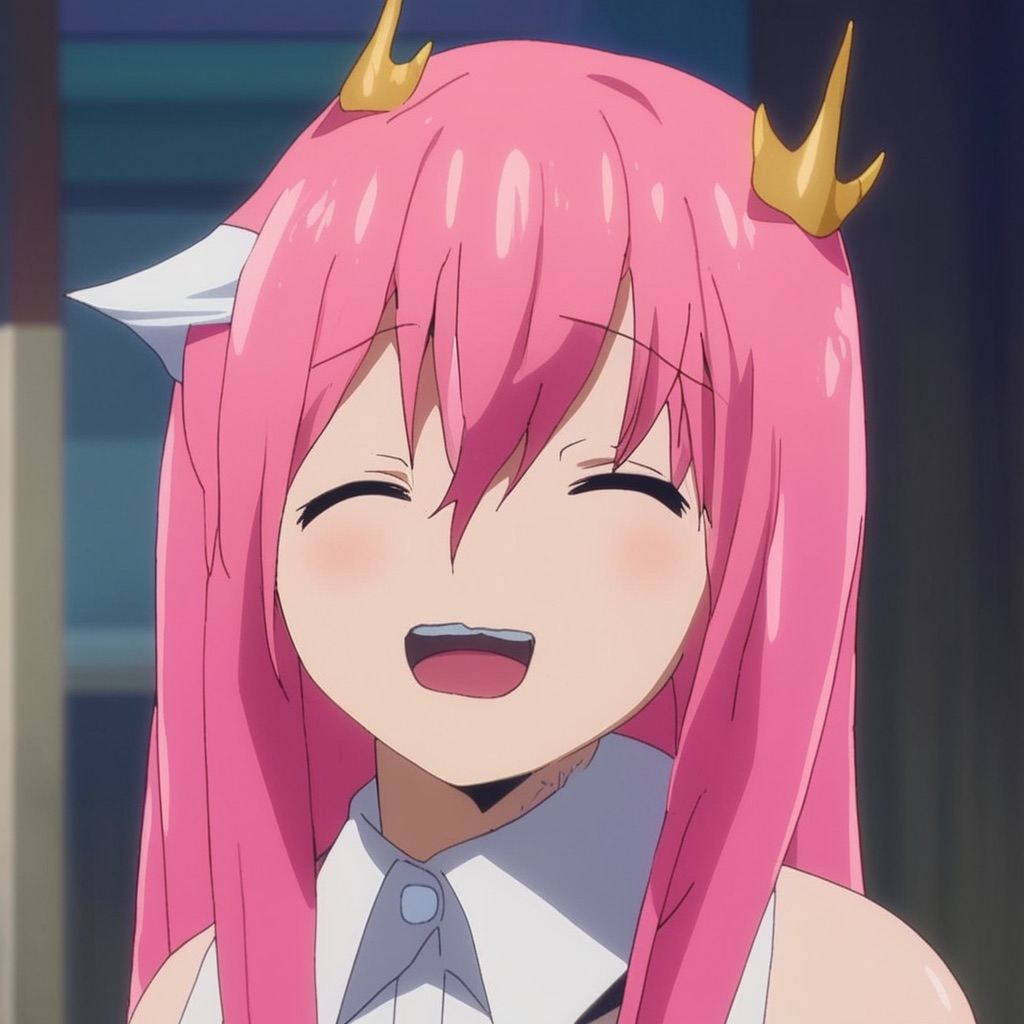} &
        \includegraphics[width=0.3\textwidth]{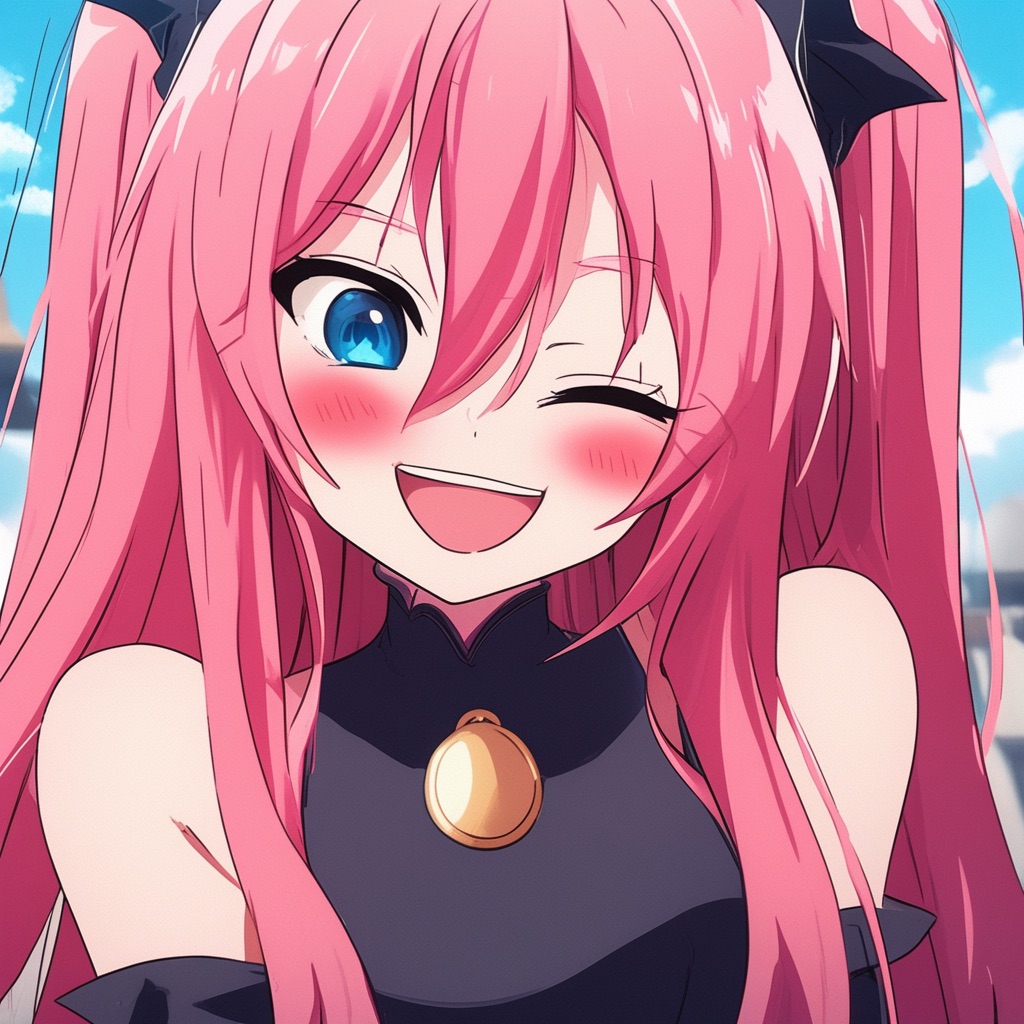} \\
        \multicolumn{4}{l}{\begin{tabular}{l}Hunyuan $\times$ Pick-a-pic Prompt: ``milim, pink hair, that awesome time i got reincarnated as a slime''\end{tabular}} \\

        \includegraphics[width=0.3\textwidth]{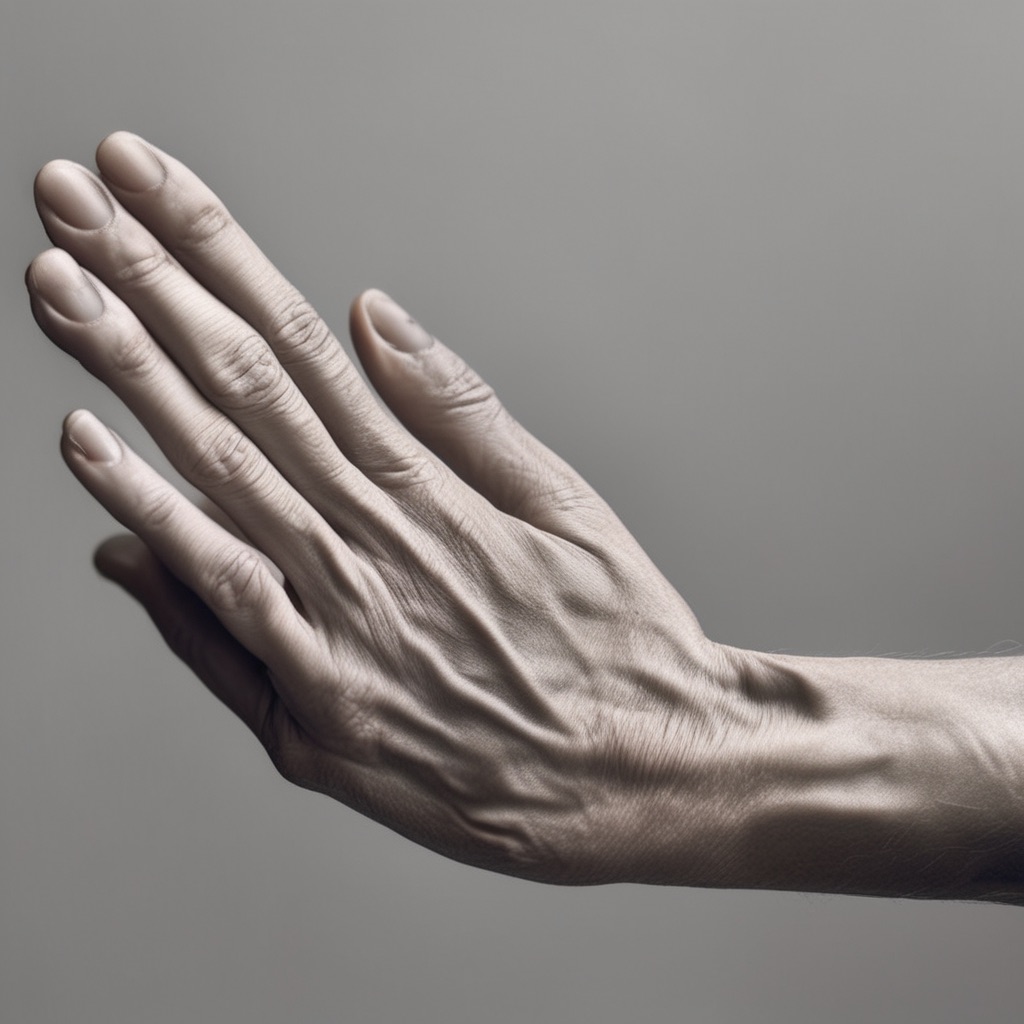} &
        \includegraphics[width=0.3\textwidth]{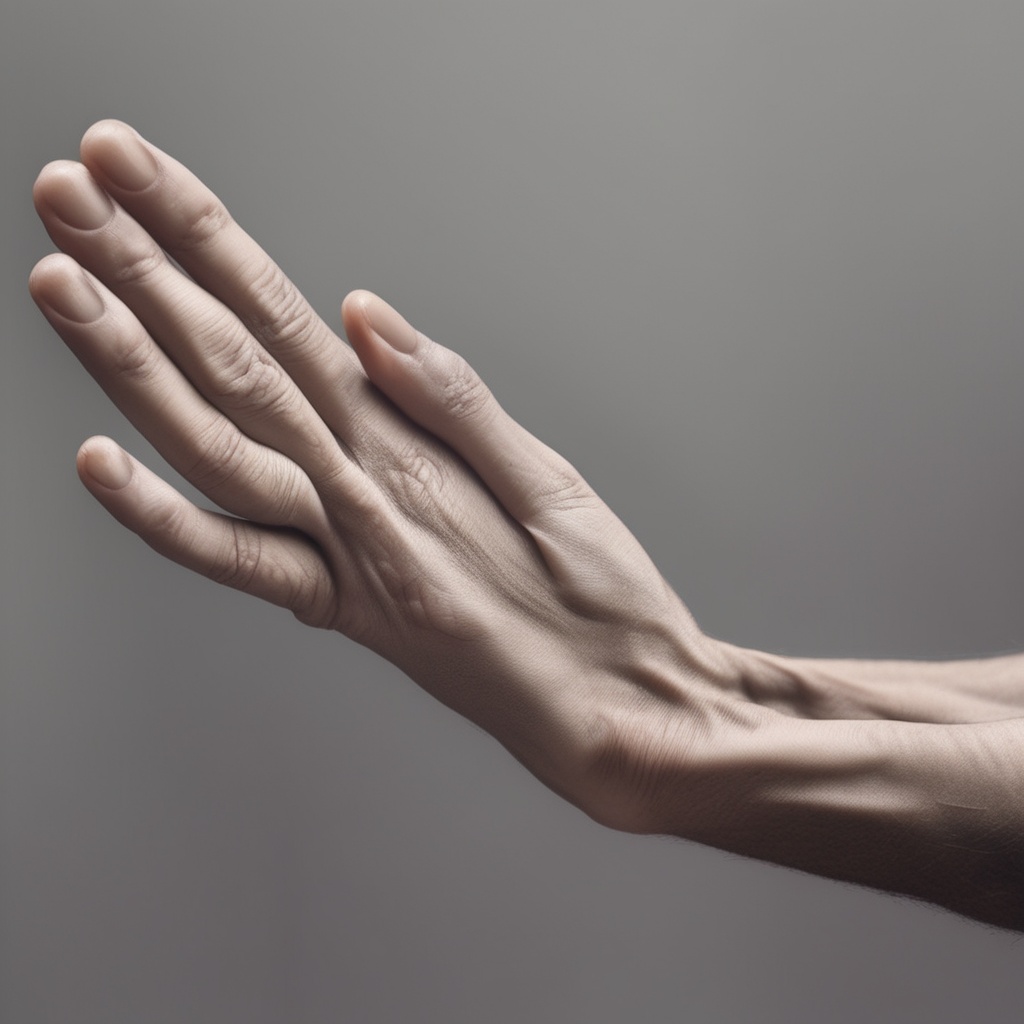} &
        \includegraphics[width=0.3\textwidth]{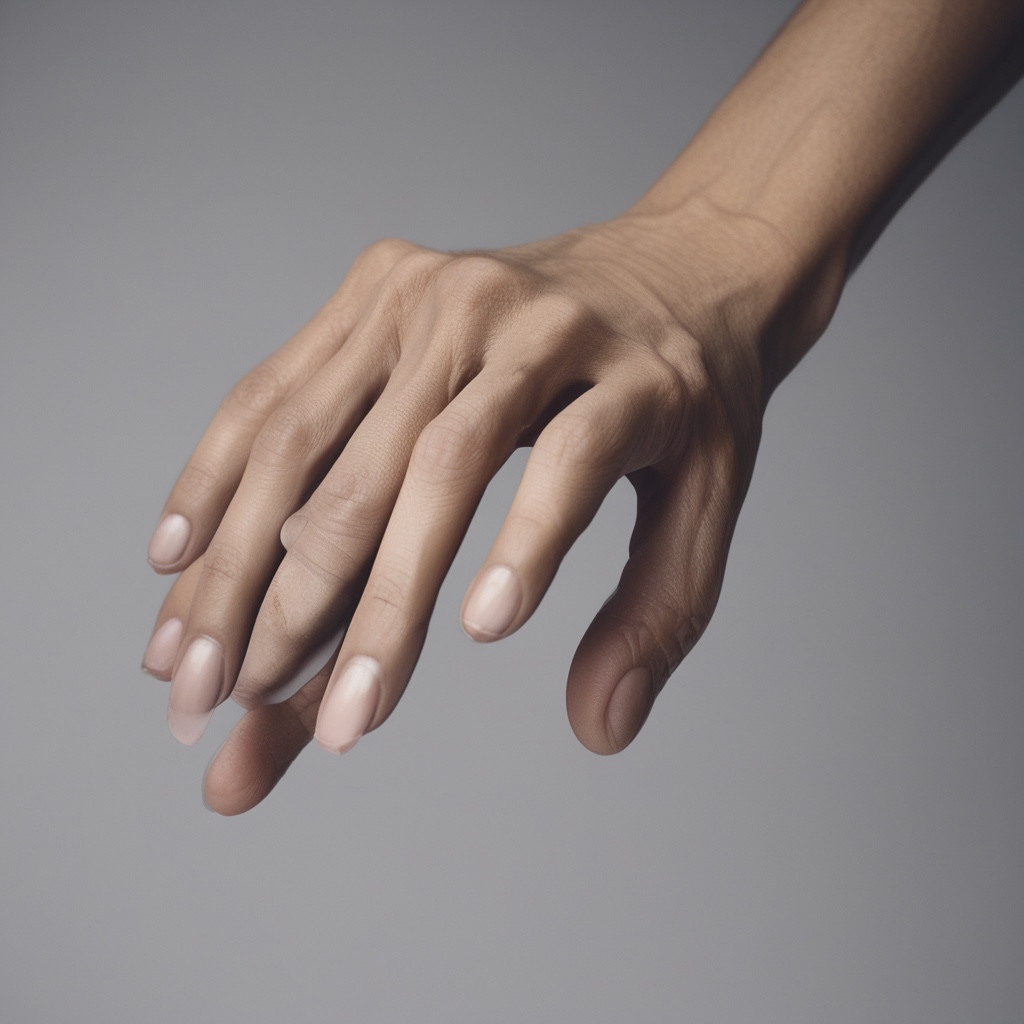} &
        \includegraphics[width=0.3\textwidth]{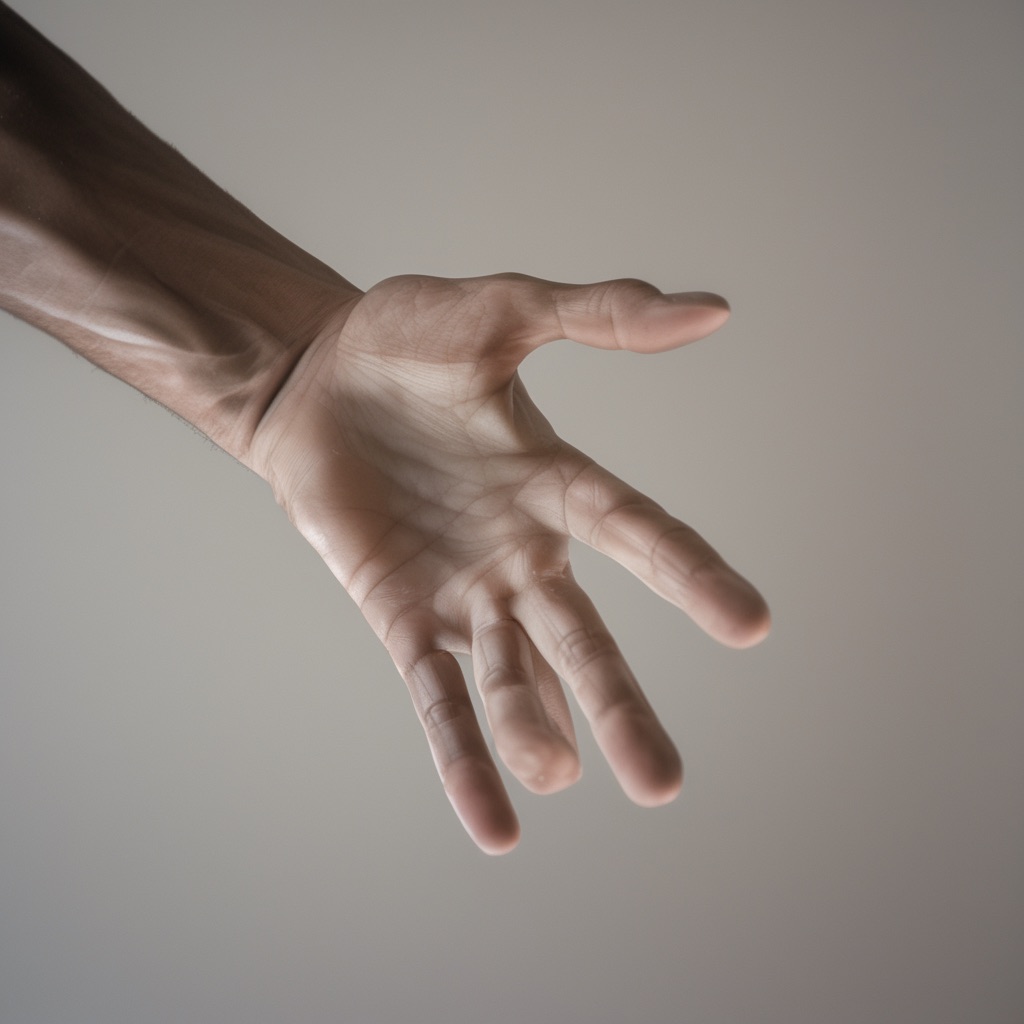} \\
        \multicolumn{4}{l}{\begin{tabular}{l}SDXL $\times$ Pick-a-pic Prompt: ``a closeup photo of a human hand''\end{tabular}} \\
        
    \end{tabular}
    }
    \caption{Qualitative visual examples. DM and prompt details provided. Best viewed in color. We provide further examples in the supplementary materials.} %
    \label{fig:mainbody_examples}
    \vspace{-4mm}
\end{figure}

\noindent\textbf{Qualitative Image Comparison.} We %
consider a qualitative visual comparison. Figure~\ref{fig:mainbody_examples} provides some %
examples and we report %
additional results, including on CivitAI fine-tunes and Low-Rank Adapters~\cite{dettmers2024qlora} %
in the supplementary. Here, we have a clearer picture of exactly what each method is doing. Specifically, only \np\ and NoiseAR will substantially change the result from the stochastic baseline, while Golden Noise only makes small edits within the confines of the initial noise. For example, the `hypnotoad from Futurama' does not wear a suit, but does have reptilian eyes. While our method and NoiseAR remove the suit and maintain the proper eye shape, Golden Noise instead preserves the suit and erroneously gives the hypnotoad human-like eyes. 

Further, consider the anime image of `milim', where the standard baseline noise produces artifacts on the character's cheek which Golden Noise does not correct for. Finally, one on-going challenge for DMs, especially older ones like SDXL, has been accurately representing human anatomy~\cite{wang2024towards}, especially hands~\cite{ma2025evaluating, narasimhaswamy2024handiffuser, kamali2025characterizing}. Despite this hurdle, \np\ is able to perform initial noise optimization to sufficiently provide a fairly reasonable and realistic image with no extra fingers and minimal distortions.

\subsection{Prompt-Aware DM Capability Estimation}
\label{sec:eval_prompt}
\begin{table}[t!]
    \centering
    \caption{Prompt-only prediction results on the test set for all four DMs. We enumerate the distribution of Pick-a-Pic prompt scores: mean, standard deviation and range %
    of the per-prompt mean scores $\mu_{S_p}$. We report predictor performance when \np\ predicts $\mu_{S_p}$ for a given prompt using only the prompt embedding.}
    \label{tab:prompt_score_result}
    \scalebox{\scaleboxratio}{
    \begin{tabular}{c|c|c|ccc} \toprule
    \textbf{Model} & \textbf{Score Distribution} & \textbf{Score Range} & \textbf{SRCC~(↑)} & \textbf{MAE~(↓)} & \textbf{MAPE~(↓)} \\ \midrule
    PixArt-$\Sigma$ & $\mathcal{N}(21.75, 1.48)$ & $[17.35, 26.01]$ & 0.6922 & 0.8920 & 3.98\% \\
    Hunyuan      & $\mathcal{N}(21.72, 1.44)$ & $[17.40, 26.73]$ & 0.7409 & 0.7818 & 3.68\% \\
    DreamShaper  & $\mathcal{N}(22.05, 1.40)$ & $[17.38, 26.51]$ & 0.7258 & 0.7287 & 3.33\% \\
    SDXL         & $\mathcal{N}(21.76, 1.31)$ & $[18.85, 26.09]$ & 0.7371 & 0.6418 & 3.05\% \\
    \bottomrule
    \end{tabular}
    }
\end{table}

We now evaluate the ability %
to provide accurate feedback on how well a DM can generate images for a given prompt. Following Section~\ref{sec:method_prompt_estimation}, this is done by masking the noise encoder $\Phi_{noise}$ and only rely on $\Phi_{prompt}$ to predict $\mu_{S_p}$ instead of $S_{p, I}$. Specifically, we use a special 
dataloader where all images in a batch share the %
prompt encoding, 
but differ in noise and score. %
From the batch we compute a single ground-truth target as the average of all single score targets $\mu_{S_p}$, and perform one inference pass on just the prompt encoding which is consistent throughout the batch. We perform this task on our pretrained predictors for each DM using their specific prompt-noise test %
set.

Table~\ref{tab:prompt_score_result} reports our findings in terms of one ranking metric, SRCC, and two regression metrics, MAE and MAPE. 
\np\ is able to provide competent feedback to an end-user on DM generation capacity for a given prompt. Specifically, we note that our method achieves less than 4\% MAPE on all four DMs and achieves over 0.7 SRCC on three of four DMs. Additionally, we can infer the suitability of each DM on the Pick-a-Pic prompts from the distribution of true scores $S_{p, I}$. Generally, SDXL, a popular academic benchmark, is a safe bet as it has a higher minimum (i.e., only minimum above 18) than the other DMs, yet falls short of Hunyuan in terms of top-end performance.

\begin{table}[t!]
    \centering
    \caption{Hardware metrics comparing \np\ to Golden Noise. We measure the parameter count, FLOPs~\cite{mills2023aiop}, latency on two devices and checkpoint (Ckpt.) size of the \np\ predictor compared to Golden Noise's NPNet.}
    \label{tab:hardware}
    \scalebox{\scaleboxratio}{
    \begin{tabular}{ccccccc} \toprule
    \textbf{Model} & \textbf{Method} & \textbf{Params~(↓)} & \textbf{FLOPs~(↓)} & \textbf{RTX PRO5000~(↓)} & \textbf{DGX Spark~(↓)} & \textbf{Ckpt.~(↓)} \\ \midrule
    \multirow{2}{*}{Hunyuan} & Golden Noise & 29.84M & 9.29G & 8.61ms (1.0x) & 16.44ms (1.0x) & 119.40MB \\
                             & \np          & 49.92M & 9.57G & 1.75ms (4.9x) & 4.71ms (3.5x) & 101.26MB \\ \midrule
    \multirow{2}{*}{SDXL/D.S.} & Golden Noise & 30.47M & 9.29G & 8.45ms (1.0x) & 25.05ms (1.0x) & 122.01MB \\
                               & \np        & 49.92M & 6.57G & 1.54ms (8.2x) & 3.24ms (7.7x) & 100.21MB \\
    \bottomrule
    \end{tabular}
    }
    \vspace{-4mm}
\end{table}

\subsection{Hardware Considerations}
\begin{wraptable}{rt}{0.43\textwidth}
    \centering
    \vspace{-10mm}
    \caption{Latency for Golden Noise and \np\ ($N$=100) to generate $|B|=10$ optimal noises on an Nvidia RTX 6000 Ada or DGX Spark for SDXL and DreamShaper.}
    \label{tab:latency}
    \scalebox{\scaleboxratio}{
    \begin{tabular}{ccc} \toprule
    \textbf{Device} & \textbf{Method} & \textbf{Latency~(↓)} \\ \midrule
    \multirow{2}{*}{RTX6000} & Golden Noise & 64.46ms \\
                             & \np          & 44.47ms \\ \midrule
    \multirow{2}{*}{DGX Spark} & Golden Noise & 111.97ms \\
                             & \np          & 67.25ms \\
    \bottomrule
    \end{tabular}
    }
    \vspace{-5mm}
\end{wraptable}

Finally, we consider the hardware costs of \np\ as part of a broader, deployed DM pipeline. Due to space constraints we primarily compare to Golden Noise as the most apples-to-apples method.

Table~\ref{tab:hardware} provides hardware measurements for a single inference pass of \np\ compared to Golden Noise's NPNet. Specifically, our PAINE predictor consumes more parameters %
but leverages them towards stronger prompt encoders $\Phi_{prompt}$ and is able to perform much faster inference. %

However, all measurements in Table~\ref{tab:hardware} consider a single batch size where \np, unlike Golden Noise, is explicitly a sampling approach where $N>>|B|$. Thus, we provide a more realistic test in Table~\ref{tab:latency} where we consider the time it takes to perform initial noise optimization for $|B|=10$ images on SDXL, %
one of the most popular DMs for fine-tuning~\cite{civitaiJuggernautRagnarok_by_RunDiffusion, civitaiIllustriousV20, lykon2023dreamshaper, civitaiCyberRealistic_Pony, hassaku_civit}. These results show that \np\ has a lower latency footprint than Golden Noise.

\subsection{Ablation Studies}
\label{sec:ablation}

\begin{figure}[t]
  \centering
  \begin{minipage}{0.2\linewidth}\centering
    \includegraphics[width=\linewidth]{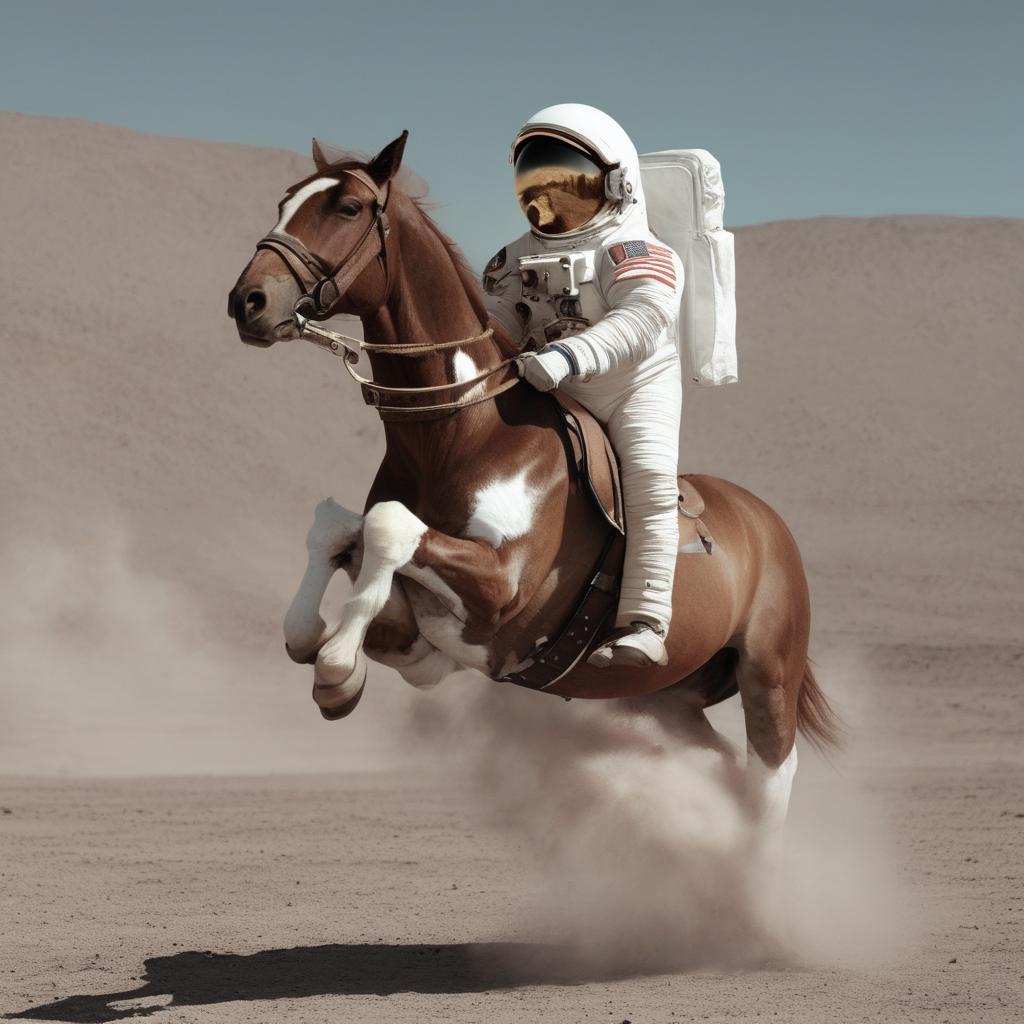}\\ \textbf{(a)} $N{=}20$
  \end{minipage}\hfill
  \begin{minipage}{0.2\linewidth}\centering
    \includegraphics[width=\linewidth]{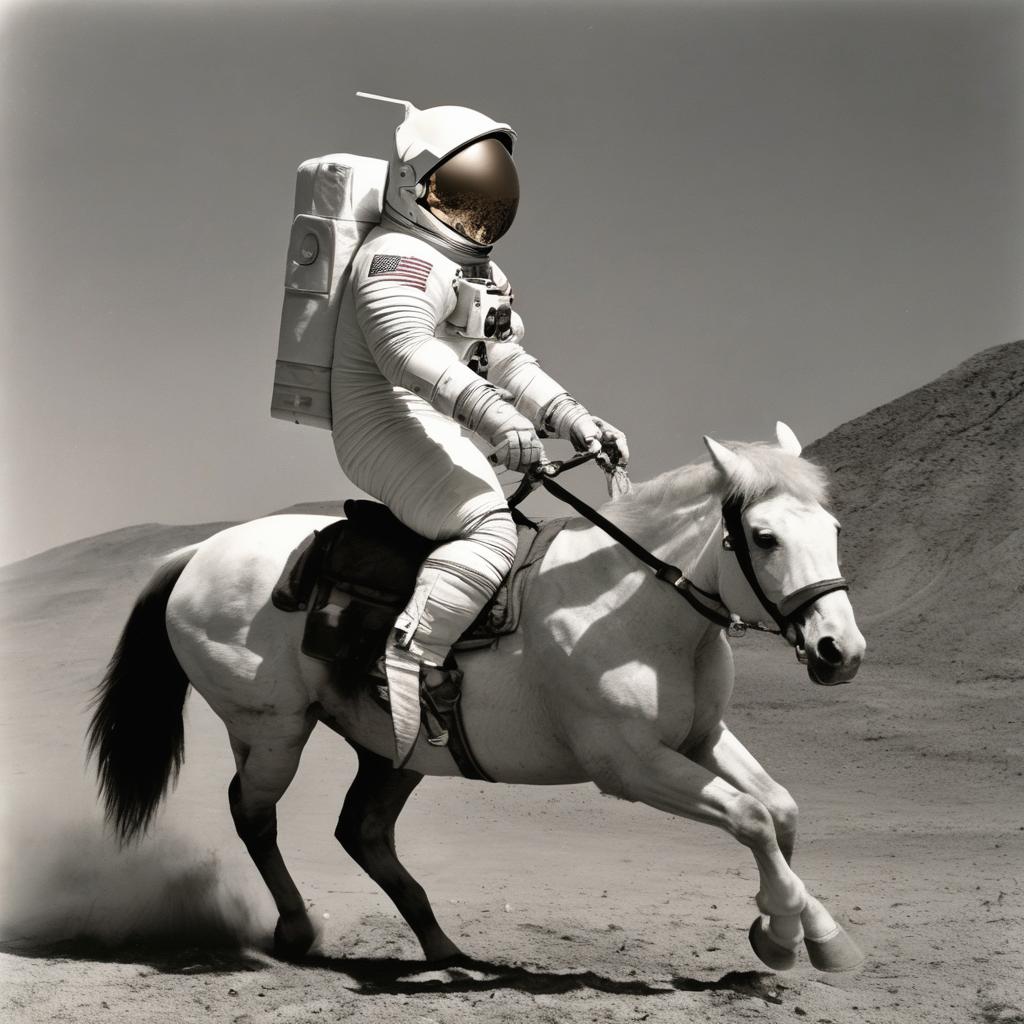}\\ \textbf{(b)} $N{=}80$
  \end{minipage}\hfill
  \begin{minipage}{0.2\linewidth}\centering
    \includegraphics[width=\linewidth]{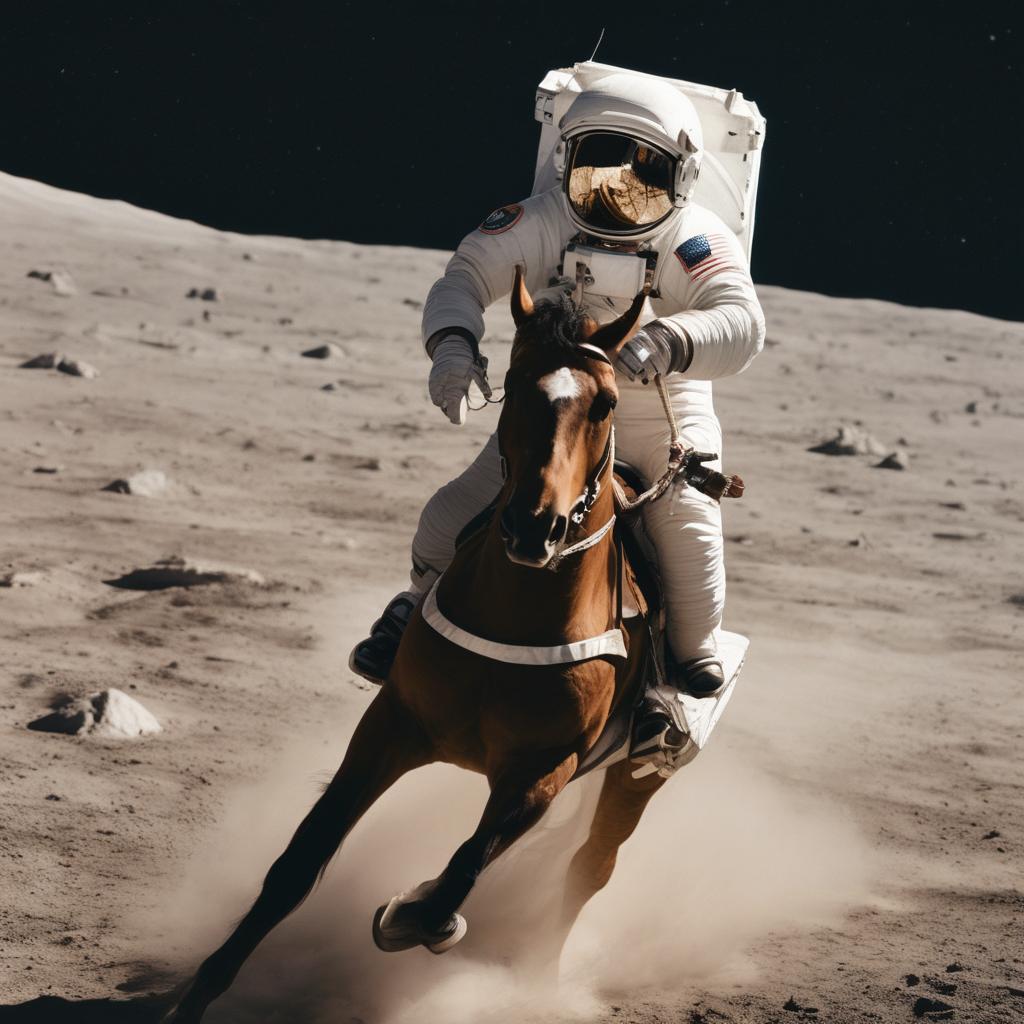}\\ \textbf{(c)} $N{=}320$
  \end{minipage}\hfill
  \begin{minipage}{0.2\linewidth}\centering
    \includegraphics[width=\linewidth]{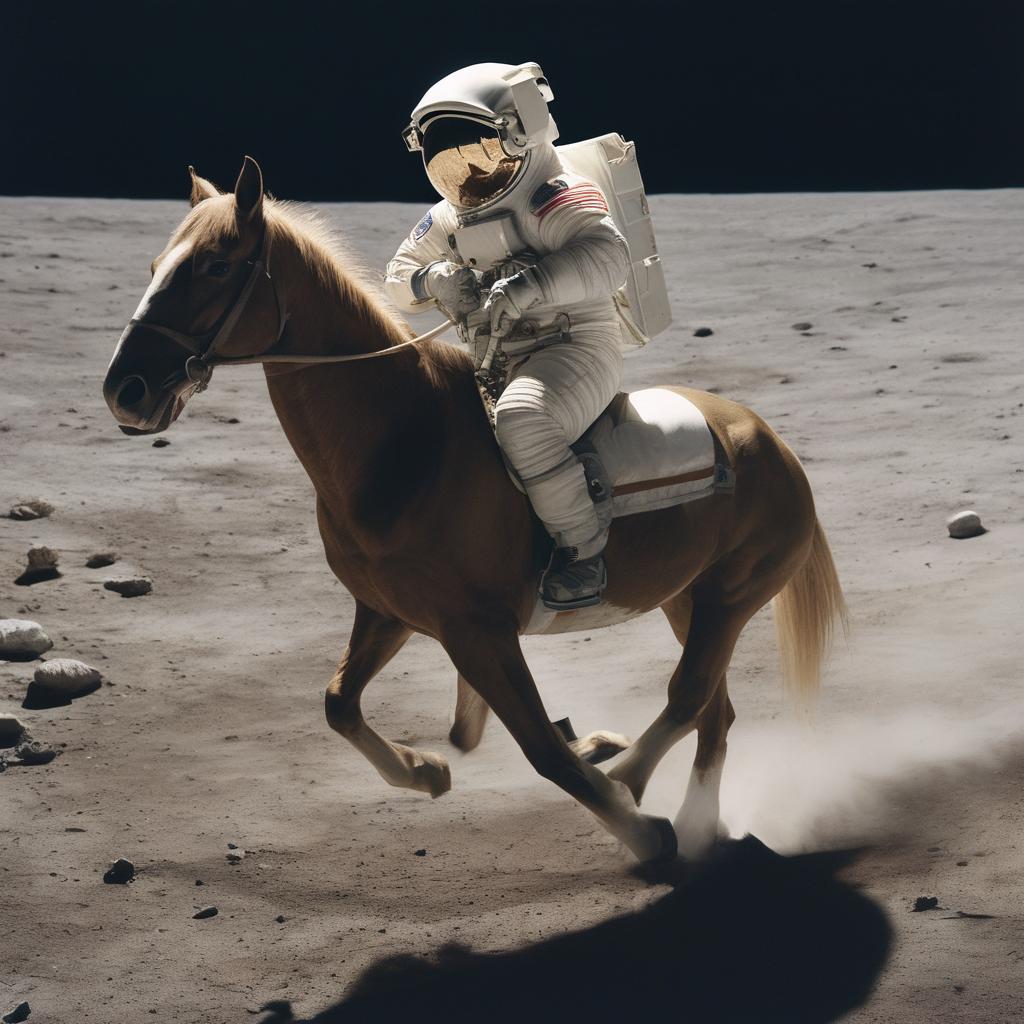}\\ \textbf{(d)} $N{=}1280$
  \end{minipage}
  \caption{Effect of %
  $N$ on \np\ selection for `a photograph
  of an astronaut riding a horse'. Larger $N$ yields better-composed selections
  with diminishing returns.}
  \label{fig:n_astronaut}
  \vspace{-6mm}
\end{figure}

\begin{wrapfigure}{rt}{0.55\linewidth}
    \centering
    \vspace{-6mm}
    \includegraphics[width=\linewidth]{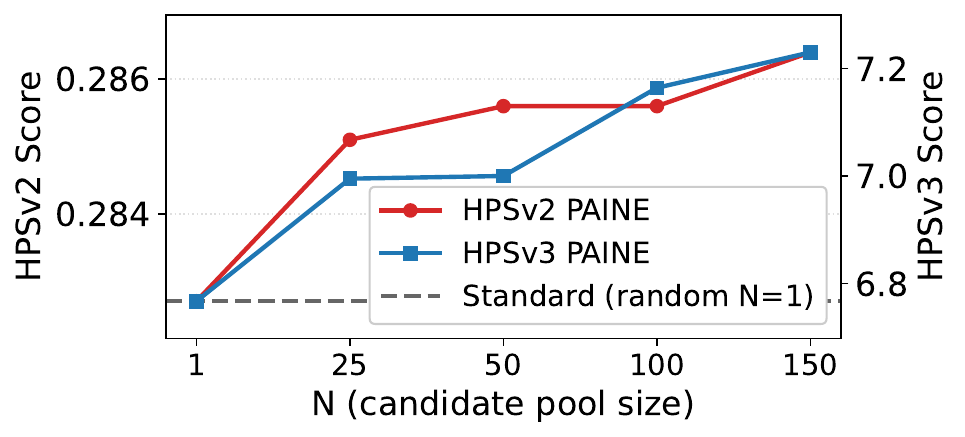}
    \caption{$N$ vs. mean $S_{p, I}$ for \np\ on SDXL for 1k random prompts for HPSv2/v3~\cite{wu2023human, ma2025hpsv3}.}
    \label{fig:n_sweep}
    \vspace{-6mm}
\end{wrapfigure}
We ablate the effect of $N$ here and %
provide additional ablations in the supplementary.

Figure~\ref{fig:n_astronaut} illustrates the corresponding qualitative effect for the prompt `a photograph of an astronaut riding a horse': as $N$ grows from 20 to 1280, \np\ selects increasingly well-composed %
samples, though with diminishing returns as $N$ increases. %
Further, %
Figure~\ref{fig:n_sweep} illustrates the effect of increasing %
$N$ compared to HPSv2/v3 scores on SDXL. Intuitively, the graph shows a monotonic increase: Higher $N$ leads to equal or greater preference scores. 

\section{Conclusions}%
\label{sec:conclusion}

We propose \np, a simple and lightweight method for initial noise optimization on Text-to-Image (T2I) Diffusion Models (DM). Specifically, \np\ achieves superior generative quality whilst also providing feedback on DM generative ability for a user-provided prompt. Specifically, our method leverages advances in estimating human preferences from prompt-image pairs to directly estimate a human preference score using %
prompt encodings and stochastically-sampled initial noises. \np\ achieves superior qualitative and quantitative results compared to existing fine-tuning-free approaches whilst %
incurring a smaller hardware latency if integrated into an existing DM pipeline.

\bibliographystyle{splncs04}
\bibliography{main}

\newpage
\appendix

\section*{Supplementary Materials}

We provide additional information and materials to support the main manuscript. Specifically, we provide additional experimental setup and execution details, extensive quantitative evaluation results for NoiseAR~\cite{li2025noisear} and GenEval~\cite{ghosh2023geneval}, additional results comparing to the RLHF-driven \cite{eyring2025noise} on the few-step SANA-Sprint~\cite{chen2025sana}, further ablation studies and %
additional qualitative image samples.  

\section{Additional Implementation Details}
\label{sec:supp_impl}

We provide further details on the implementation of \np. Further, we discuss how we conduct experiments with baseline methods. 

\subsection{Training Details}
\label{sec:supp_training}

\noindent\textbf{Dataset.} We train one Naïve PAINE predictor per DM. %
For each DM we sample 5k prompts from the Pick-a-Pic~\cite{kirstain2023pick} training set and generate 20 images per prompt, each seeded with an independent Gaussian noise tensor, yielding 100k $(p, X_T, S_{s, I})$ tuples per DM.  All preference scores are computed before training begins. Prompt embeddings are pre-encoded with the DM's native text encoder and cached as \texttt{.pt} files; noise tensors are likewise stored on disk. We split data by prompt into 80\%/10\%/10\%  into distinct training, validation, and test sets. %
Target scores are z-score normalized $z=\sfrac{(x-\mu)}{\sigma}$ using the training-split mean and standard
deviation, which we store %
alongside the checkpoint for inference-time de-normalization.

\noindent\textbf{Grouped Batching.} We implement %
a custom \texttt{PromptGroupedBatchSampler} that groups $k = 6$ prompts
per batch. %
Each prompt has ${\sim}20$ noise samples and each batch contains
${\sim}120$ samples spanning 6 prompts. %

\noindent\textbf{Loss function.} We implement a mixed regression/ranking loss function: 
\begin{equation}
\mathcal{L} = \mathcal{L}_{\mathrm{MAE}} + %
\mathcal{L}_{\mathrm{SRCC}},
\end{equation}
where $\mathcal{L}_{\mathrm{MAE}}$ is the mean absolute error between the predicted
and ground-truth (normalized) scores.
$\mathcal{L}_{\mathrm{SRCC}}$ is the differentiable Spearman Rank Correlation
Coefficient loss from Blondel~\etal~\cite{blondel2020fast}, computed
\emph{per prompt group} and averaged over groups within the batch
(regularization strength $= 10^{-2}$). Following \cite{mills2024autobuild, mills2025qua2sedimo}, %
the ranking SRCC loss encourages the model to rank noises correctly within each prompt
rather than minimizing global regression error.

\noindent\textbf{Optimizer and scheduler.}
We use AdamW~\cite{loshchilov19AdamW} with learning rate $10^{-4}$ and weight decay $10^{-8}$.  Gradients are clipped to a maximum $\ell_2$-norm of 1.0.  The learning rate is halved (\texttt{factor=0.5}) if validation SRCC does not improve for 5 consecutive epochs(\texttt{ReduceLROnPlateau}). We train for 50 epochs and select the checkpoint with the highest validation SRCC as the final model.

\noindent\textbf{Evaluation metrics.}
During training we monitor two %
metrics: %
\begin{itemize}
\item \textbf{SRCC}: global Spearman rank correlation between predicted and
      ground-truth scores across all validation samples, 
      computed \emph{per prompt} and macro-averaged.  Targets are min-max
      scaled to $[0, 5]$ before gain computation (exponential gain,
      \texttt{exp2}).
\item \textbf{MAE}: mean absolute error on raw (de-normalized) score values.
\end{itemize}

\subsection{Baseline Implementation}
\label{sec:supp_baselines}

\noindent\textbf{Standard baseline.} We simply use the vanilla DM pipeline from HuggingFace Diffusers~\cite{von-platen-etal-2022-diffusers} without any initial noise optimization. 

\noindent\textbf{Golden Noise}~\cite{zhou2025golden}. We use the official open-source NPNet checkpoints. %
For PixArt-$\Sigma$, which borrows the $X_T \in \mathbb{R}^{4 \times 128 \times 128}$ for $1024^2$ image latent space from SDXL, but has no provided checkpoint, we 
apply the SDXL weights and text encoder as a proxy. %

\noindent\textbf{NoiseAR}~\cite{li2025noisear}. We use the open-source %
pretrained weights. The checkpoint for both SDXL and DreamShaper %
is shared across both DMs; for Hunyuan-DiT the DiT-specific checkpoint is used. For PixArt-$\Sigma$ we reuse the SDXL weights. %
Image generation follows the same settings as Golden Noise above.

\section{Extended Quantitative Results}
\label{sec:supp_quant}

We substantiate our background preliminary results on the impact of a prompt on preference score distribution. We also expand 
upon the quantitative results comparing \np\ to contemporary baseline approaches on a variety of benchmarks. %

\begin{figure}[H]
  \centering

  \begin{minipage}{\linewidth}
      \centering
      \includegraphics[width=\linewidth]{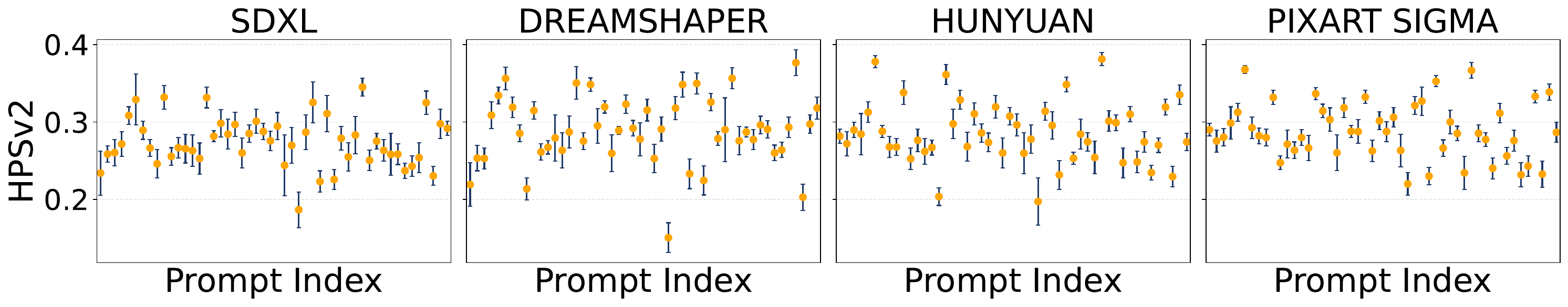}
      \textbf{(a)} HPSv2~\cite{wu2023human}\\[2mm]
  \end{minipage}

  \vspace{2mm}

  \begin{minipage}{\linewidth}
      \centering
      \includegraphics[width=\linewidth]{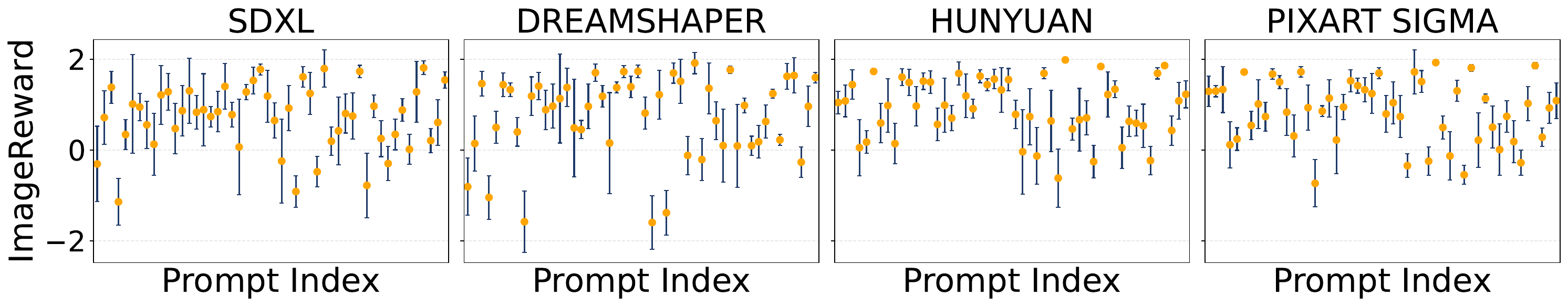}
      \textbf{(b)} ImageReward~\cite{xu2023imagereward}\\[2mm]
  \end{minipage}

  \vspace{2mm}

  \begin{minipage}{\linewidth}
      \centering
      \includegraphics[width=\linewidth]{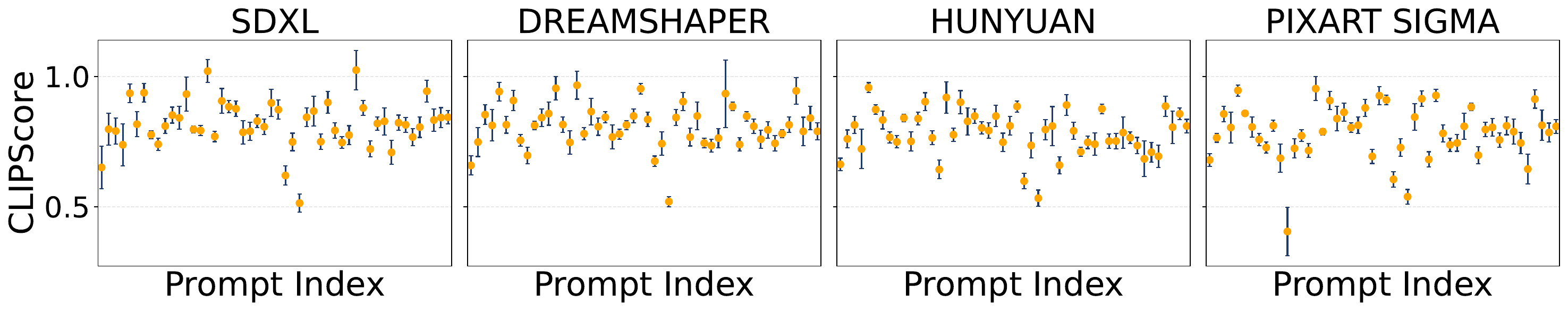}
      \textbf{(c)} CLIPScore~\cite{hessel2021clipscore}\\[2mm]
  \end{minipage}

  \caption{Extending Figure~\ref{fig:score_dists} to different human preference/prompt adherence metrics.}
  \label{fig:score_dists_appendix}
  \vspace{-5mm}
\end{figure}

\noindent\textbf{Extended prompt score distributions.} 
Figure~\ref{fig:score_dists_appendix} extends Figure~\ref{fig:score_dists} across different human preference metrics for the same DMs. Specifically, we provide additional results on HPSv2~\cite{wu2023human}, ImageReward~\cite{xu2023imagereward} and CLIPScore~\cite{hessel2021clipscore}. As we can see, the distribution of scores by mean $\mu_{S_{p, I}}$ and standard deviation $\sigma_{S_{p, I}}$ primarily depend on the prompt $p$ across different DMs and preference metrics. Further, while some prompts stand out as easier or harder to achieve a high score for than others, by visual inspection the exact distributions vary across DM with no perfect correlation. 

The most interesting empirical finding of Figure~\ref{fig:score_dists_appendix} is how much variance there is within ImageReward. Although the scales of different human preference metrics are not directly comparable, we can visually observe a much larger average variance in score distributions compared to other methods, especially on the U-Net DMs, SDXL and its fine-tune, DreamShaper. %

\noindent\textbf{Additional T2I metrics.} Table~\ref{table:main_exp_supp} provides additional quantitative results on prompt datasets like Pick-a-Pic or DrawBench and various human preference metrics like HPSv2/v3. Specifically, we compare  %
\np\ to NoiseAR~\cite{li2025noisear}, which is a costlier (compared to \np\ and Golden Noise~\cite{zhou2025golden}), RLHF and fine-tuning approach that performs initial noise optimization by modifying the %
DM noise after every timestep in an autoregressive manner. %

\begin{table}[!t]
\begin{center}
\caption{T2I Quantitative Comparison on several human preference benchmarks and prompt datasets. Same setup as Table~\ref{table:main_exp_full}. Best and second best results in \textbf{bold} and \textit{italics}, respectively.}

\scalebox{\scaleboxratio}{
\begin{tabular}{ccccccc}
\toprule
Model & Dataset & Method & HPSv2~(↑) & HPSv3~(↑) & ImageReward~(↑) & PickScore~(↑) \\ \midrule
\multirow{12}{*}{PixArt-$\Sigma$}
&  \multirow{3}{*}{Pick-a-Pic} & Standard     & 29.83 & \textit{8.60} & 81.24 & \textbf{21.90} \\
&                              & NoiseAR      & \textit{29.91} & 8.18 & \textit{84.58} & 21.71 \\ 
&                              & \np          & \textbf{30.82} & \textbf{8.73} & \textbf{86.02} & \textit{21.82} \\ \cline{2-7} 
&  \multirow{3}{*}{HPDv2}      & Standard     & 30.91 & \textit{10.53} & \textbf{106.45} & \textit{22.98} \\
&                              & NoiseAR      & \textit{31.88} & 9.75 & 102.93 & 22.78 \\ 
&                              & \np          & \textbf{32.71} & \textbf{10.58} & \textit{106.11} & \textbf{23.01} \\ \cline{2-7} 
&  \multirow{3}{*}{HPSv2}      & Standard     & \textit{31.70} & \textit{11.93} & \textit{103.21} & \textit{22.68} \\
&                              & NoiseAR      & 30.57 & 11.07 & 96.47 & 22.38 \\ 
&                              & \np          & \textbf{31.79} & \textbf{11.96} & \textbf{103.23} & \textbf{22.70} \\ \cline{2-7} 
&  \multirow{3}{*}{DrawBench}  & Standard     & 28.54 & \textbf{9.60} & \textit{73.41} & \textit{22.33} \\
&                              & NoiseAR      & \textit{28.92} & 8.66 & 60.12 & 22.13 \\ 
&                              & \np          & \textbf{30.03} & \textit{9.57} & \textbf{75.60} & \textbf{22.34} \\ \midrule 
\multirow{12}{*}{Hunyuan}
&  \multirow{3}{*}{Pick-a-Pic} & Standard & 29.48 & 8.06 & 85.47 & 21.81 \\
&                              & NoiseAR & \textit{30.27} & \textbf{8.56} & \textbf{103.7} & \textbf{21.93} \\ 
&                              & \np     & \textbf{30.77} & \textit{8.24} & \textit{89.52} & \textit{21.86} \\ \cline{2-7} 
&  \multirow{3}{*}{HPDv2}      & Standard & 31.04 & 10.01 & 109.27 & 22.88 \\
&                              & NoiseAR & \textit{31.42} & \textbf{10.17} & \textbf{121.17} & \textbf{23.01} \\ 
&                              & \np     & \textbf{32.79} & \textit{10.02} & \textit{114.78} & \textit{23.00} \\ \cline{2-7} 
&  \multirow{3}{*}{HPSv2}      & Standard & 30.06 & 11.42 & 106.41 & \textit{22.58} \\
&                              & NoiseAR & \textit{30.32} & \textbf{11.60} & \textbf{116.05} & \textbf{22.63} \\ 
&                              & \np     & \textbf{31.70} & \textit{11.43} & \textit{108.42} & \textit{22.58} \\ \cline{2-7} 
&  \multirow{3}{*}{DrawBench}  & Standard & 28.49 & \textit{9.18} & 84.45 & 22.31 \\
&                              & NoiseAR & \textit{29.14} & \textbf{9.53} & \textit{89.53} & \textbf{22.48} \\ 
&                              & \np     & \textbf{30.16} & 9.10 & \textbf{89.88} & \textit{22.33}\\ \midrule 

\multirow{12}{*}{DreamShaper}
&  \multirow{3}{*}{Pick-a-Pic} & Standard     & 32.51 & \textit{10.33} & 98.06 & 22.60 \\
&                              & NoiseAR      & \textbf{33.64} & \textbf{10.44} & \textbf{111.12} & \textbf{22.71} \\ 
&                              & \np          & \textit{33.27} & 10.29 & \textit{106.60} & \textit{22.64} \\ \cline{2-7} 
&  \multirow{3}{*}{HPDv2}      & Standard     & 33.95 & \textbf{12.38} & \textbf{129.51} & \textbf{23.88} \\
&                              & NoiseAR      & \textit{34.66} & \textit{12.34} & \textit{131.85} & \textit{23.70} \\ 
&                              & \np          & \textbf{34.85} & 12.01 & 119.49 & 23.59\\ \cline{2-7} 
&  \multirow{3}{*}{HPSv2}      & Standard     & 34.39 & 13.90 & 121.91 & 23.42 \\
&                              & NoiseAR      & \textbf{34.65} & \textbf{13.97} & \textbf{126.23} & \textbf{23.48} \\ 
&                              & \np          & \textit{34.44} & \textit{13.94} & \textit{123.24} & \textit{23.44} \\ \cline{2-7} 
&  \multirow{3}{*}{DrawBench}  & Standard     & 30.17 & 11.57 & 96.37 & 22.92 \\
&                              & NoiseAR      & \textit{31.05} & \textbf{11.71} & \textbf{106.56} & \textbf{23.05} \\ 
&                              & \np          & \textbf{31.58} & \textit{11.70} & \textit{97.28} & \textit{22.93} \\ \midrule 
\multirow{12}{*}{SDXL}
&  \multirow{3}{*}{Pick-a-Pic} & Standard     & 28.40 & 7.10 & 57.94 & 21.65 \\
&                              & NoiseAR      & \textbf{29.07} & \textit{7.39} & \textit{75.29} & \textbf{21.85} \\ 
&                              & \np          & \textit{28.85} & \textbf{7.90} & \textbf{80.86} & \textit{21.82} \\ \cline{2-7} 
&  \multirow{3}{*}{HPDv2}      & Standard     & 29.66 & 8.73 & 82.66 & 22.83 \\
&                              & NoiseAR      & \textit{30.77} & \textbf{9.45} & \textbf{108.04} & \textbf{23.05} \\ 
&                              & \np          & \textbf{31.70} & \textit{8.95} & \textit{100.78} & \textit{22.93} \\ \cline{2-7} 
&  \multirow{3}{*}{HPSv2}      & Standard     & \textit{28.69} & \textit{10.24} & 85.06 & 22.52 \\
&                              & NoiseAR      & \textbf{29.10} & \textbf{10.56} & \textbf{96.07} & \textit{22.65} \\ 
&                              & \np          & 28.67 & 10.21 & \textit{92.37} & \textbf{23.01} \\ \cline{2-7} 
&  \multirow{3}{*}{DrawBench}      & Standard     & 26.83 & 7.88 & 60.08 & 22.28 \\
&                              & NoiseAR      & \textit{27.72} & \textbf{8.40} & \textbf{71.31} & \textbf{22.60} \\ 
&                              & \np          & \textbf{28.66} & \textit{8.16} & \textit{66.00} & \textit{22.31} \\ 
\bottomrule
\end{tabular}
}
\label{table:main_exp_supp}
\end{center}
\end{table}

As we can see from Table~\ref{table:main_exp_supp}, \np\ achieves competitive performance on several DMs such as Hunyuan and PixArt-$\Sigma$ especially, across a variety of prompt sets and preference scores like HPSv2 or v3. Specifically, \np\ achieves the best performance in over 22 of 64 test scenarios and the second best performance in over 30 scenarios. Overall, these results further corroborate the efficacy and viability of \np\ as an effective and low-cost method for achieving initial noise optimization. 

\noindent\textbf{Full GenEval comparison.} 
We provide a full categorical breakdown on the GenEval~\cite{ghosh2023geneval} in Table~\ref{table:geneval_full}. To note, GenEval measures semantic/compositional constraints, e.g., `a photo of two vases' or `four dogs standing on a street' rather than human preference, and \np\ is not optimized for it. 
Regardless, we observe that in general, \np\ achieves competent performance in several evaluation categories, specifically single or two object generation, as well as colors and position. Overall, like before, \np\ achieves the second-best performance, only behind NoiseAR on three of four DMs. This finding is impressive considering NoiseAR is a costlier fine-tuning-based method and we did not optimize \np\ for GenEval, but PickScore~\cite{kirstain2023pick} primarily. 

\begin{landscape}
\begin{table}[!t]
\begin{center}
\caption{Full quantitative Performance on GenEval~\cite{ghosh2023geneval} across all categories. Best and second best results in \textbf{bold} and \textit{italics}, respectively.}

\scalebox{\scaleboxratio}{
\begin{tabular}{ccccccccc}
\toprule
Model & Method & Single Obj.~(↑) & Two Obj.~(↑) & Counting~(↑) & Colors~(↑) & Position~(↑) & Color Attrib.~(↑) & Overall~(↑) \\ \midrule
\multirow{4}{*}{PixArt-$\Sigma$}
&  Standard     & \textit{99.06} & 60.86 & 47.19 & \textbf{85.37} & \textbf{15.25} & 25.00 & 55.45 \\
&  Golden Noise & 98.75 & 61.87 & 47.19 & 82.71 & 12.25 & \textbf{28.00} & 55.12 \\
&  NoiseAR      & 98.75 & \textit{63.89} & \textbf{58.44} & 81.12 & 12.75 & 21.75 & \textbf{56.16} \\ 
&  \np          & \textbf{99.38} & \textbf{65.15} & \textit{49.38} & \textit{83.51} & \textit{14.00} & \textit{25.25} & \textit{56.11} \\ \cline{1-9}
\multirow{4}{*}{Hunyuan}
&  Standard     & \textit{97.81} & 82.83 & 61.25 & 88.30 & \textit{17.75} & 46.75 & \textit{65.78} \\
&  Golden Noise & 97.19 & 81.57 & \textit{62.19} & \textbf{89.36} & \textbf{18.00} & \textbf{47.75} & \textbf{66.00} \\
&  NoiseAR      & \textbf{98.44} & \textit{83.03} & \textbf{66.88} & \textbf{89.36} & 12.00 & 44.75 & 65.75 \\ 
&  \np          & 97.19 & \textbf{83.59} & 59.38 & \textit{88.56} & 15.00 & \textit{47.50} & 65.20 \\ \cline{1-9} 
\multirow{4}{*}{DreamShaper}
&  Standard     & \textbf{100} & \textit{82.83} & \textit{44.69} & 84.84 & \textit{13.25} & 30.00 & 59.26 \\
&  Golden Noise & \textbf{100} & \textbf{85.35} & 41.56 & \textbf{85.90} & 11.50 & \textit{30.75} & 59.18 \\
&  NoiseAR      & \textbf{100} & \textbf{85.35} & 40.94 & \textit{85.64} & \textbf{15.50} & \textbf{32.00} & \textbf{59.90} \\ 
&  \np          & \textbf{100} & 80.56 & \textbf{45.31} & \textbf{85.90} & \textit{13.25} & 29.25 & \textit{59.38} \\ \cline{1-9} 
\multirow{4}{*}{SDXL}
&  Standard     & \textit{98.12} & 71.21 & 37.81 & \textbf{83.78} & 9.25 & \textit{21.50} & 53.61 \\
&  Golden Noise & \textbf{98.75} & 67.68 & 37.50 & \textit{82.98} & \textit{13.00} & 21.00 & 53.49 \\
&  NoiseAR      & 96.25 & \textbf{77.78} & \textbf{43.44} & 82.91 & 11.25 & \textbf{27.56} & \textbf{55.08} \\ 
&  \np          & \textit{98.12} & \textit{72.22} & \textit{40.00} & 82.21 & \textbf{18.15} & 19.50 & \textit{54.05} \\ 

\bottomrule
\end{tabular}
}
\label{table:geneval_full}
\end{center}
\end{table}

\end{landscape}

\section{SANA-Sprint}                             \label{sec:supp_sanasprint}

We further compare \np\ to an HyperNoise~\cite{eyring2025noise}. This approach is more akin to NoiseAR~\cite{li2025noisear} in that it is an RLHF and fine-tuning-based initial noise optimization approach, requiring up to 6 H100 GPUs to execute, rather than \np\ and Golden Noise~\cite{zhou2025golden} which are fine-tuning-free and only require a single consumer-grade GPU. However, different from NoiseAR, HyperNoise is tailored towards few-step~\cite{sauer2023adversarial} DMs, some of which utilize radically different latent spaces~\cite{esser2024scaling, xie2024sana} than what we consider in the main paper, so an additional comparison helps to further demonstrate the robustness of our method. 

Specifically, we extend \np\ to SANA-Sprint~\cite{chen2025sana}, a single-step DM whose latent space consists of $32$ channels with higher downsampling, as opposed to prior results which work within the $4$ channel latent space of SDXL. 
\np\ trains a standalone predictor on PickScore as the target metric using the same procedure as the other DMs (Section~\ref{sec:supp_training}).

\begin{table}[t]
\centering
\caption{SANA-Sprint T2I evaluation on four prompt benchmarks.
\np\ is trained on PickScore. Best results in \textbf{bold},
second best in \textit{italics}.}
\label{tab:sana_results}
\scalebox{\scaleboxratio}{
\begin{tabular}{ll cccc}
\toprule
Dataset & Method & HPSv2 ($\uparrow$) & PickScore ($\uparrow$)
& ImageReward ($\uparrow$) & HPSv3 ($\uparrow$) \\
\midrule
\multirow{3}{*}{Pick-a-Pic}
& Standard   & 30.10 & 22.03 & 94.64  & 8.69 \\
& HyperNoise & \textbf{32.31} & 21.92 & \textbf{125.07} & 8.84 \\
& \np        & \textit{30.83} & \textbf{22.00} & \textit{98.61}  & \textbf{8.85} \\
\midrule
\multirow{3}{*}{HPDv2}
& Standard   & 32.03 & \textbf{23.04} & 93.20  & 11.14 \\
& HyperNoise & \textbf{34.25} & \textit{23.05} & \textbf{133.20} & \textbf{11.71} \\
& \np        & \textit{32.23} & 22.96 & \textit{106.36} & \textit{11.27} \\
\midrule
\multirow{3}{*}{HPSv2}
& Standard   & 30.13 & 22.73 & 105.47 & 12.59 \\
& HyperNoise & \textbf{33.38} & \textit{22.74} & \textbf{133.79} & \textbf{13.30} \\
& \np        & \textit{31.53} & \textbf{22.77} & \textit{107.25} & \textit{12.73} \\
\midrule
\multirow{3}{*}{DrawBench}
& Standard   & 28.58 & \textit{22.48} & 90.86  & 10.52 \\
& HyperNoise & \textbf{31.27} & 22.43 & \textbf{113.41} & \textit{11.04} \\
& \np        & \textit{30.07} & \textbf{22.47} & \textit{86.74}  & \textbf{10.92} \\
\bottomrule
\end{tabular}
}
\end{table}

Table~\ref{tab:sana_results} provides the quantitative results comparing \np\ to HyperNoise on a variety of T2I prompt sets and preference metrics. \np\ outperforms HyperNoise on several prompt sets and preference metrics, e.g., Pick-a-Pic and DrawBench, while also providing competitive performance on HPDv2 and HPSv2. Thus, these results demonstrate not only the generalizability of \np\ to generate high-quality artwork on DMs that utilize different latent spaces, but to do so competitively against methods that require much heftier initial upfront compute investment to fine-tune specific DMs. 

\begin{figure}[t!]
  \centering
  \setlength{\tabcolsep}{0pt}          %
  \renewcommand{\arraystretch}{1.2}
  \footnotesize

  \makebox[\linewidth][c]{%
    \begin{tabular}{@{}ccc@{}}
      Standard & HyperNoise & \np \\[2pt]

      \includegraphics[width=0.32\linewidth]{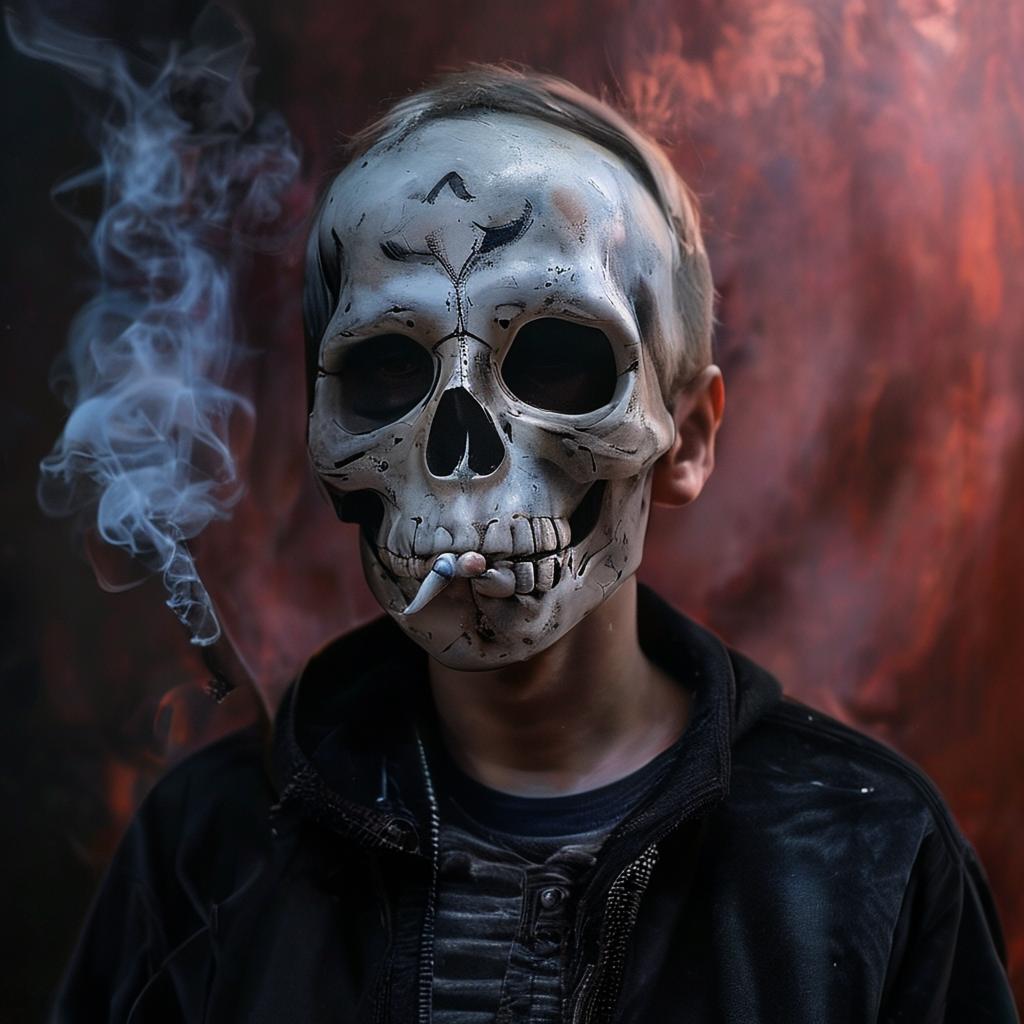} &
      \includegraphics[width=0.32\linewidth]{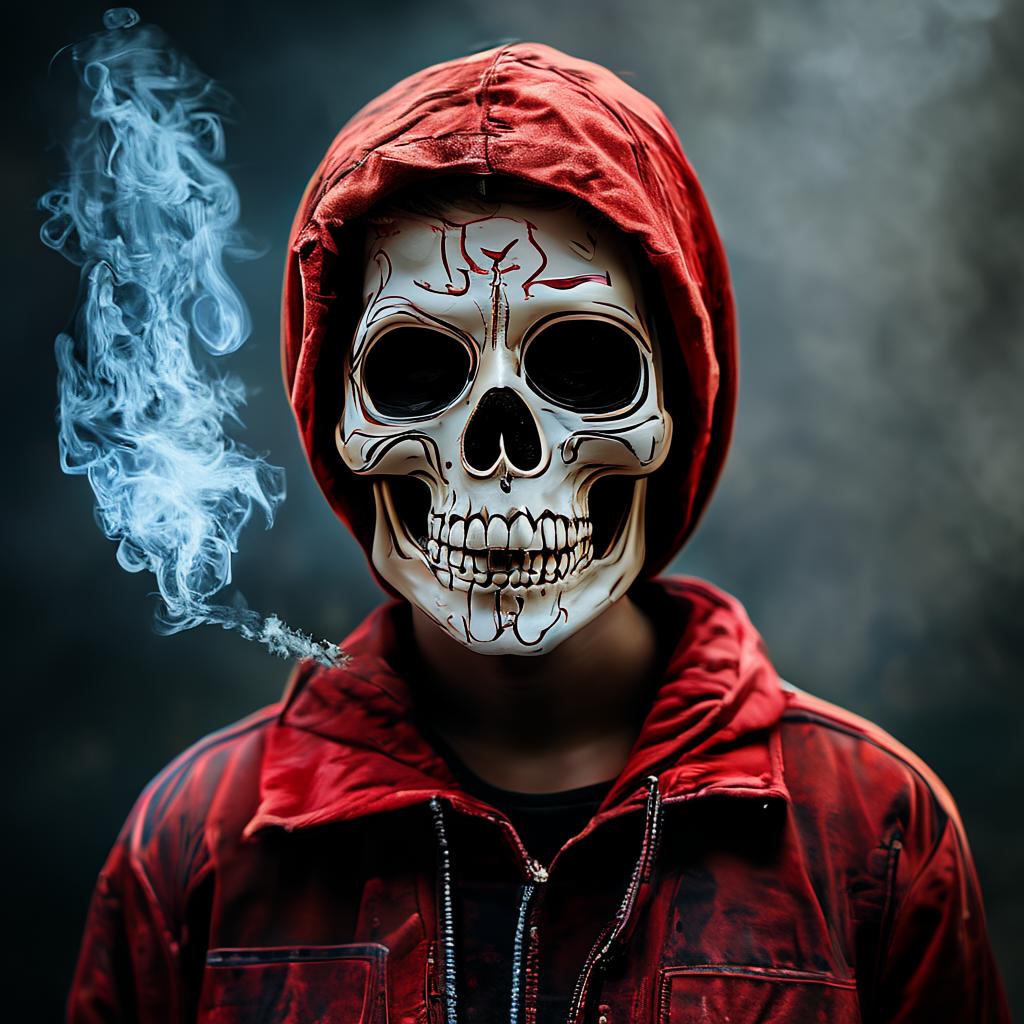} &
      \includegraphics[width=0.32\linewidth]{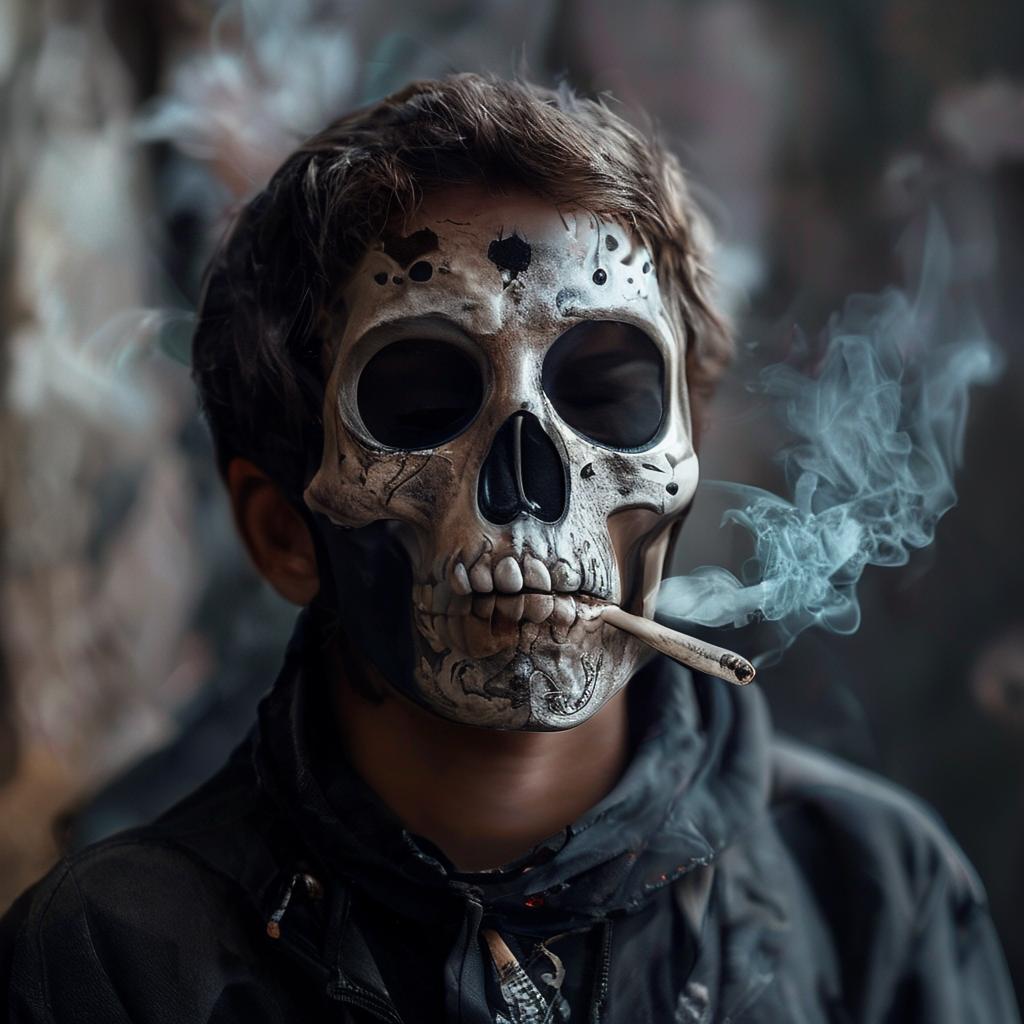} \\
      \multicolumn{3}{@{}l@{}}{\footnotesize ``Teenage boy wearing a skull mask and smoking.''} \\[3pt]

      \includegraphics[width=0.32\linewidth]{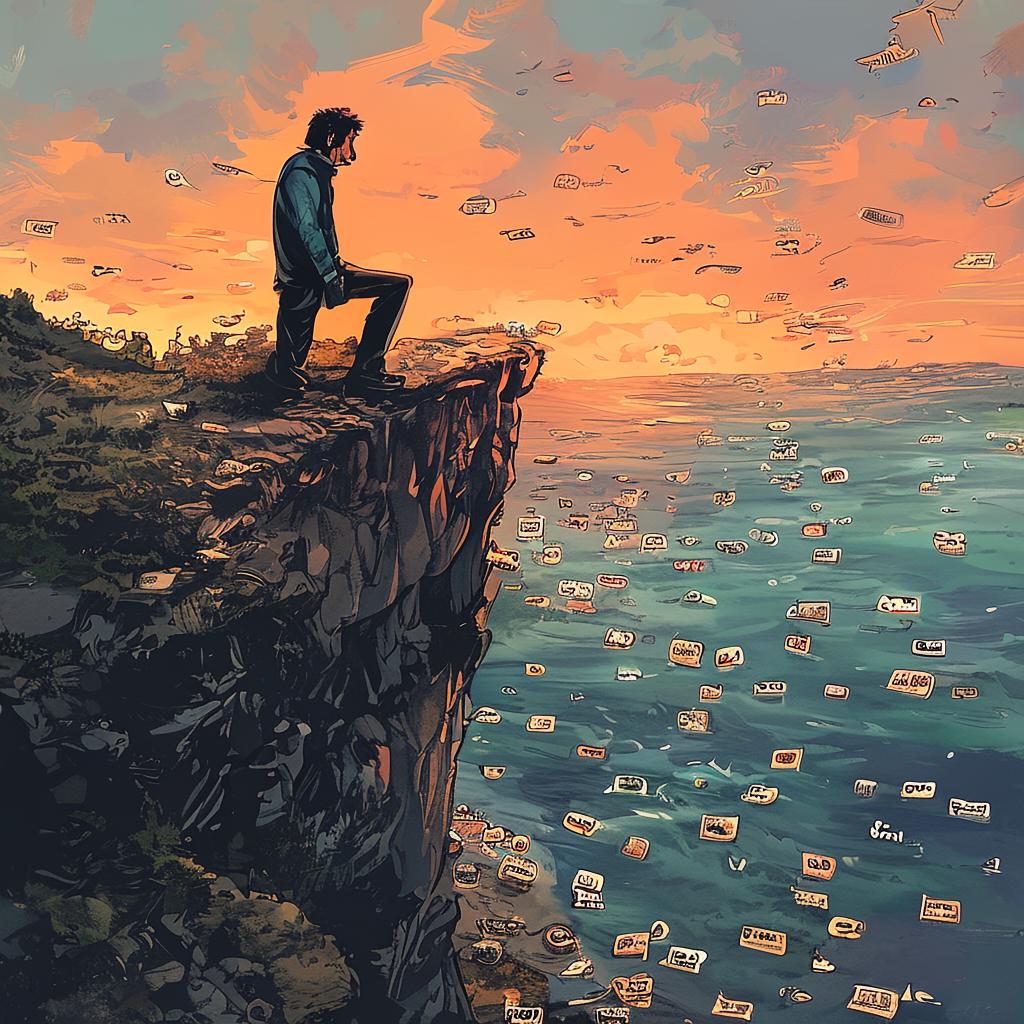} &
      \includegraphics[width=0.32\linewidth]{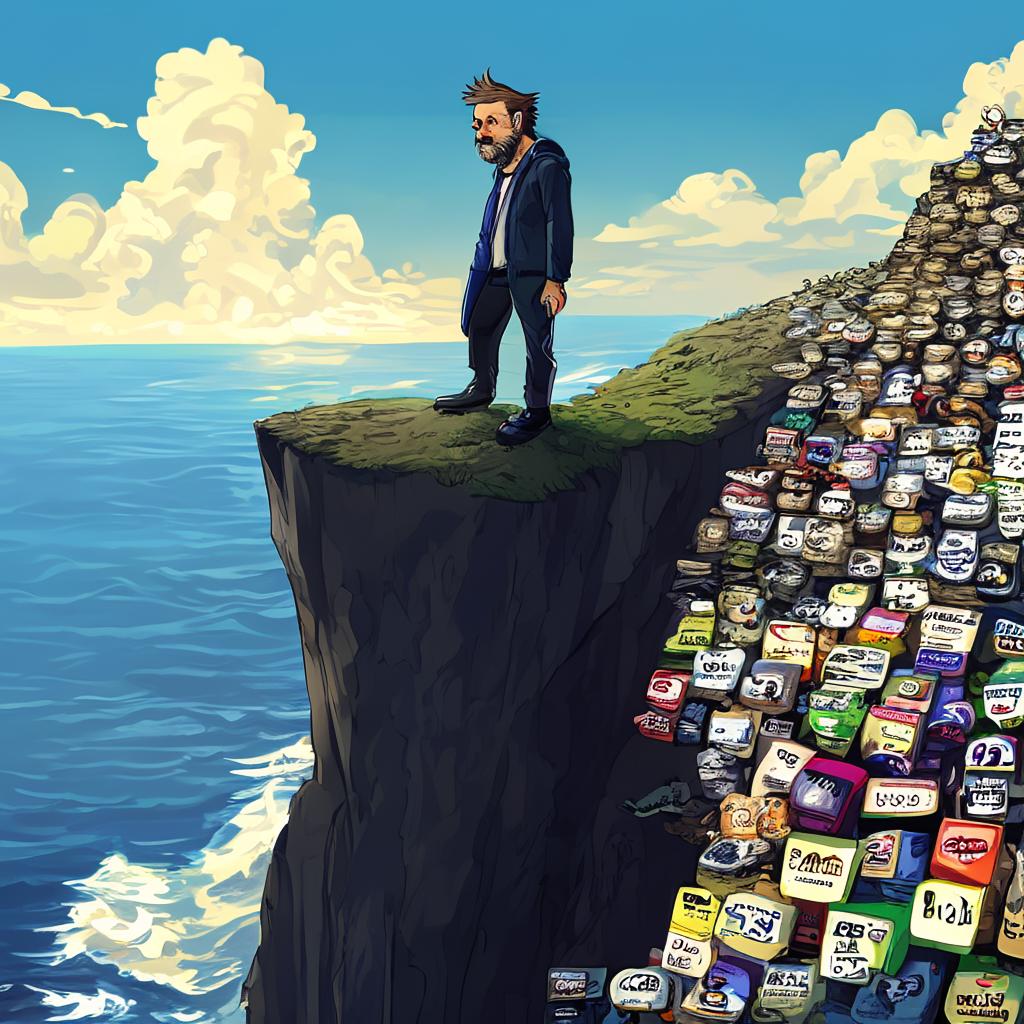} &
      \includegraphics[width=0.32\linewidth]{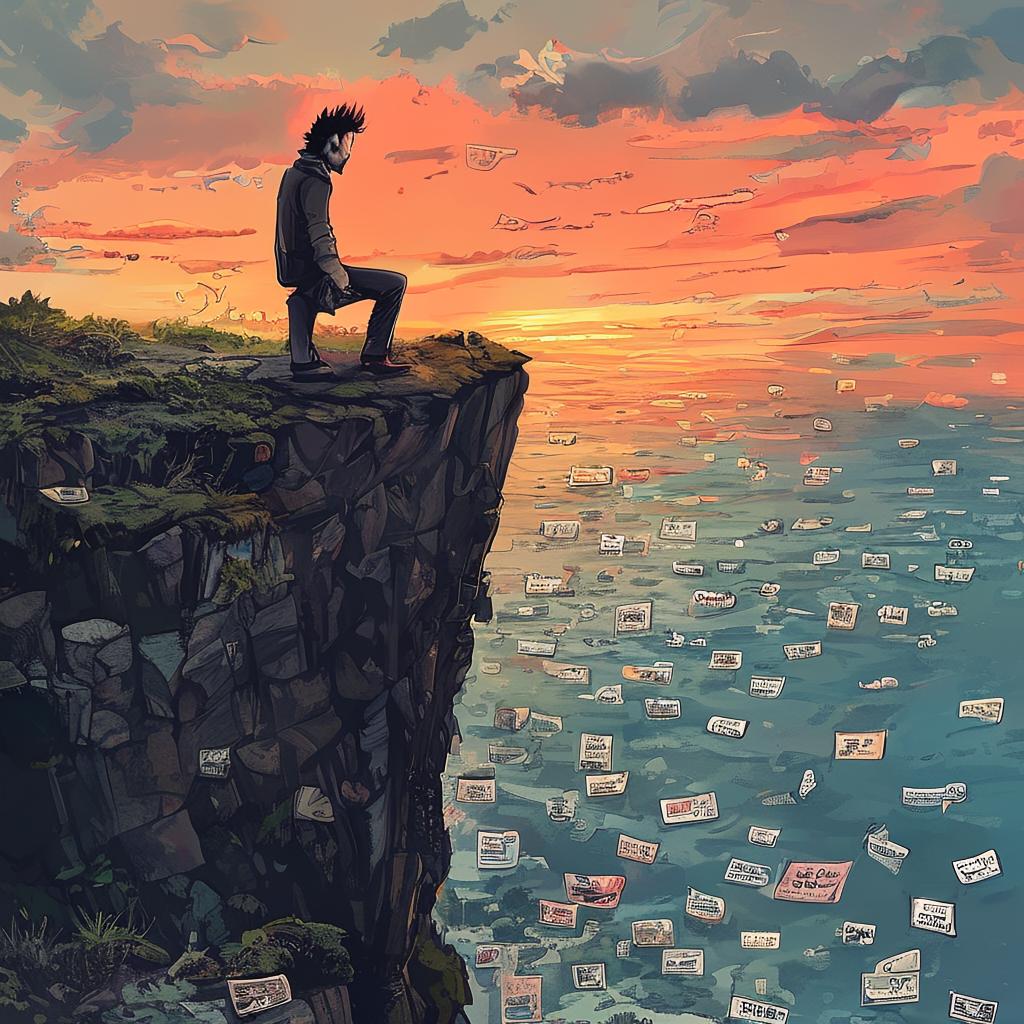} \\
      \multicolumn{3}{@{}l@{}}{\footnotesize ``A Wojack looking over a sea of memes from a cliff on 4chan.''} \\[3pt]

      \includegraphics[width=0.32\linewidth]{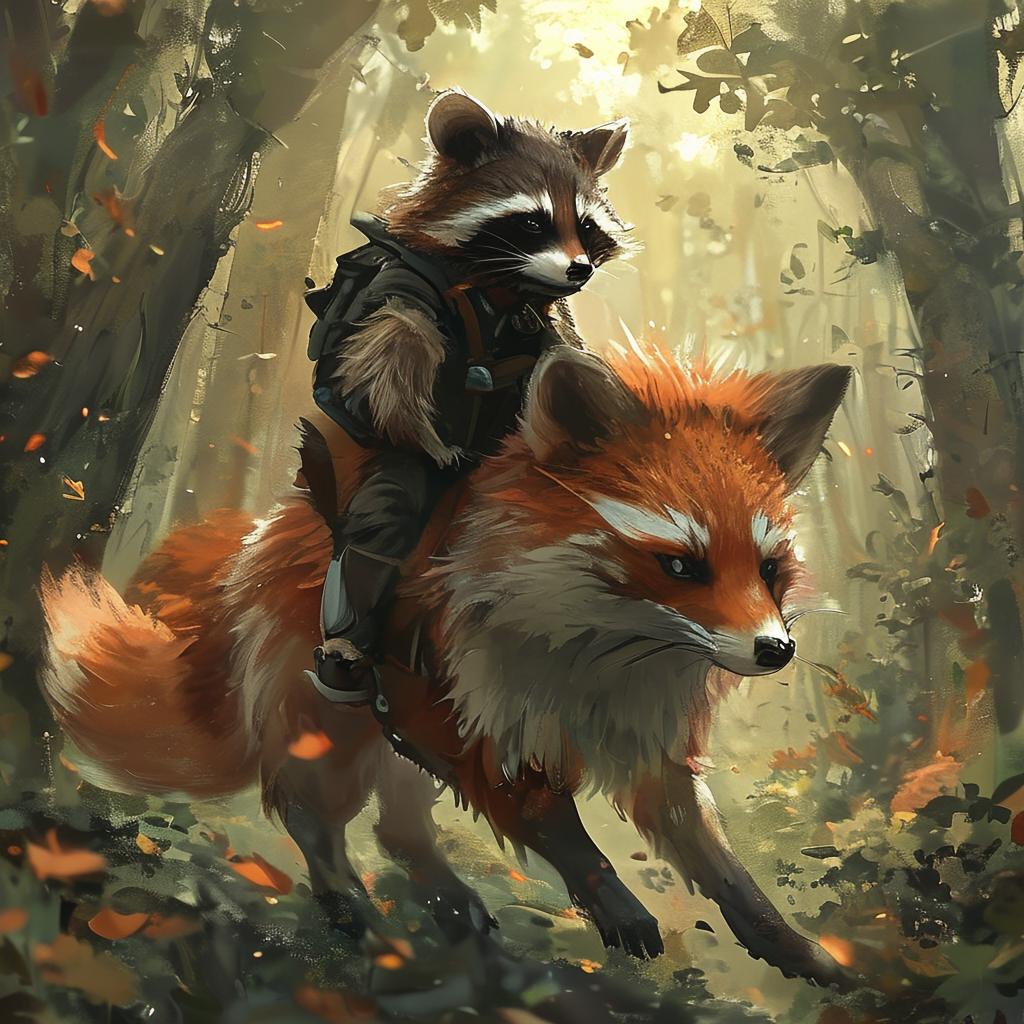} &
      \includegraphics[width=0.32\linewidth]{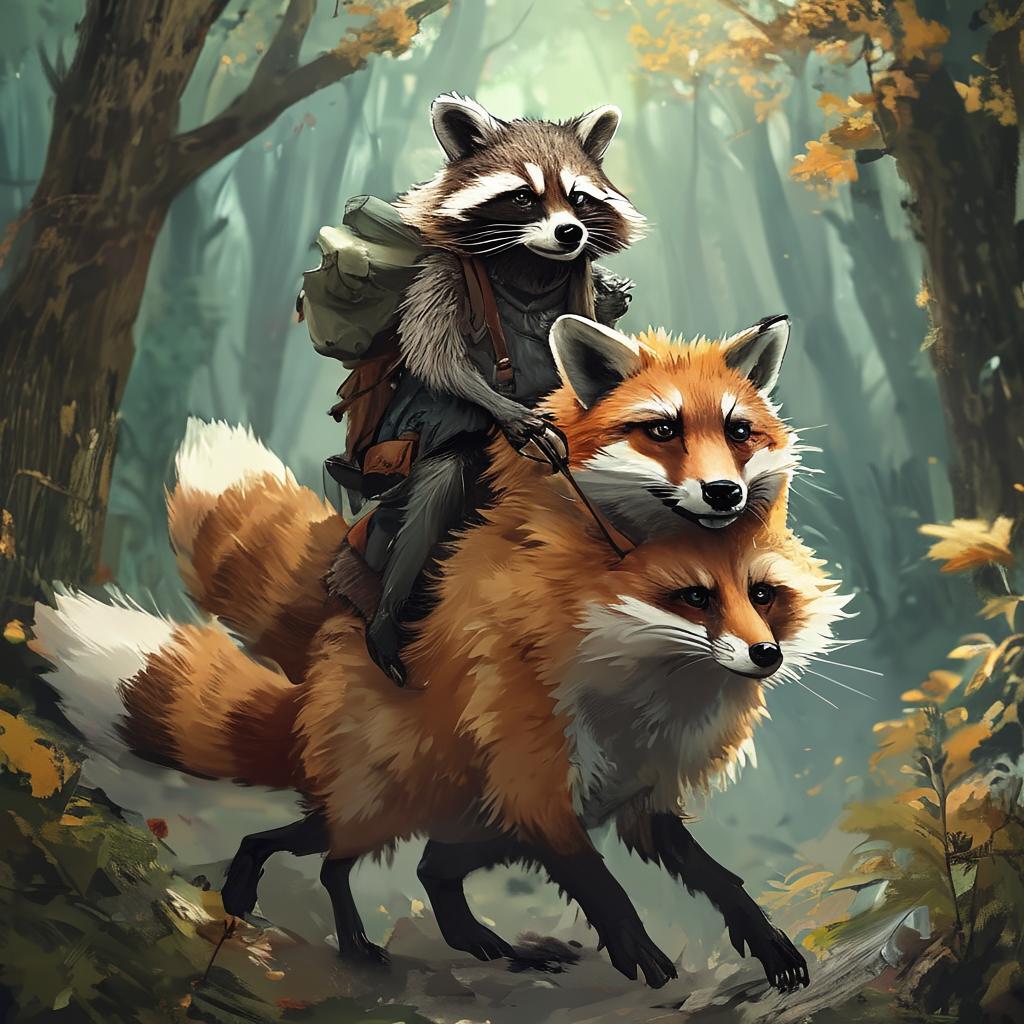} &
      \includegraphics[width=0.32\linewidth]{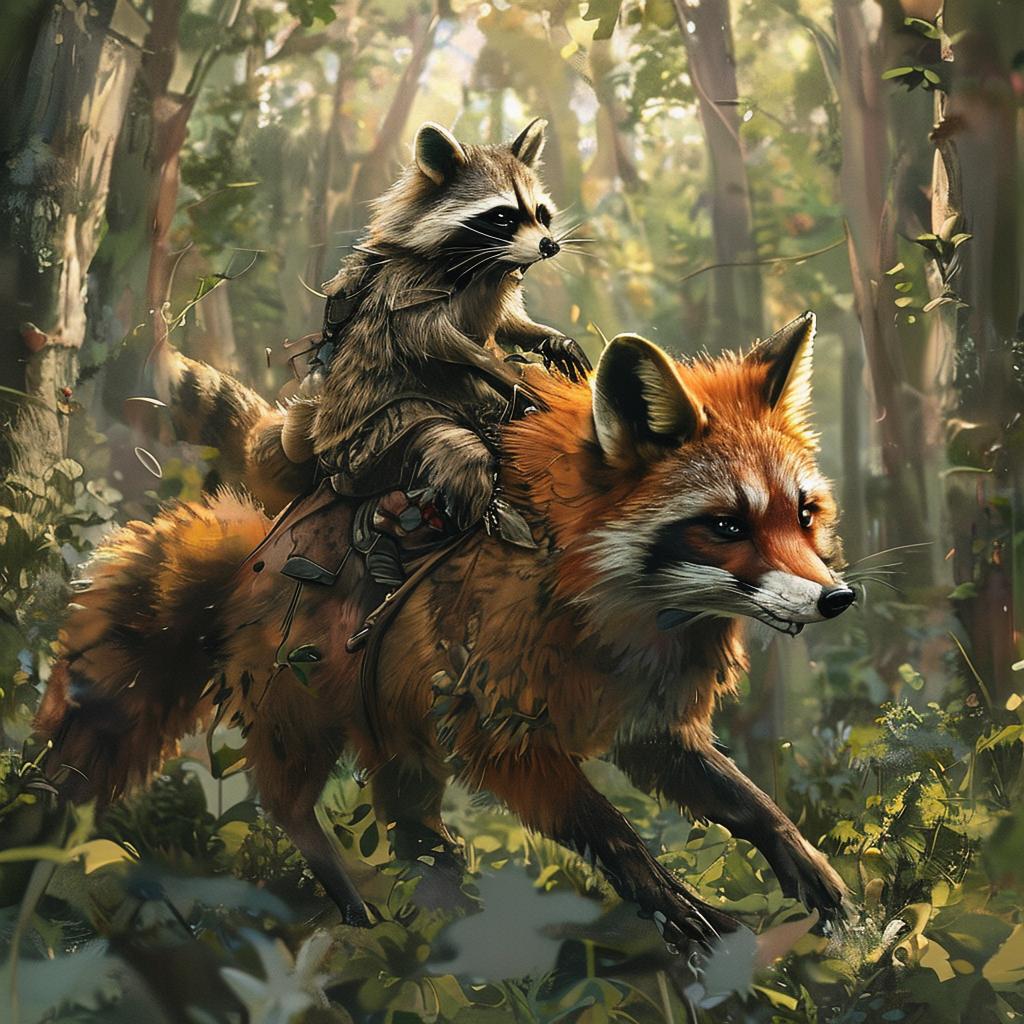} \\
      \multicolumn{3}{@{}l@{}}{\footnotesize ``A racoon riding an oversized fox through a forest in a furry art anime still.''}
    \end{tabular}%
  }%

  \caption{Qualitative results comparing \np\ to HyperNoise on SANA-Sprint. All prompts from HPDv2~\cite{wu2023human}.}
  \label{fig:sana_examples}
  \vspace{-6mm}
\end{figure}

Additionally, Figure~\ref{fig:sana_examples} provides some qualitative examples on SANA-Sprint comparing \np\ to HyperNoise and the standard, optimization-free baseline. These results further illustrate the deployment efficacy and competitiveness of our method. 

\section{Additional Ablation Studies}
\label{sec:supp_ablation}

\begin{table}[t]                                                                                                                                                             
\centering                                                                                                                                                                   
\caption{Ablation studies on SDXL with Pick-a-Pic evaluation benchmark. We select PickScore as the default target, $N{=}100$ candidate noises, and MAE+SRCC as the default loss because this combination yields consistently competitive scores across all four metrics without sacrificing performance on any individual one. Best results \textbf{bolded.}} %
\label{tab:ablation}
\scalebox{\scaleboxratio}{
\begin{tabular}{llcccc}
\toprule
Ablation & Variant & HPSv2~($\uparrow$) & HPSv3~($\uparrow$) & ImageReward~($\uparrow$) & PickScore~($\uparrow$) \\
\midrule
\multirow{3}{*}{\shortstack[l]{Predictor\\Target}}
& HPSv2         & 28.69 & 7.09 & 61.79 & 21.7 \\
& PickScore     & 28.85 & \textbf{7.90} & \textbf{72.00} & \textbf{21.81} \\
& ImageReward   & \textbf{29.85} & 7.28 & 58.26 & 21.77 \\
\midrule
\multirow{2}{*}{\shortstack[l]{Loss\\Function}}
& MAE + SRCC          & 28.85 & \textbf{7.90} & 80.86 & \textbf{21.82} \\
& MAE + LambdaRank    & 28.97 & 7.03 & 80.85 & 21.80 \\
& MAE + LambdaLoss    & \textbf{29.49} & 7.24 & \textbf{81.49} & 21.73 \\
& MAE + neuralNDCG    & 28.91 & 7.16 & 80.86 & 21.76 \\
\midrule
\multirow{2}{*}{\shortstack[l]{Text\\Encoder}}
& AttnPool            & 28.85 & \textbf{7.90} & \textbf{72.00} & \textbf{21.81} \\
& PerTokenScalar      & \textbf{30.11} & 7.36 & 70.48 & 21.79 \\

\bottomrule
\end{tabular}
}
\end{table}

We now provide extensive ablation studies of the design elements/choice in \np. Specifically, Table~\ref{tab:ablation} ablates four independent design choices of Naïve PAINE and evaluates the resulting predictors on the PickScore test set using the SDXL DM unless otherwise stated.  In all cases the default configuration matches the model reported in the main paper. %

\subsection{Predictor Target Metric}
\label{sec:supp_abl_target}

The first segment of Tab.~\ref{tab:ablation} quantifies Figure~\ref{fig:ablate_target} with actual numerical values. Specifically, we train three predictors on SDXL using HPSv2~\cite{wu2023human},
PickScore~\cite{kirstain2023pick}, and ImageReward~\cite{xu2023imagereward} as the regression target while keeping all other hyperparameters fixed (\texttt{mae+srcc} loss, $N{=}100$, AttnPool text encoder). As shown in Figure~\ref{fig:ablate_target}, %
training with PickScore as the target yields the best overall downstream PickScore performance on generated images. This pattern is consistent: the predictor generalizes best to the metric it 
was %
trained to rank.

\subsection{Loss Function}
\label{sec:supp_abl_loss}

\begin{sloppypar}
The second section of Tab.~\ref{tab:ablation} considers the loss function utilized to train \np. We compare four configurations, each pairing the MAE regression term
with a different listwise ranking objective: (i) MAE\,+\,differentiable SRCC (\texttt{mae+srcc})~\cite{blondel2020fast}, (ii) MAE\,+\,LambdaRank
(\texttt{mae+lambdarank}), (iii) MAE\,+\,LambdaLoss (\texttt{mae+lambdaloss})~\cite{wang2018lambdaloss},
and (iv) MAE\,+\,NeuralNDCG (\texttt{neuralndcg})~\cite{pobrotyn2021neuralndcg}.
In every variant the MAE term anchors the predictions to the normalized target magnitude, while the ranking term shapes the within-prompt ordering. \texttt{mae+lambdarank} and \texttt{mae+lambdaloss} are two instances of the LambdaLoss framework~\cite{wang2018lambdaloss}: For each prompt, they encourage the model to rank higher-quality candidates above lower-quality ones, giving larger weight to pairwise mistakes that would most affect NDCG@5. The two losses differ mainly in the specific NDCG-based weighting formulation they use. In contrast, \texttt{neuralndcg} uses a differentiable approximation to sorting, allowing the model to optimize a smooth surrogate of NDCG@5 over the full candidate list. All losses are computed within prompt-level candidate groups using the grouped batch sampler. Results show that \texttt{mae+srcc} remains the stronger default across DMs and metrics.
\end{sloppypar}

\subsection{Text Encoder Architecture}
\label{sec:supp_abl_textenc}

Finally, the last segment of Tab.~\ref{tab:ablation} considers the architecture of the PAINE prompt embedding neural network $\Phi_{prompt}$. Specifically, we compare two text encoder designs: (i) the proposed \texttt{AttnPool} encoder (learnable summary token + 2 self-attention blocks, 16 heads) and (ii) \texttt{PerTokenScalar}, a simpler per-token MLP that projects each token embedding to a scalar and aggregates them. AttnPool consistently outperforms PerTokenScalar, which we attribute to the attention mechanism's ability to model inter-token dependencies and focus on semantically salient tokens rather than treating each token independently.

\subsection{Sensitivity to Preference Metrics}
\label{sec:supp_abl_preference}
We ablate the sensitivity of \np\ to different human preference metrics. Specifically, we vary the choice of human preference metric utilized to train the predictor, from HPSv2~\cite{wu2023human} to ImageReward~\cite{xu2023imagereward} to PickScore~\cite{kirstain2023pick} for \np\ on SDXL. We then observe the effect on T2I quality by using these predictors to generate images using prompts from the PickScore test set.

\begin{wrapfigure}{rt}{0.65\linewidth}
    \centering
    \vspace{-6mm}
    \includegraphics[width=\linewidth]{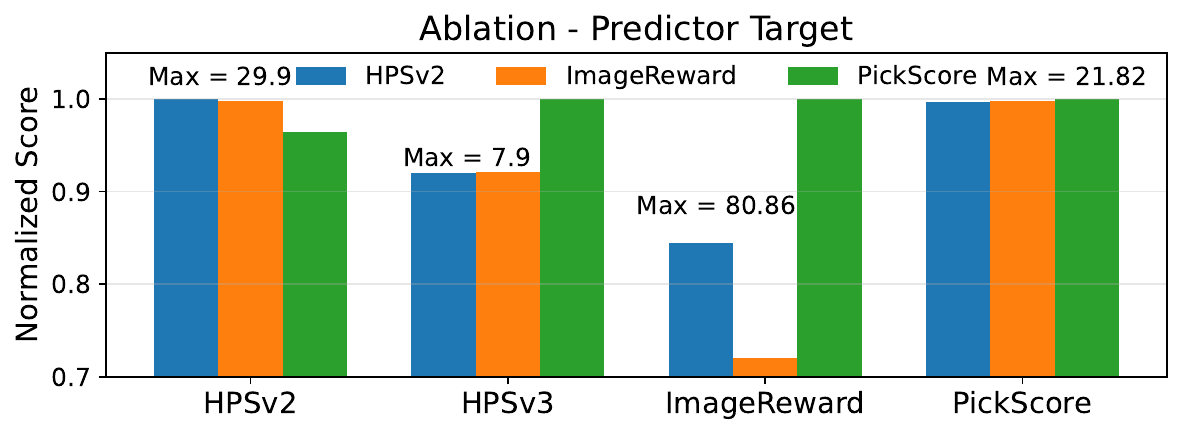}
    \caption{Comparing the performance of \np\ on the PickScore test set when we vary the choice of preference metric utilized to train the predictor. Values normalized based on the maximum (annotated).}
    \label{fig:ablate_target}
    \vspace{-6mm}
\end{wrapfigure}

Figure~\ref{fig:ablate_target} illustrates the result. First, we observe that while both HPSv2 and ImageReward, when used as a predictor metric, achieve slightly superior HPSv2 evaluation performance compared to PickScore, %
they both fall short of it on HPSv3~\cite{ma2025hpsv3} and ImageReward evaluation performance. Ironically, when we use ImageReward scores to train \np, it provides the worst ImageReward benchmark performance. Further, %
using PickScore to train the predictor also yields the best PickScore evaluative performance. This finding is one %
example of why we select %
PickScore to generate results in Section~\ref{sec:eval} as we ran many trials and generally found that it achieves %
better overall performance in comparison to HPSv2 or ImageReward, and %
compared %
to other baselines. We provide additional ablations and examples in the %
supplementary materials.

\section{Evaluation on OOD Prompts w/ LoRA}
We %
consider the applicability of \np\ when integrated with open-source DM checkpoints and Low-Rank Adapters (LoRA)~\cite{dettmers2024qlora}. Specifically, we consider SDXL-based %
fine-tunes, %
pair them with a LoRA and our \np\ predictor for SDXL. %

\begin{figure}[t!]
    \centering
    \setlength{\tabcolsep}{1.2pt}
    \renewcommand{\arraystretch}{1.2}
    \footnotesize
    \scalebox{\scaleboxratio}{
      \begin{tabular}{@{}c@{\hspace{3pt}}c@{\hspace{3pt}}c@{\hspace{3pt}}c@{}}
          Standard & Golden Noise & NoiseAR & \np \\[2pt]

          \includegraphics[width=0.30\textwidth]{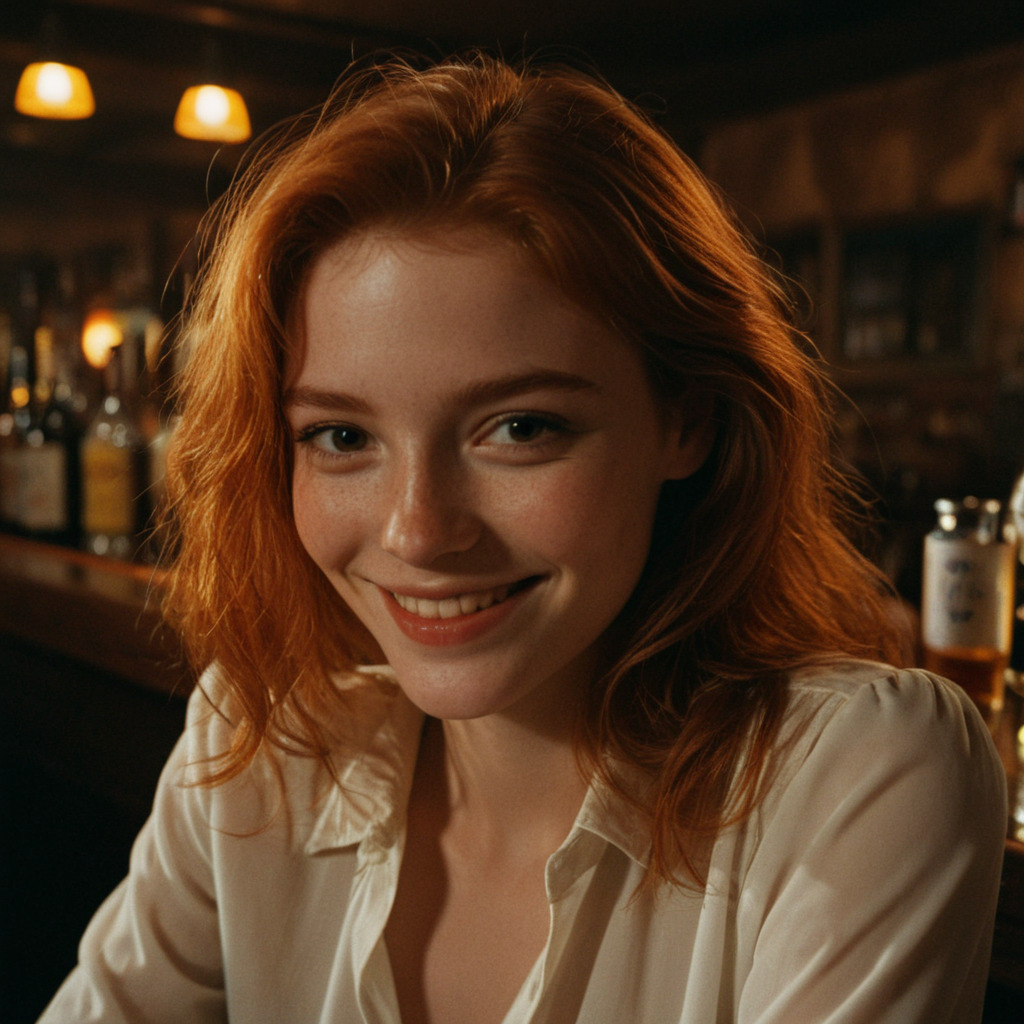} &
          \includegraphics[width=0.30\textwidth]{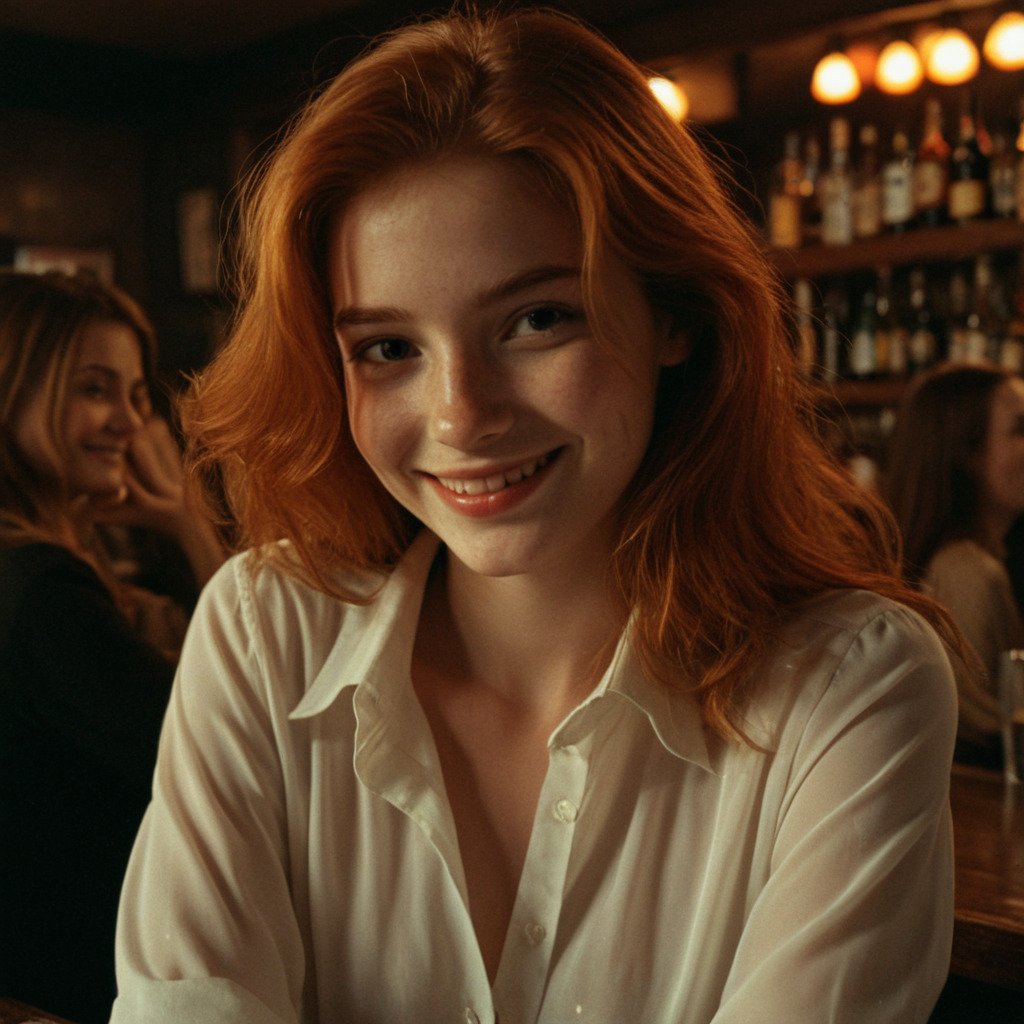} &
          \includegraphics[width=0.30\textwidth]{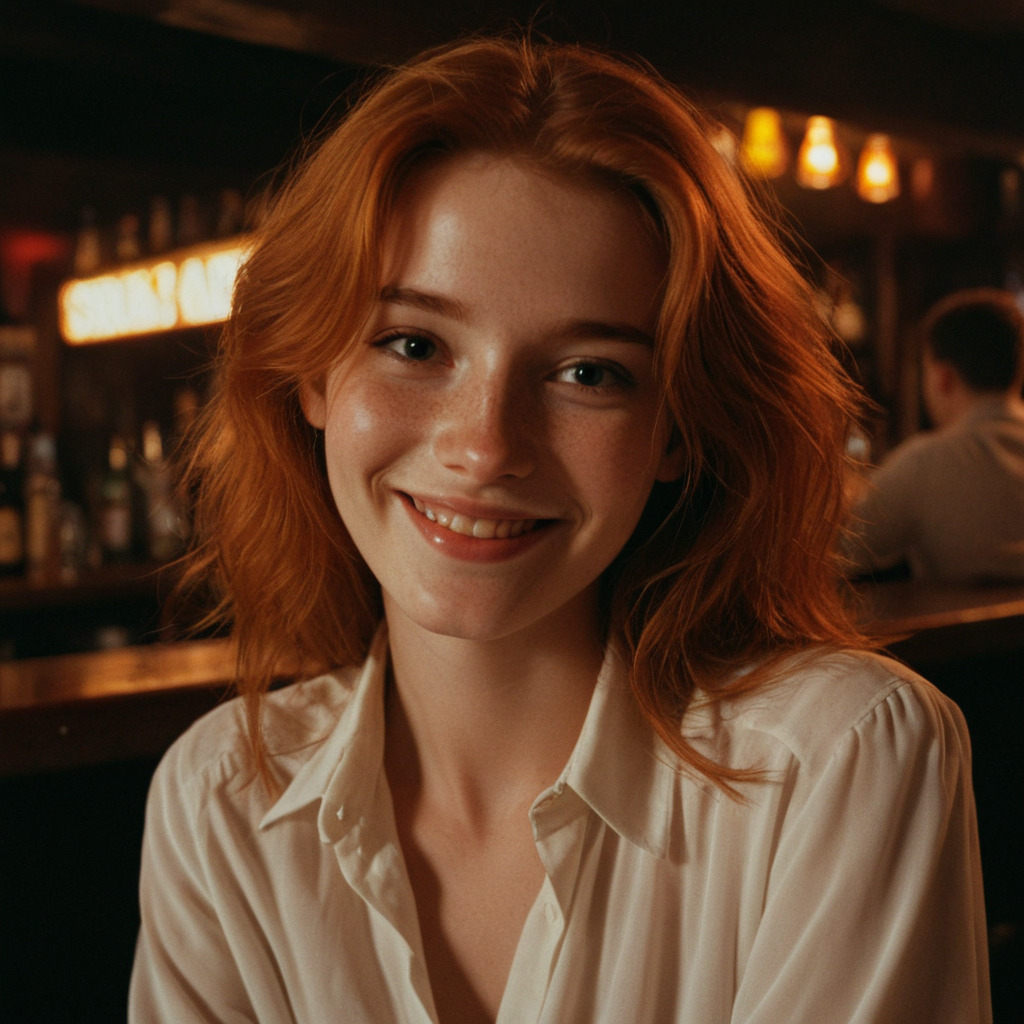} &
          \includegraphics[width=0.30\textwidth]{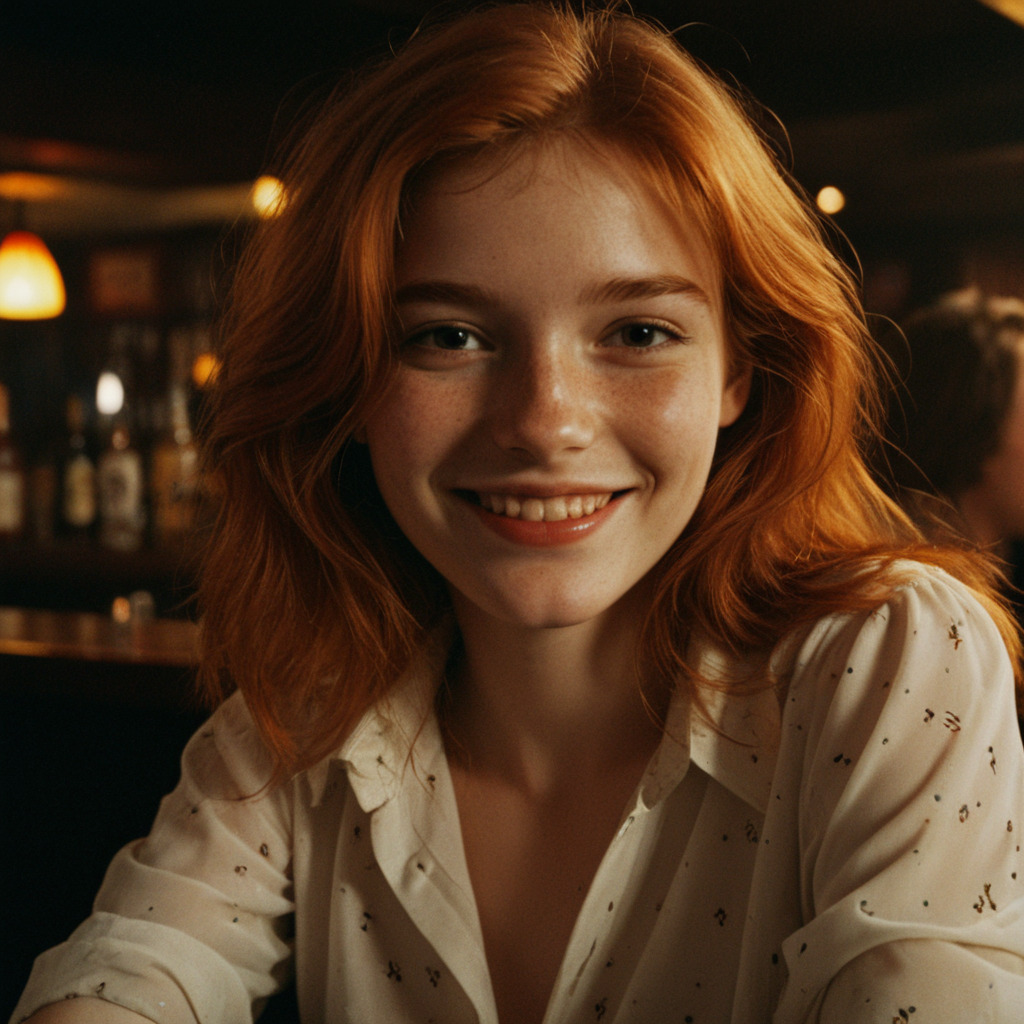} \\

          \includegraphics[width=0.30\textwidth]{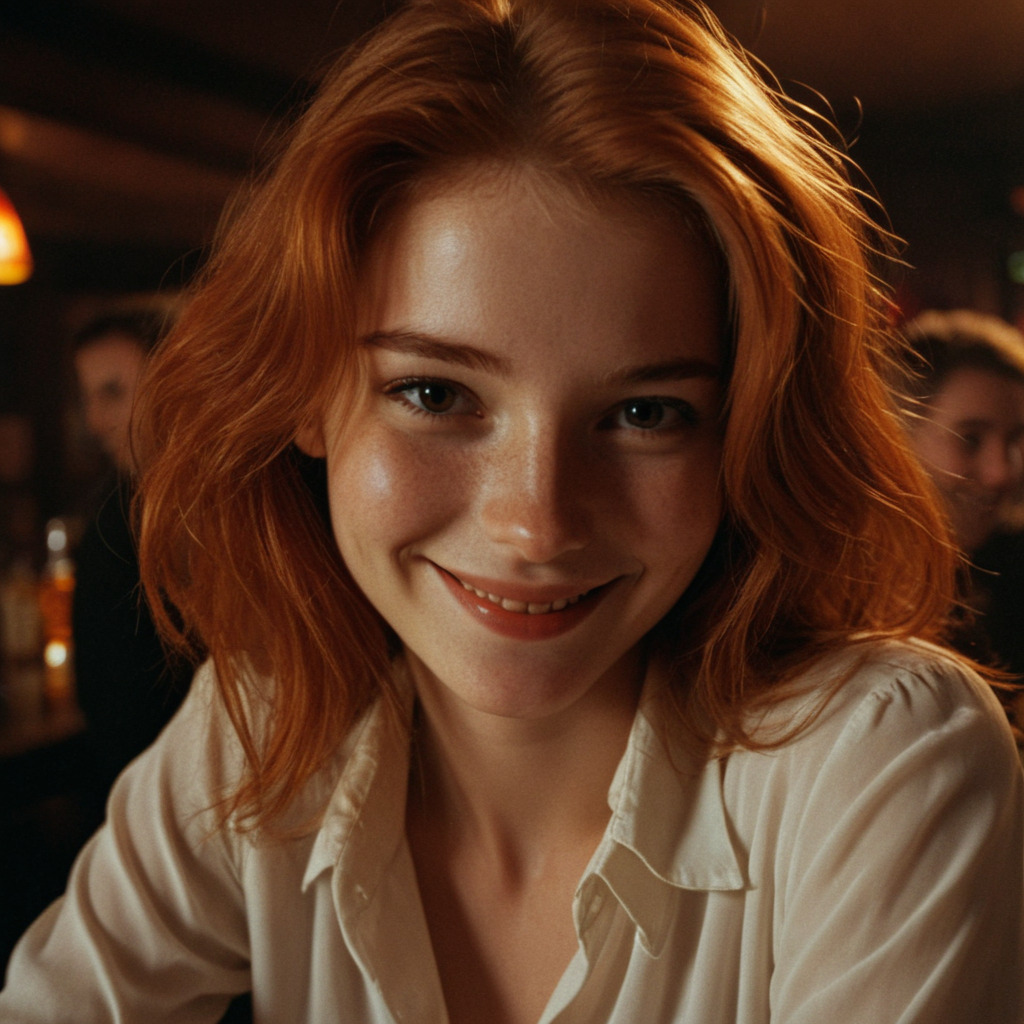} &
          \includegraphics[width=0.30\textwidth]{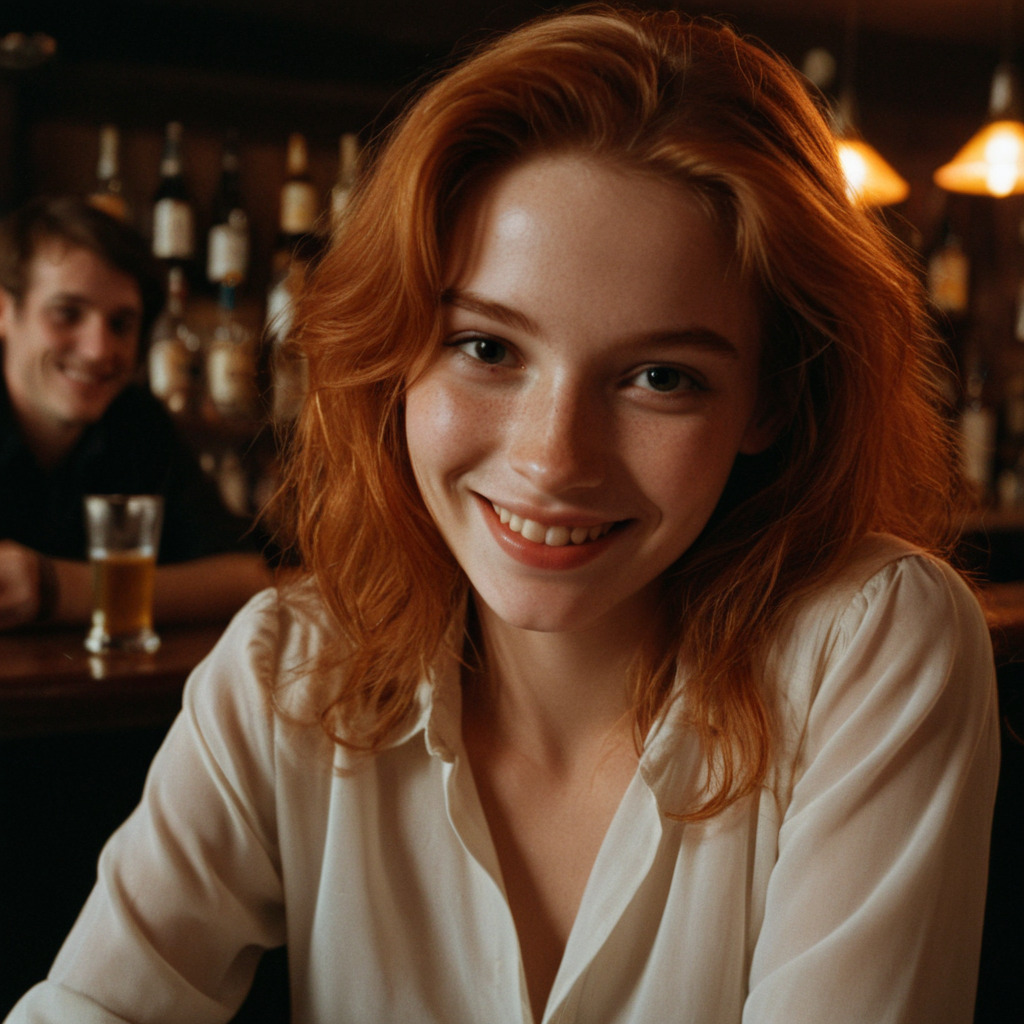} &
          \includegraphics[width=0.30\textwidth]{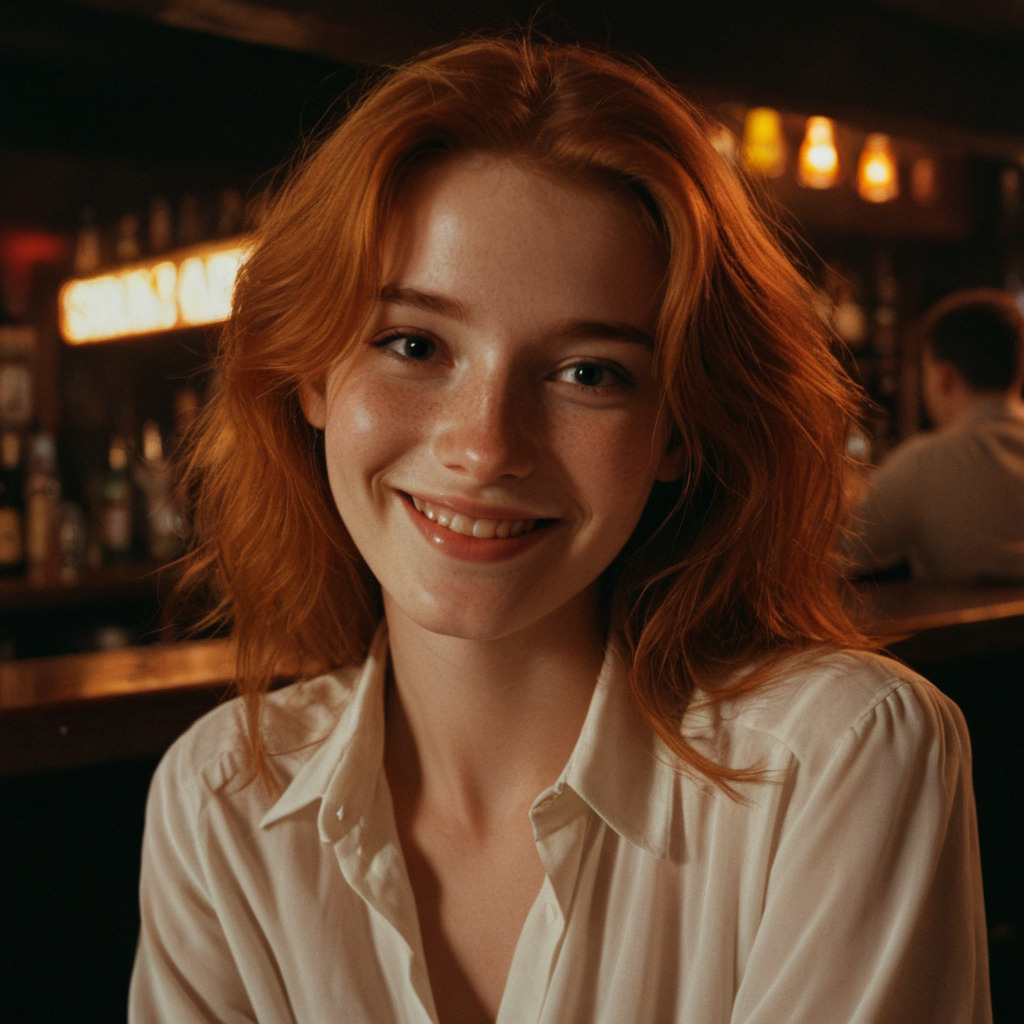} &
          \includegraphics[width=0.30\textwidth]{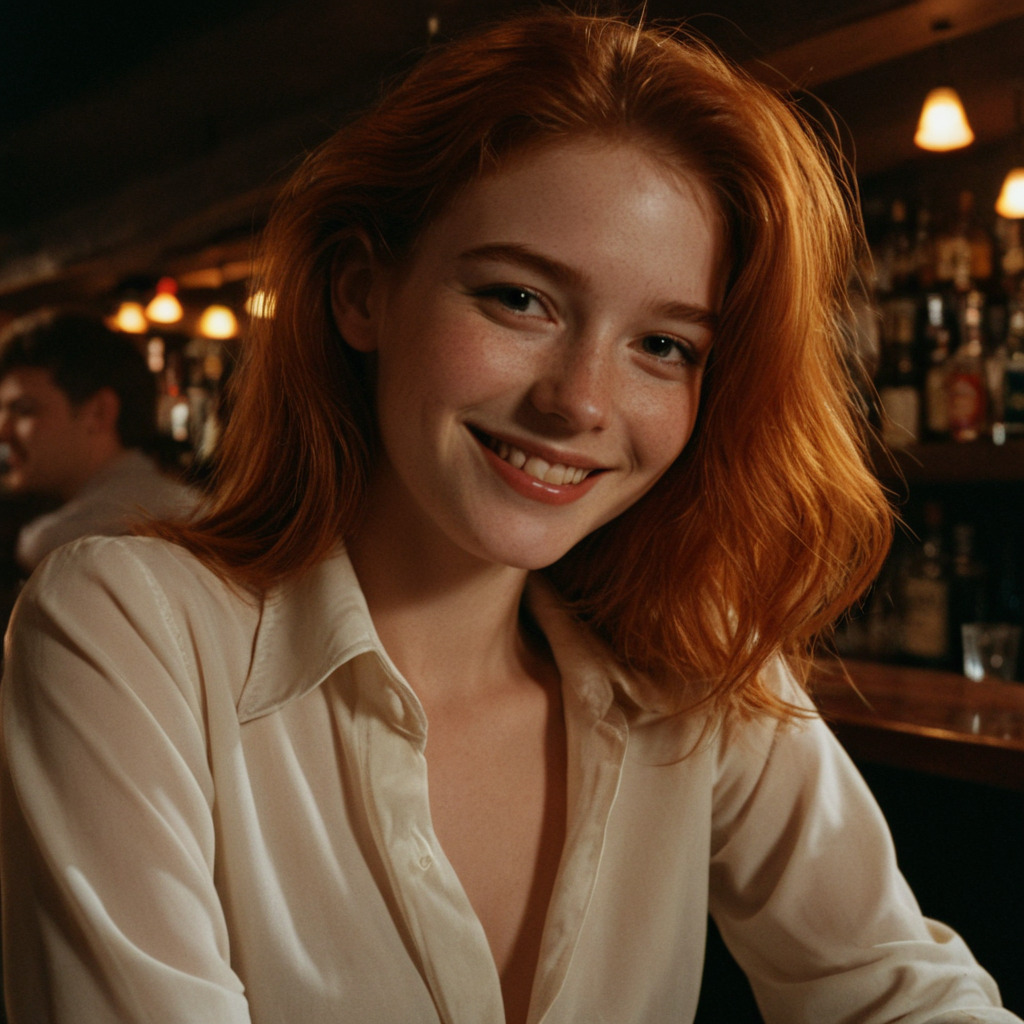} \\

          \includegraphics[width=0.30\textwidth]{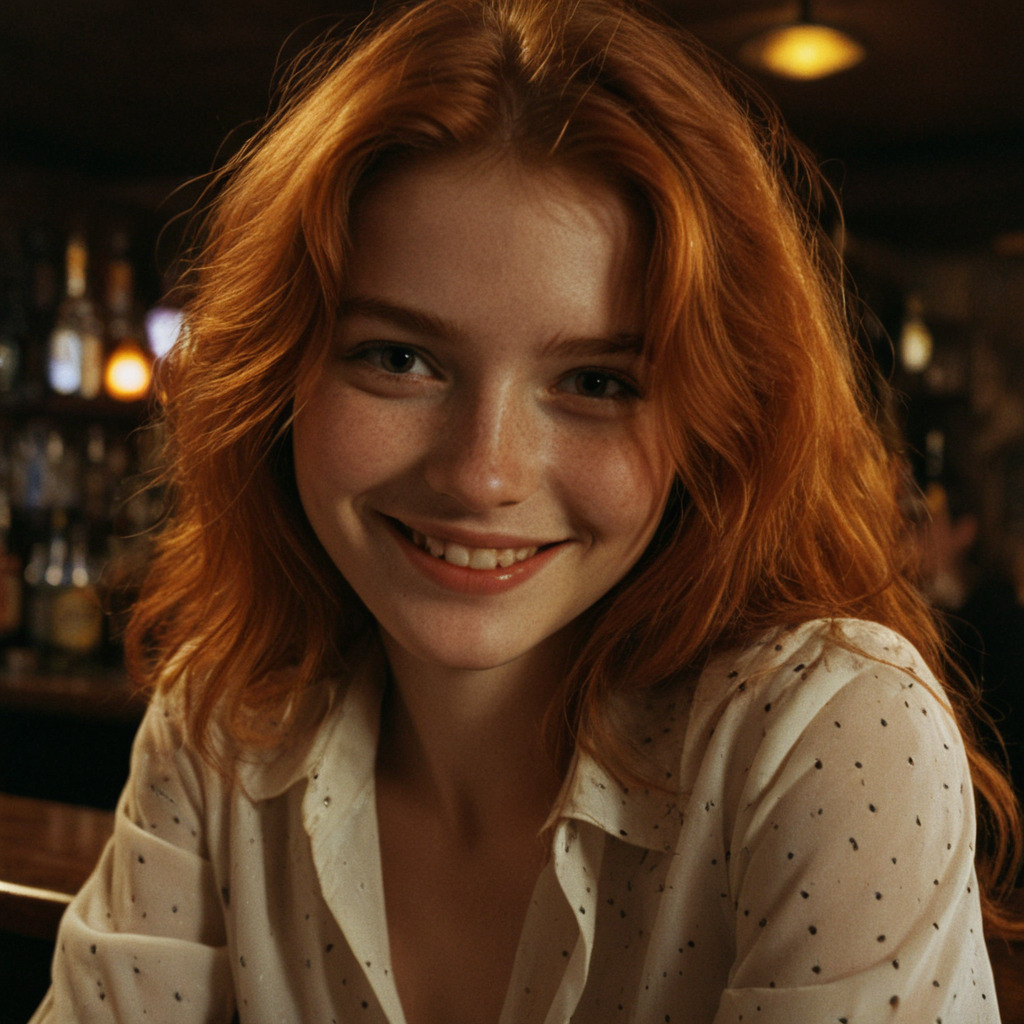} &
          \includegraphics[width=0.30\textwidth]{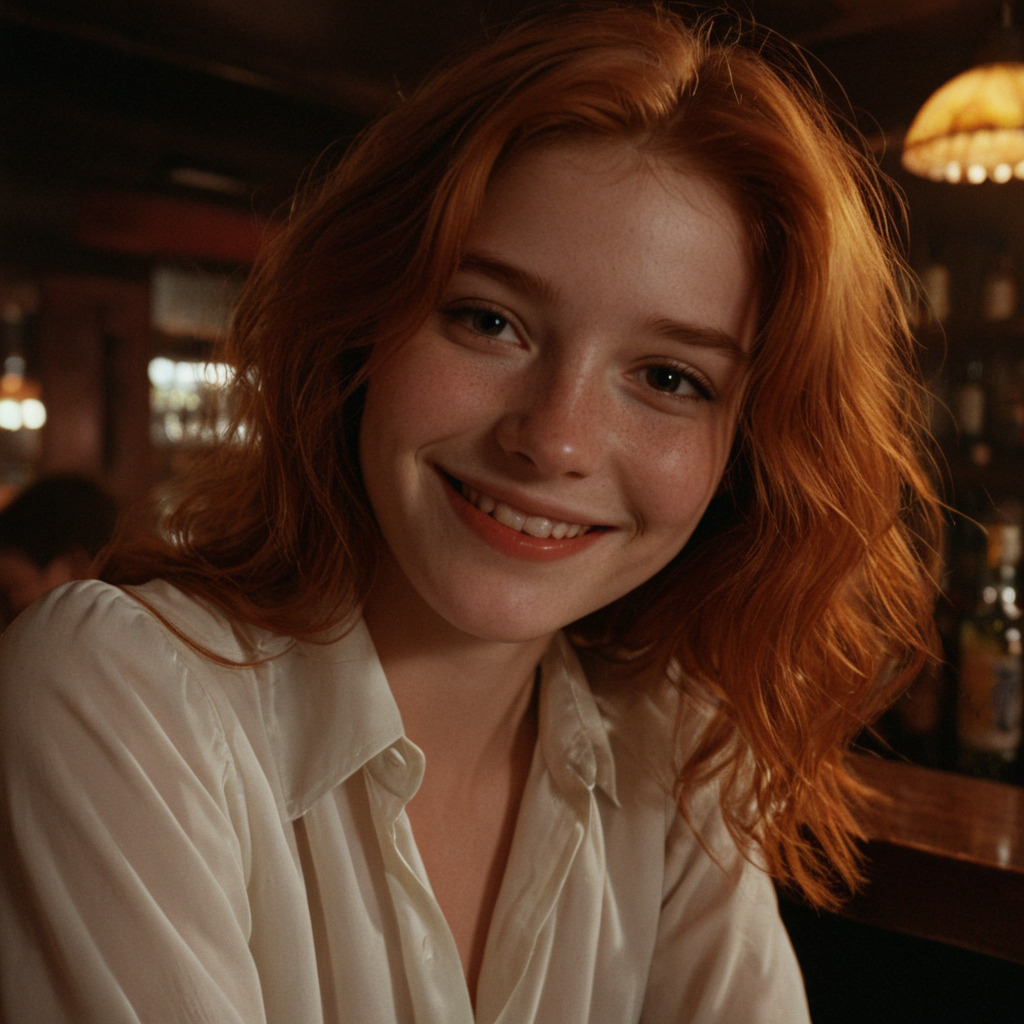} &
          \includegraphics[width=0.30\textwidth]{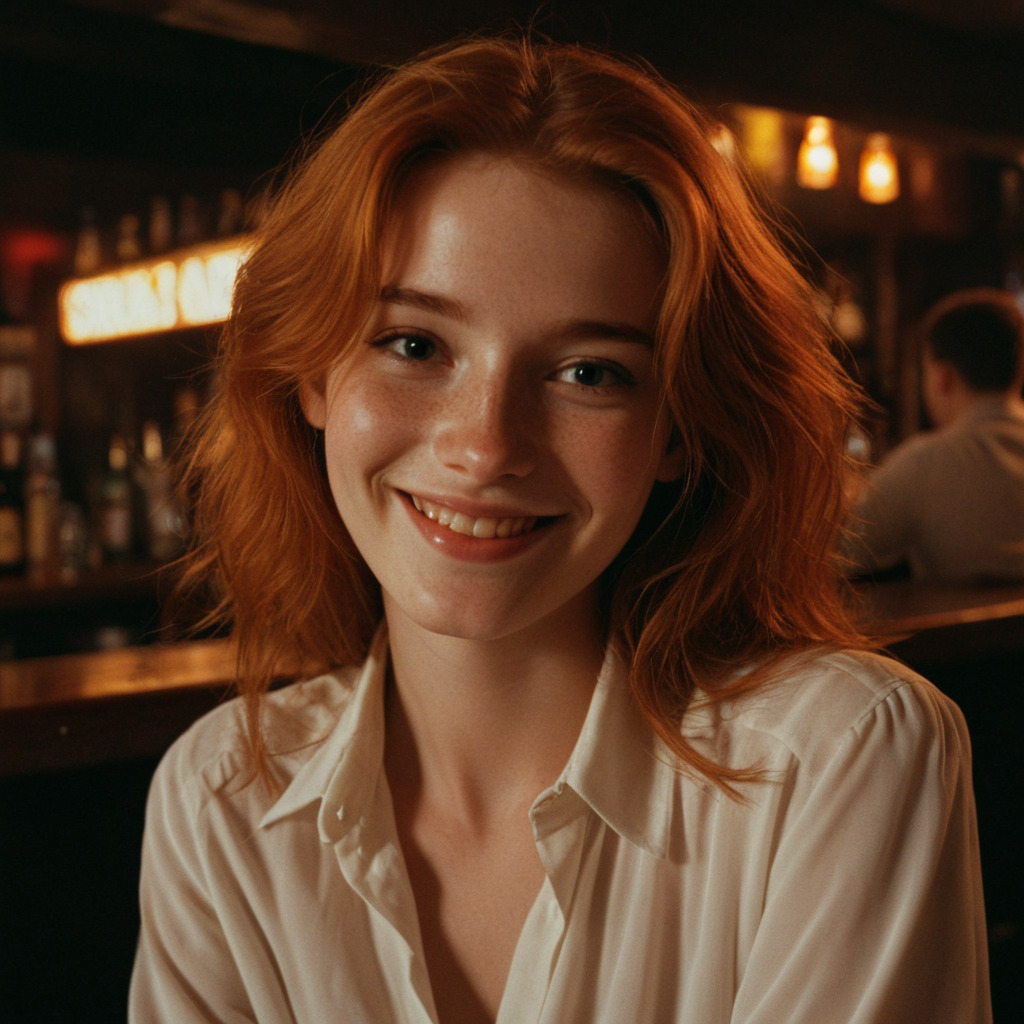} &
          \includegraphics[width=0.30\textwidth]{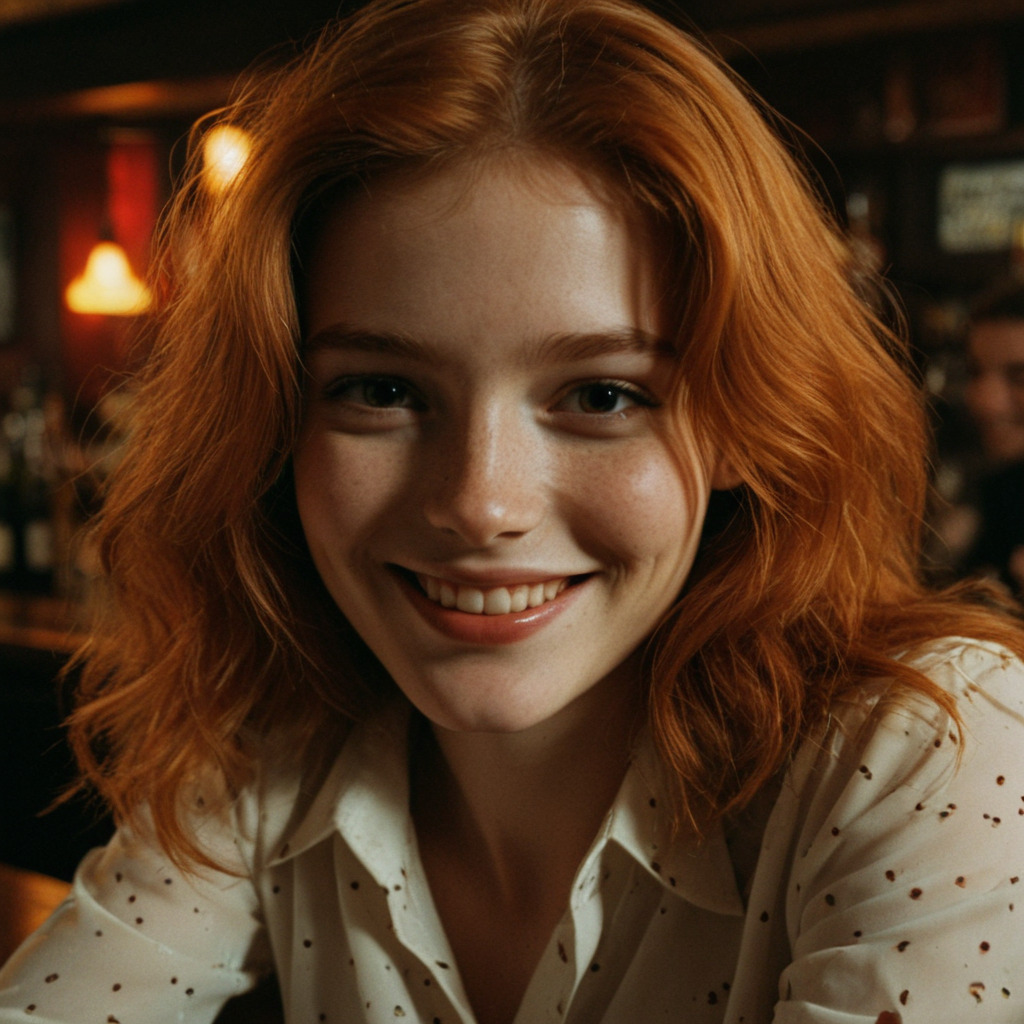} \\

          \includegraphics[width=0.30\textwidth]{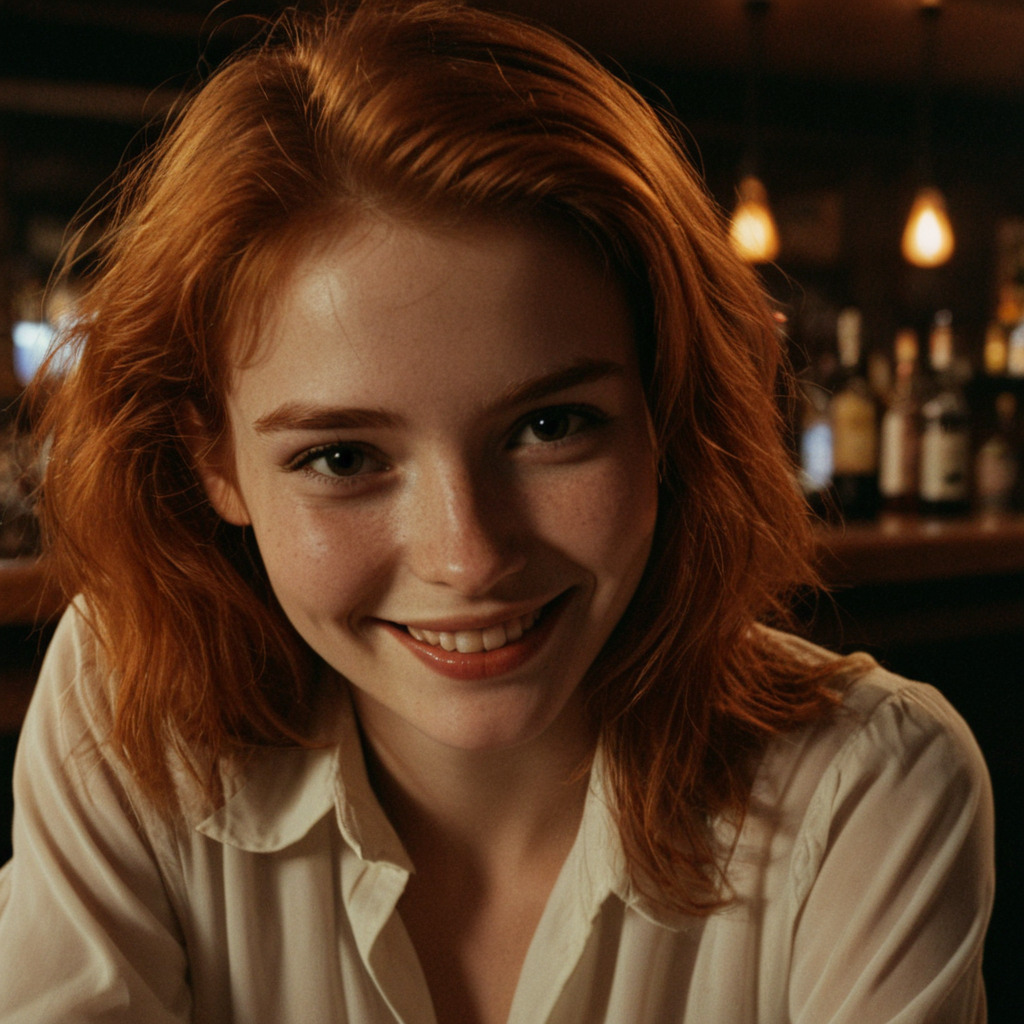} &
          \includegraphics[width=0.30\textwidth]{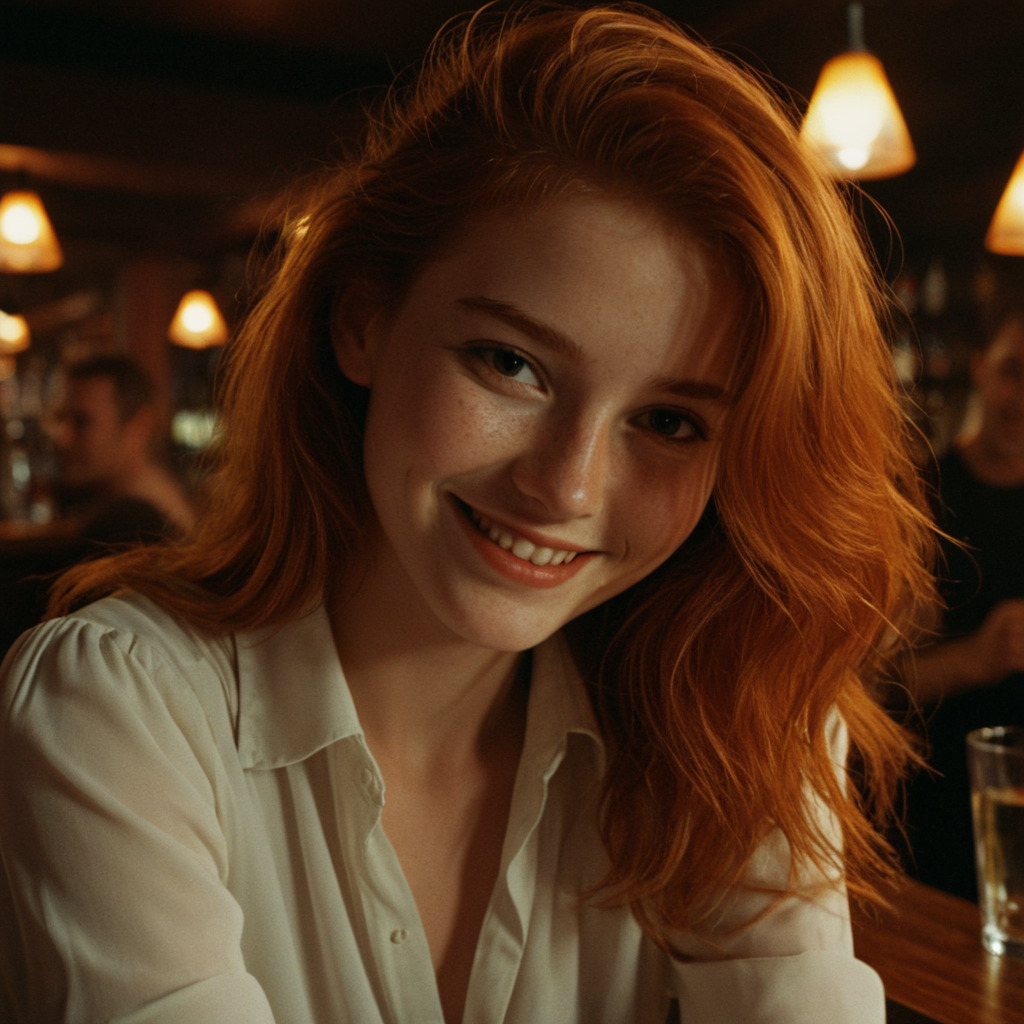} &
          \includegraphics[width=0.30\textwidth]{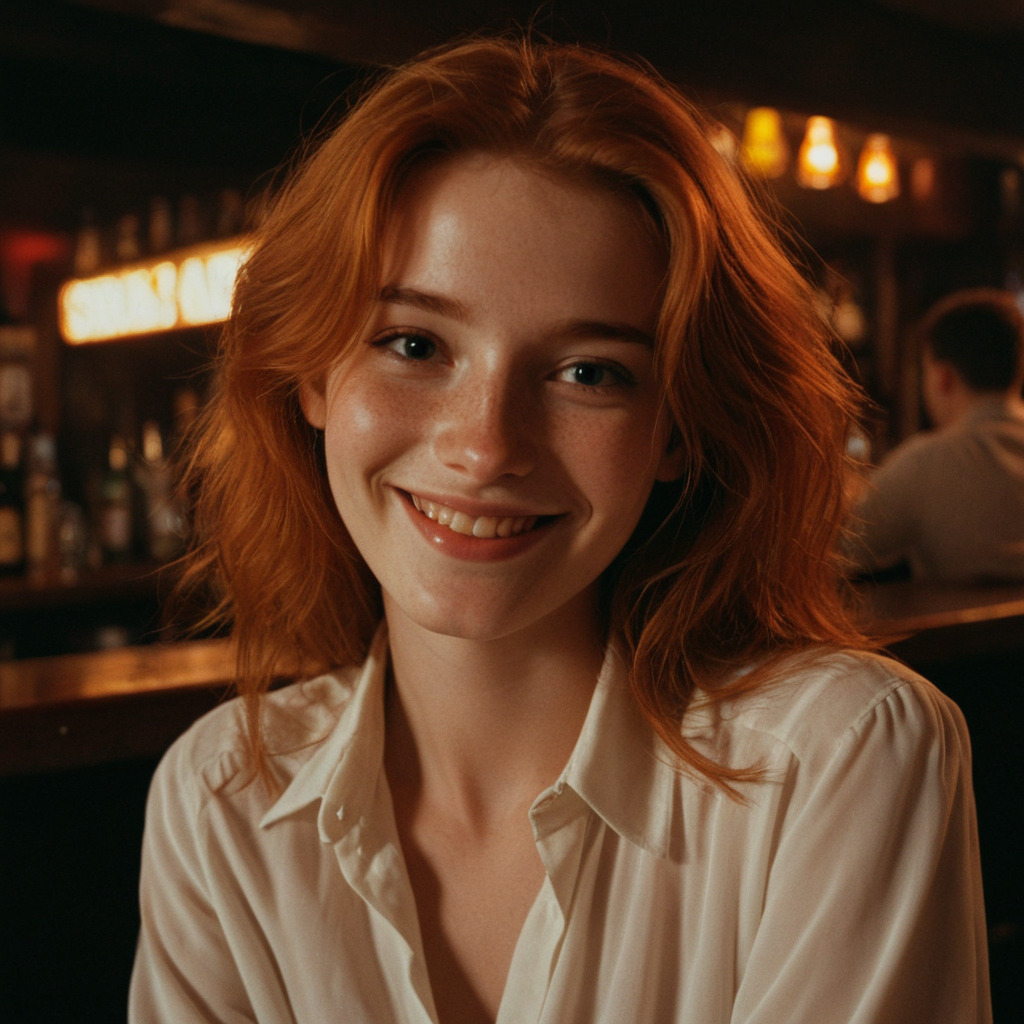} &
          \includegraphics[width=0.30\textwidth]{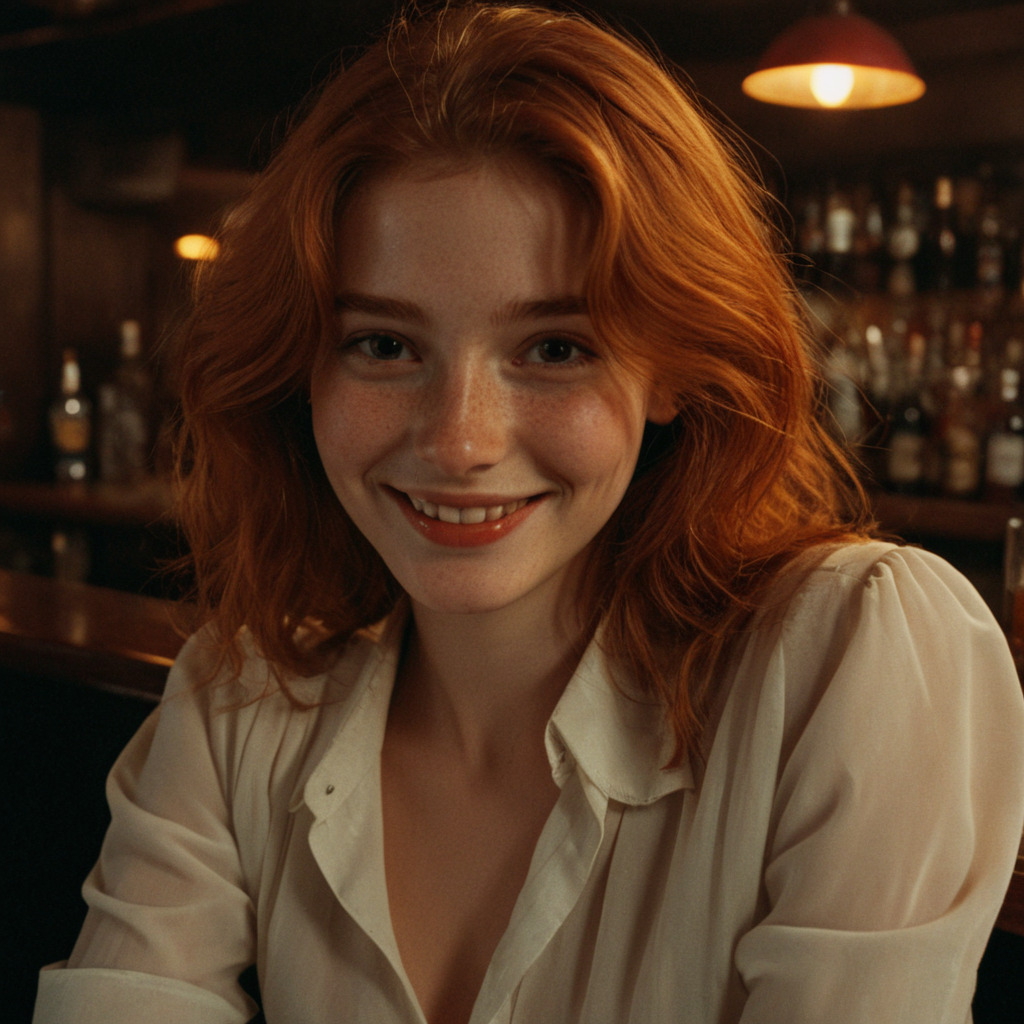} \\
          \textbf{PickScore:} \textit{25.43} & 25.31 & 25.37 & \textbf{25.68}

      \end{tabular}
      }
      \caption{Realistic image comparison when integrating noise optimization methods with open-source checkpoints and LoRAs. Prompt: `a close up photo of a 20 year old french woman in a blouse at a bar, seductive smile, ginger hair, cinematic light, film still'. Source checkpoint: \cite{civitaiJuggernautRagnarok_by_RunDiffusion}; LoRA: \cite{juggernaut_lora}. We \textbf{bold} and \textit{italicize} the best and second best average PickScore, respectively.}
      
      \vspace{-2mm}
      \label{fig:lora_examples_juggernaut_woman}
  \end{figure}

\begin{figure}[t!]
    \centering
    \setlength{\tabcolsep}{1.2pt}
    \renewcommand{\arraystretch}{1.2}
    \footnotesize
    \scalebox{\scaleboxratio}{
      \begin{tabular}{@{}c@{\hspace{3pt}}c@{\hspace{3pt}}c@{\hspace{3pt}}c@{}}
          Standard & Golden Noise & NoiseAR & \np \\[2pt]

          \includegraphics[width=0.30\textwidth]{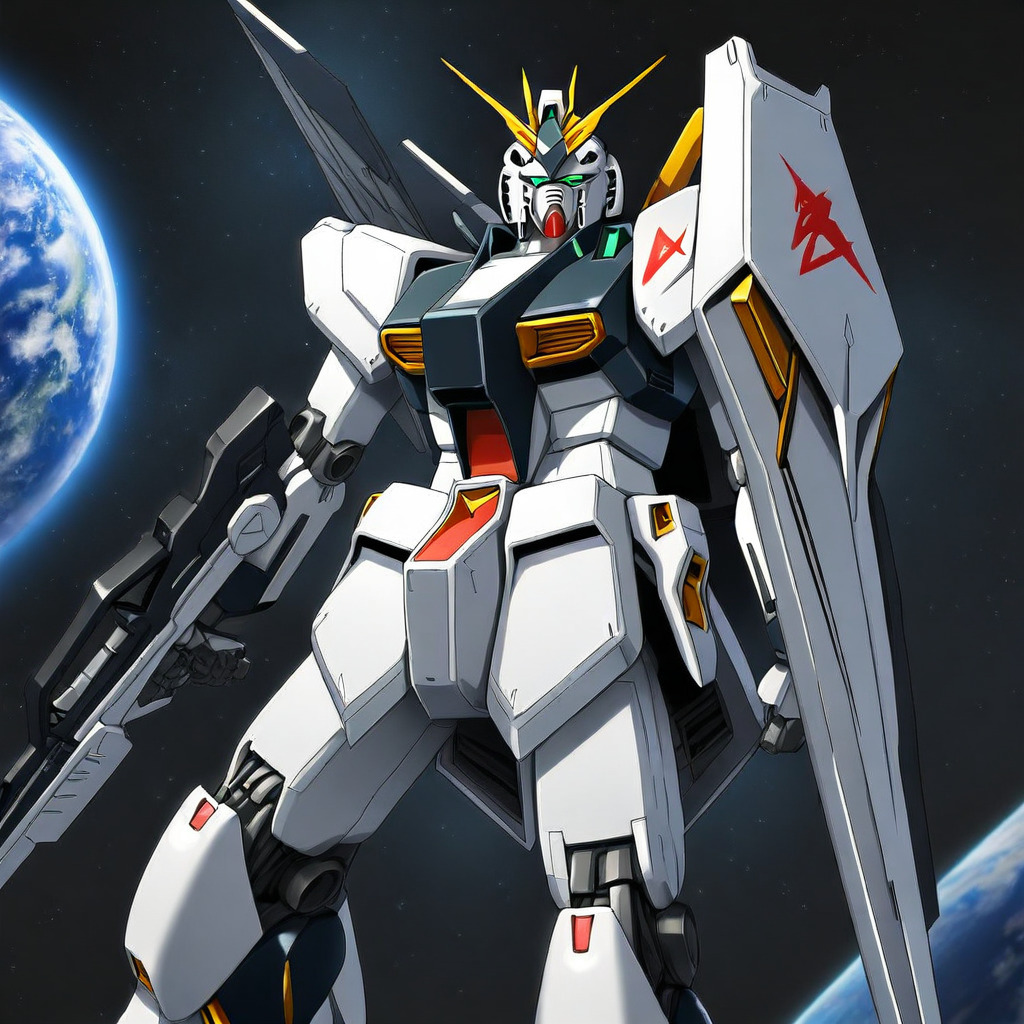} &
          \includegraphics[width=0.30\textwidth]{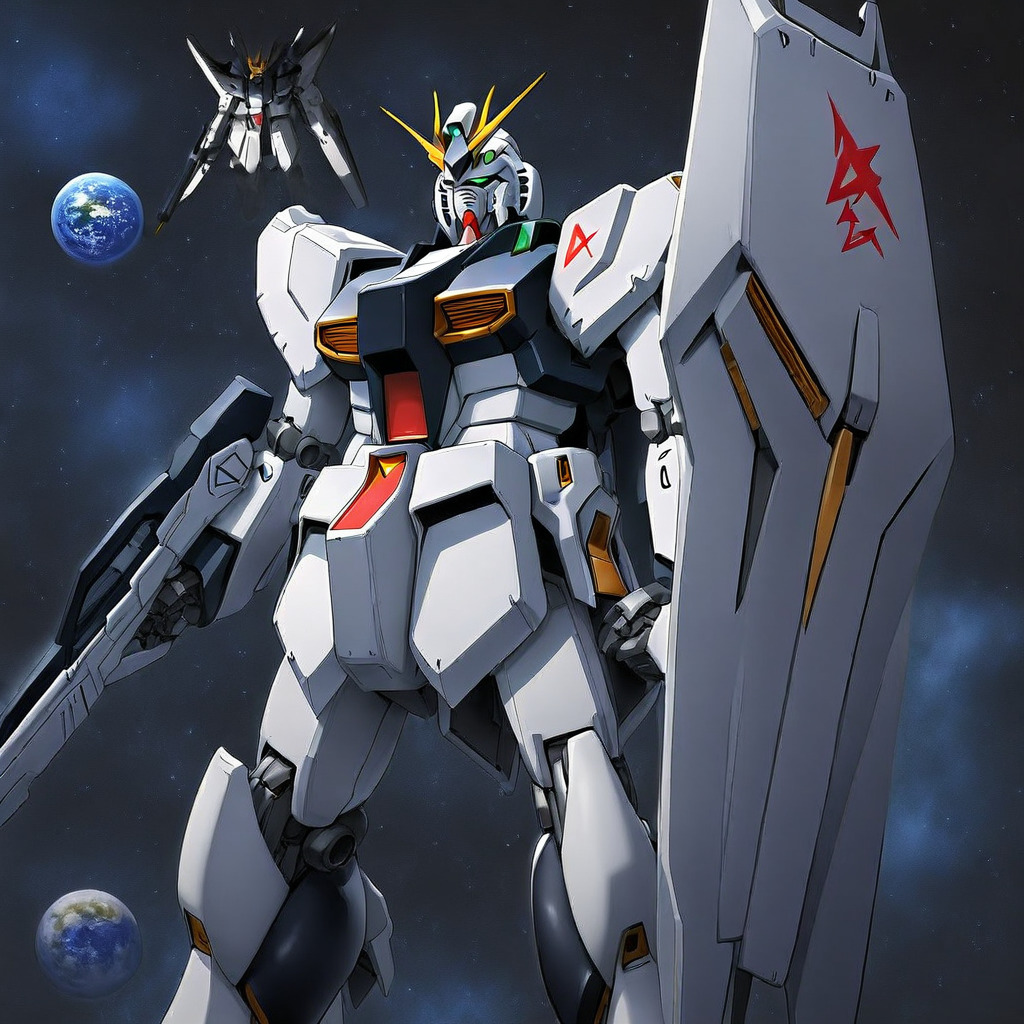} &
          \includegraphics[width=0.30\textwidth]{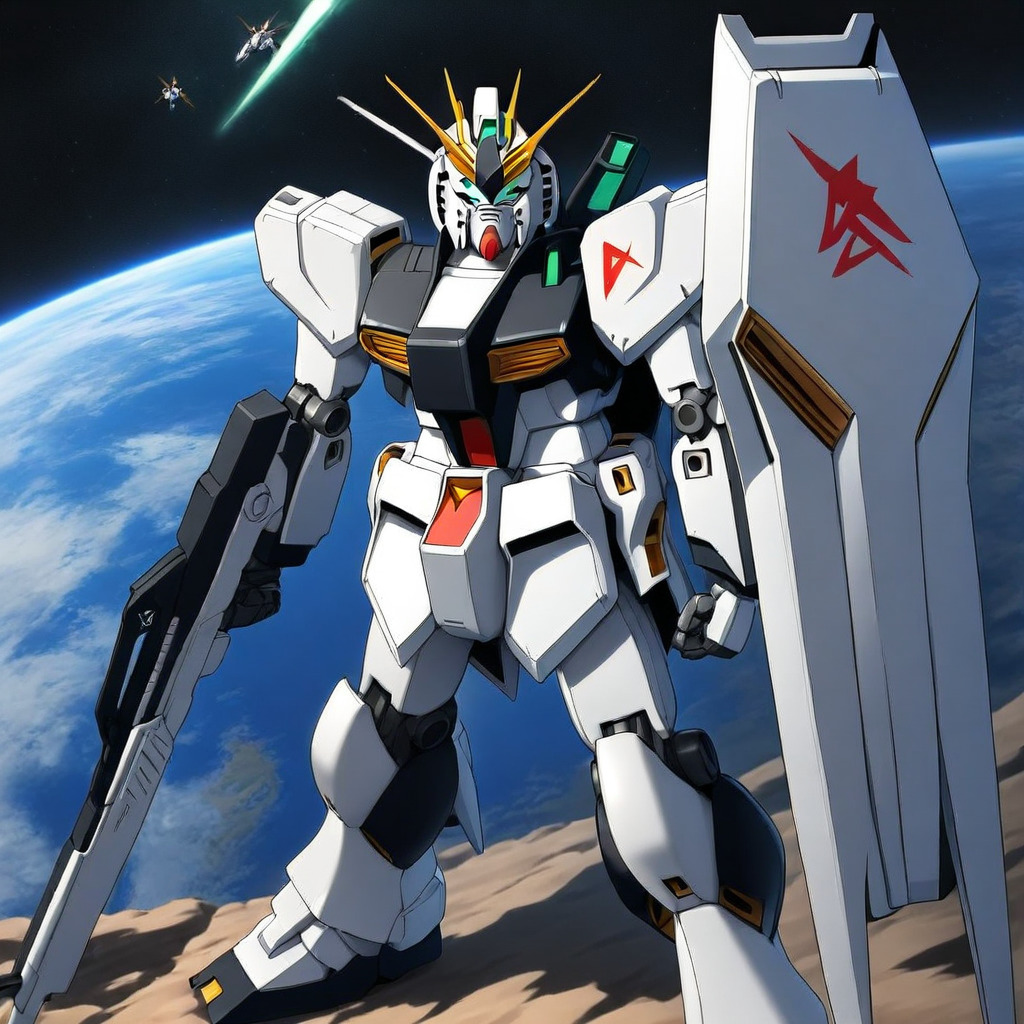} &
          \includegraphics[width=0.30\textwidth]{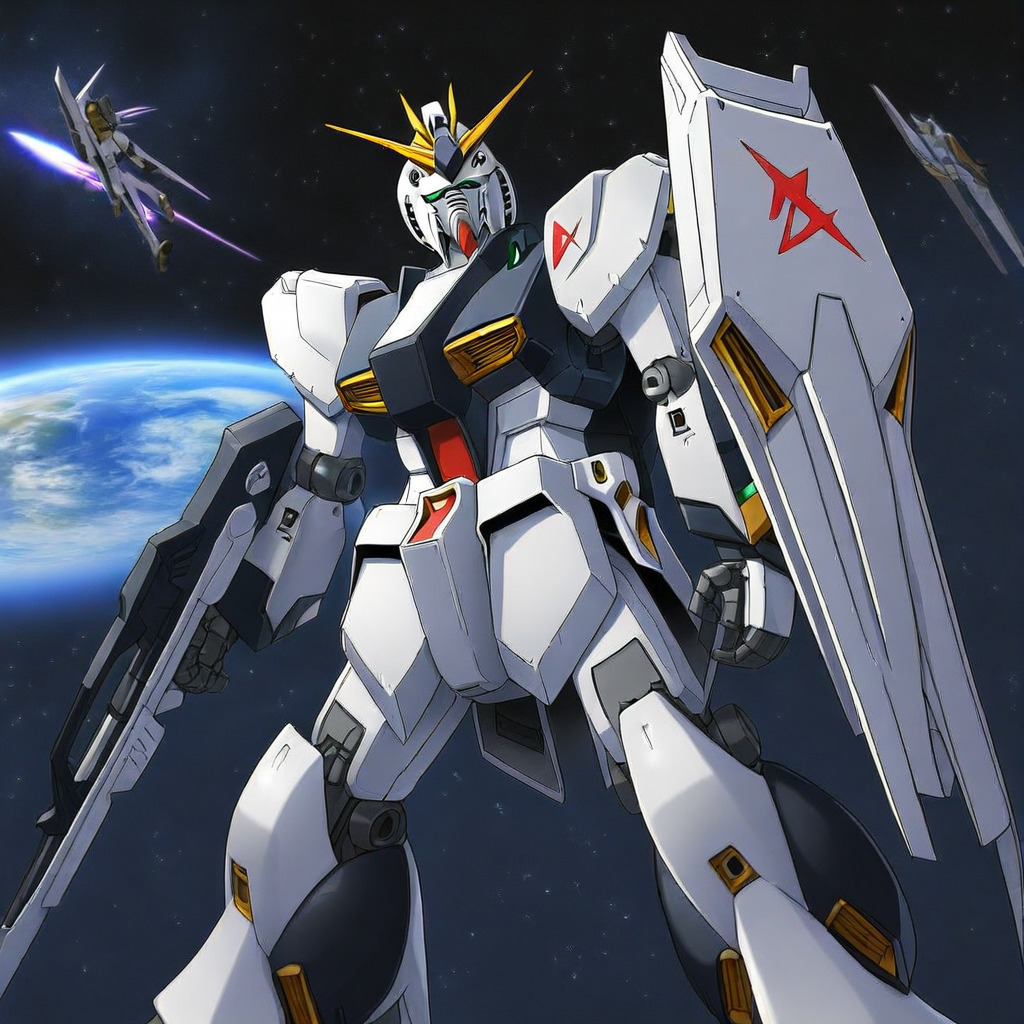} \\

          \includegraphics[width=0.30\textwidth]{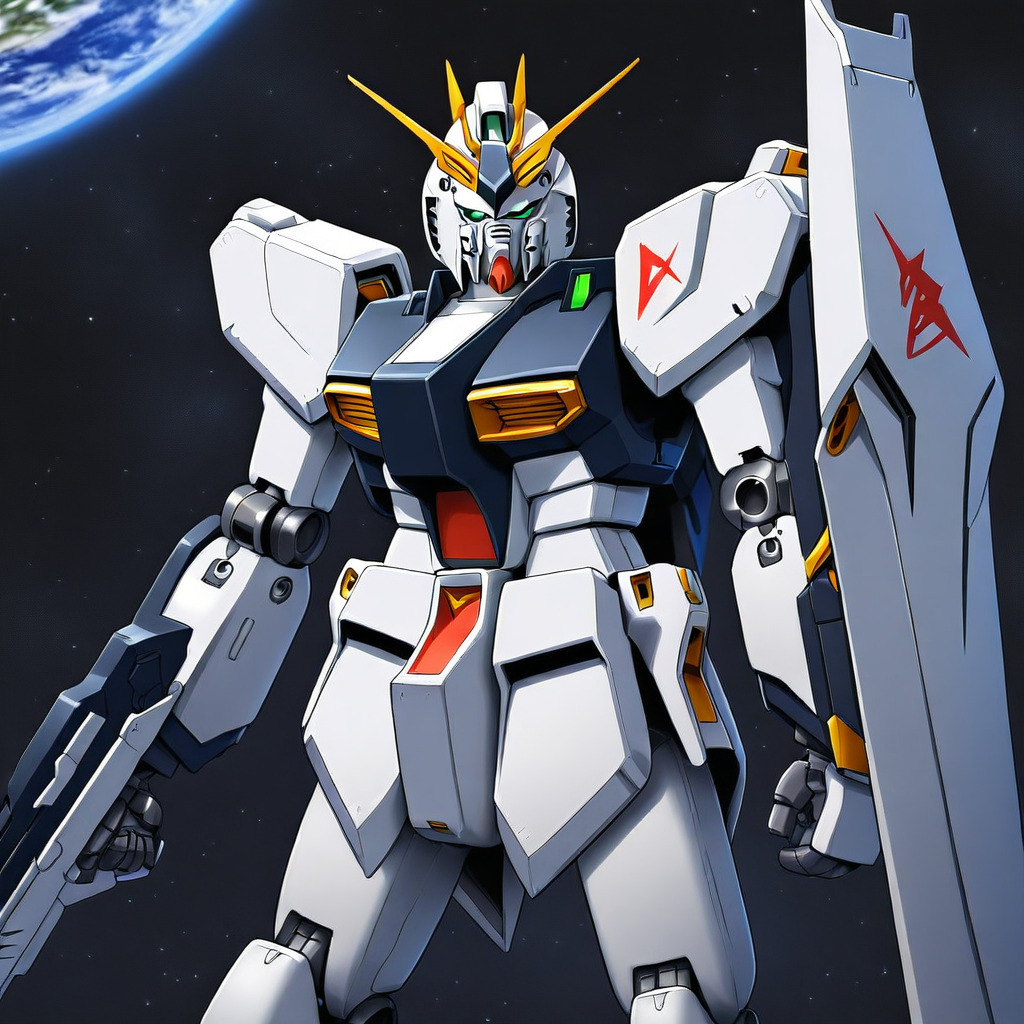} &
          \includegraphics[width=0.30\textwidth]{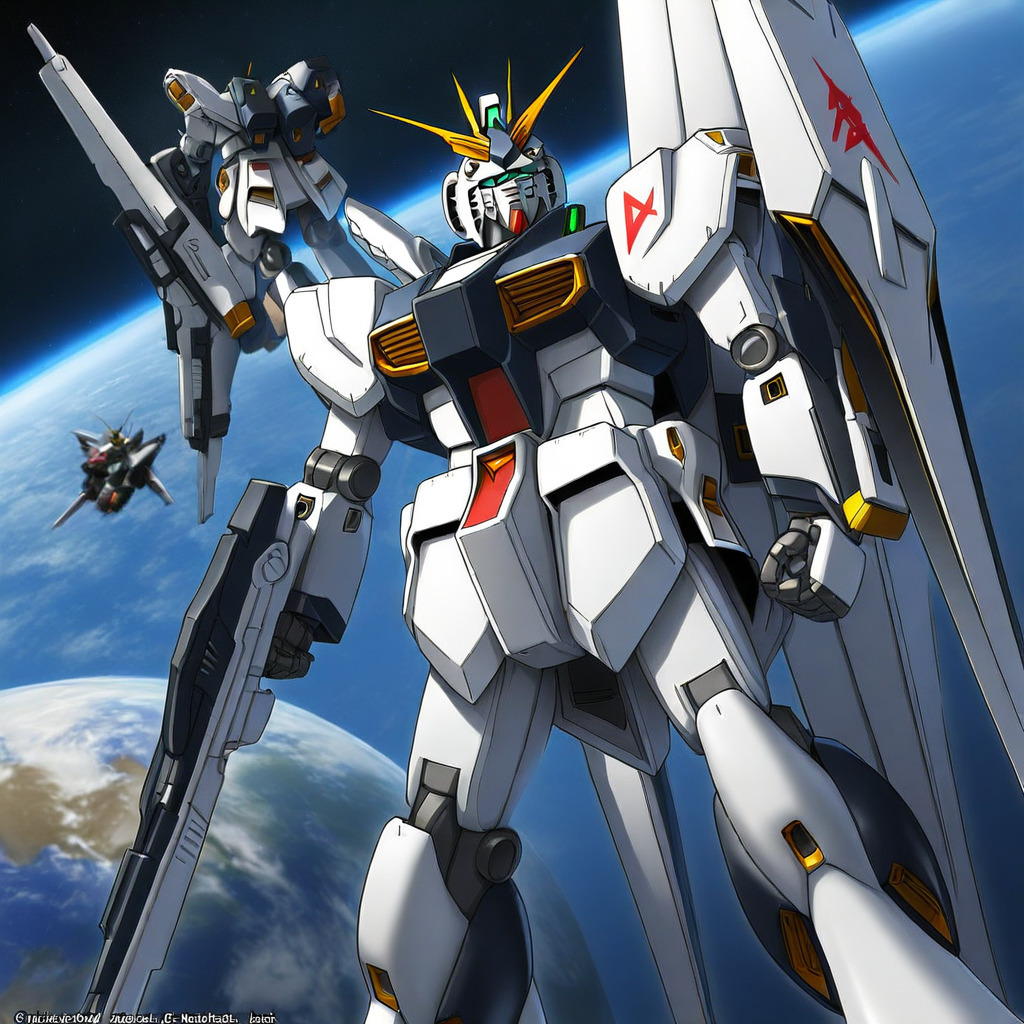} &
          \includegraphics[width=0.30\textwidth]{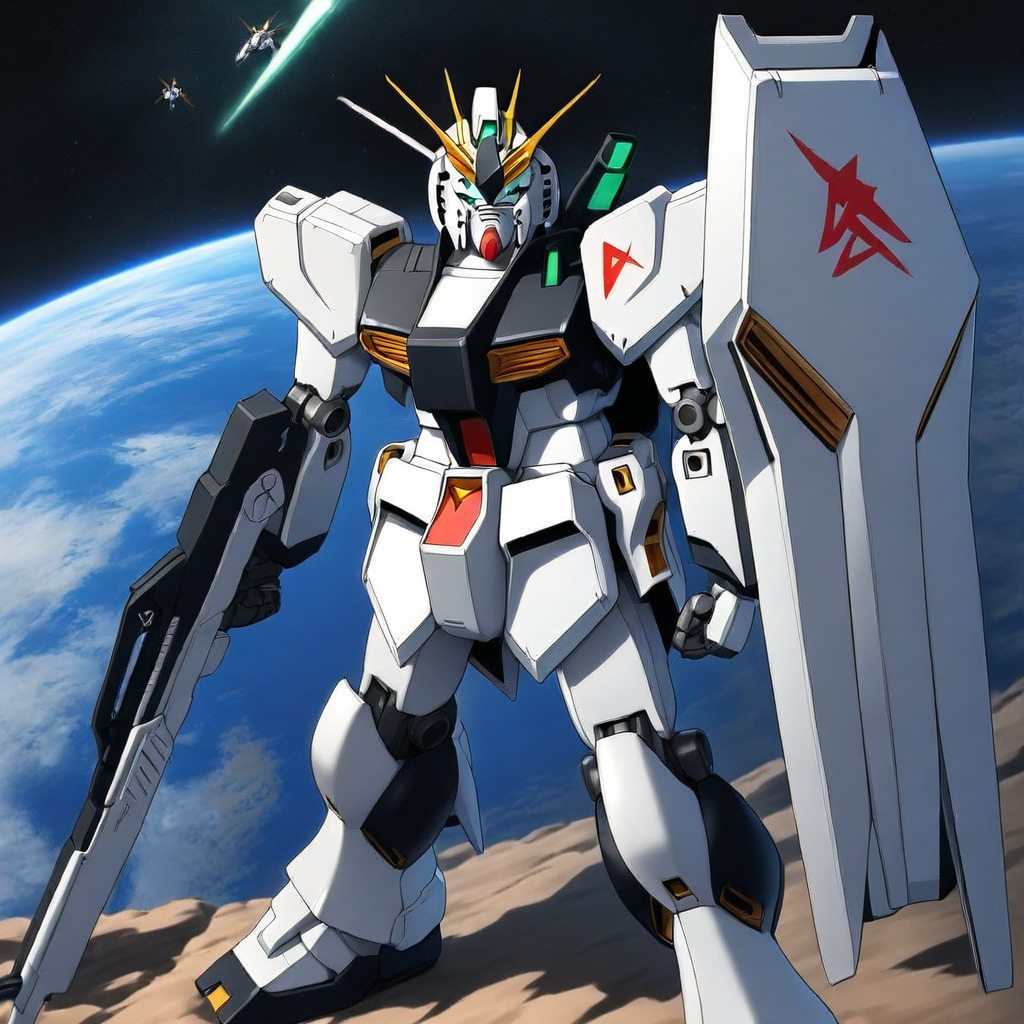} &
          \includegraphics[width=0.30\textwidth]{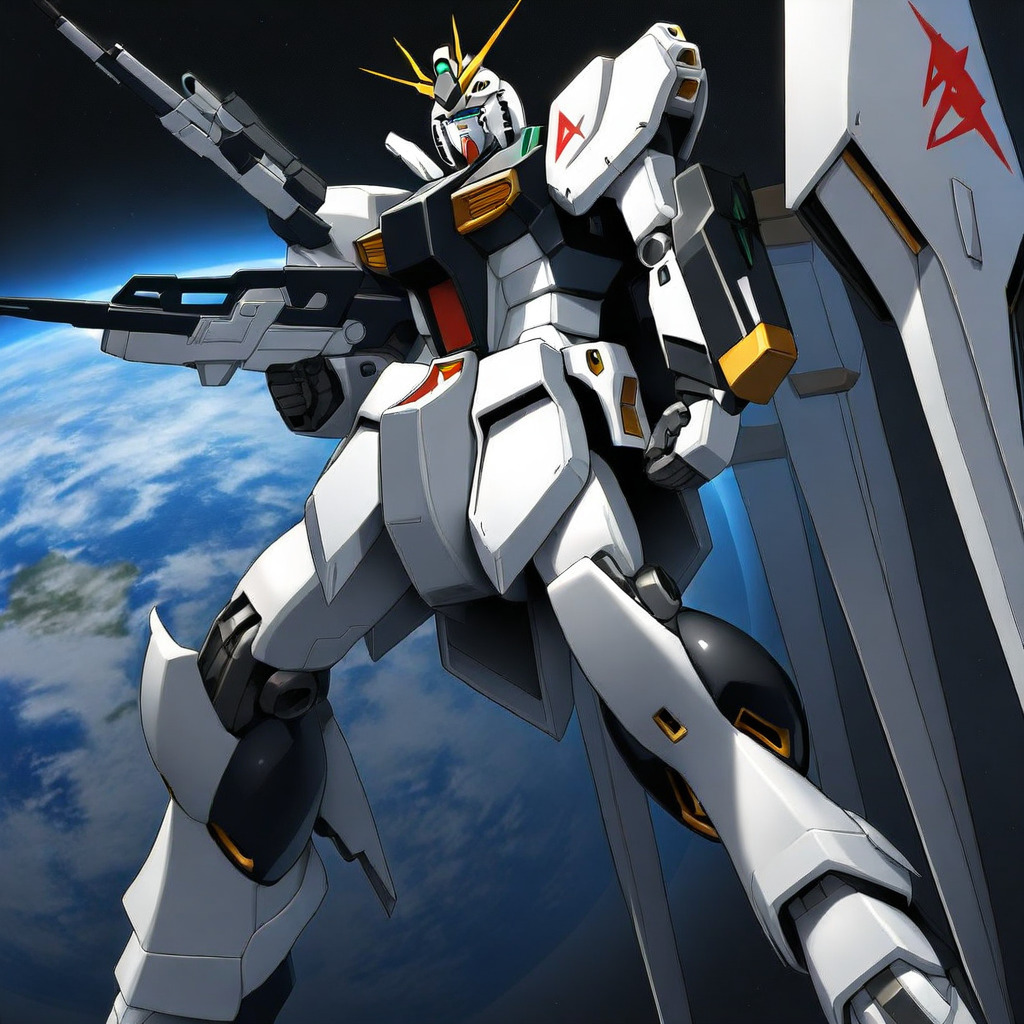} \\

          \includegraphics[width=0.30\textwidth]{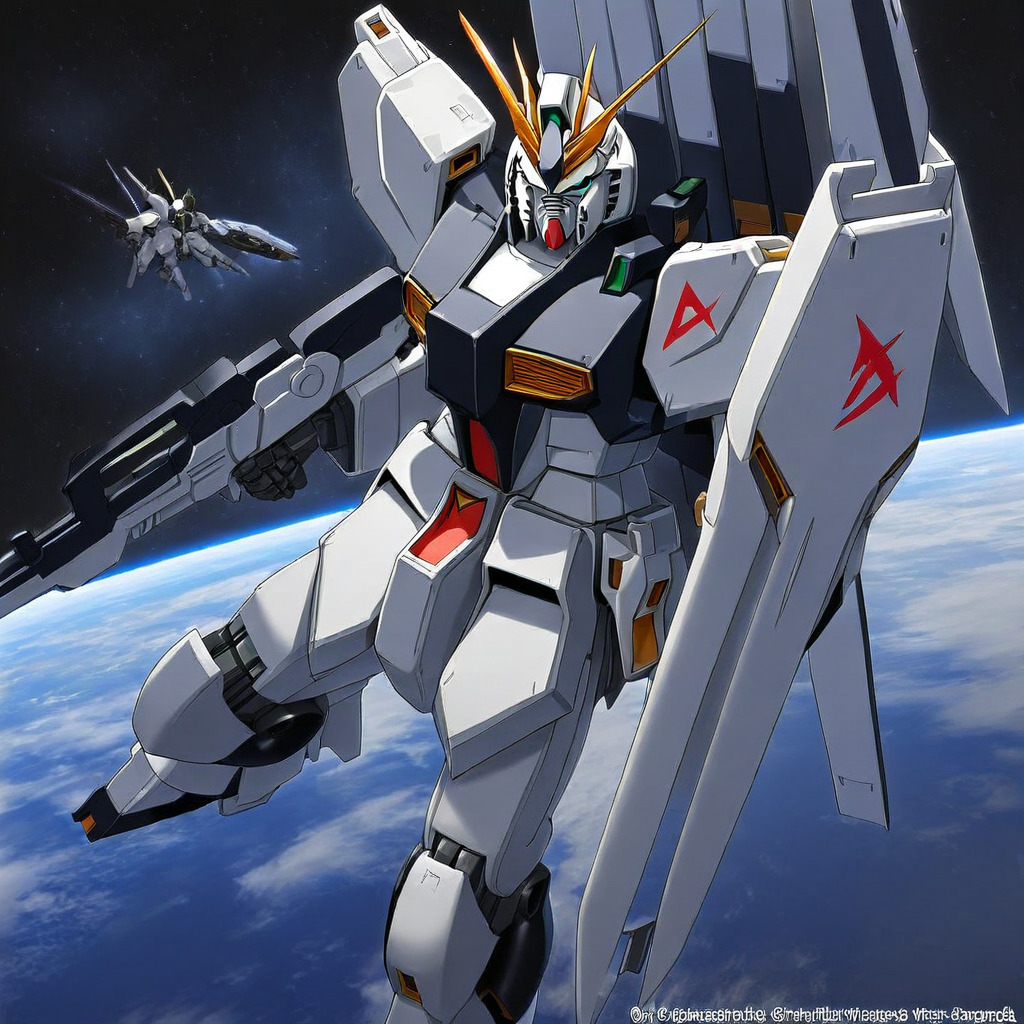} &
          \includegraphics[width=0.30\textwidth]{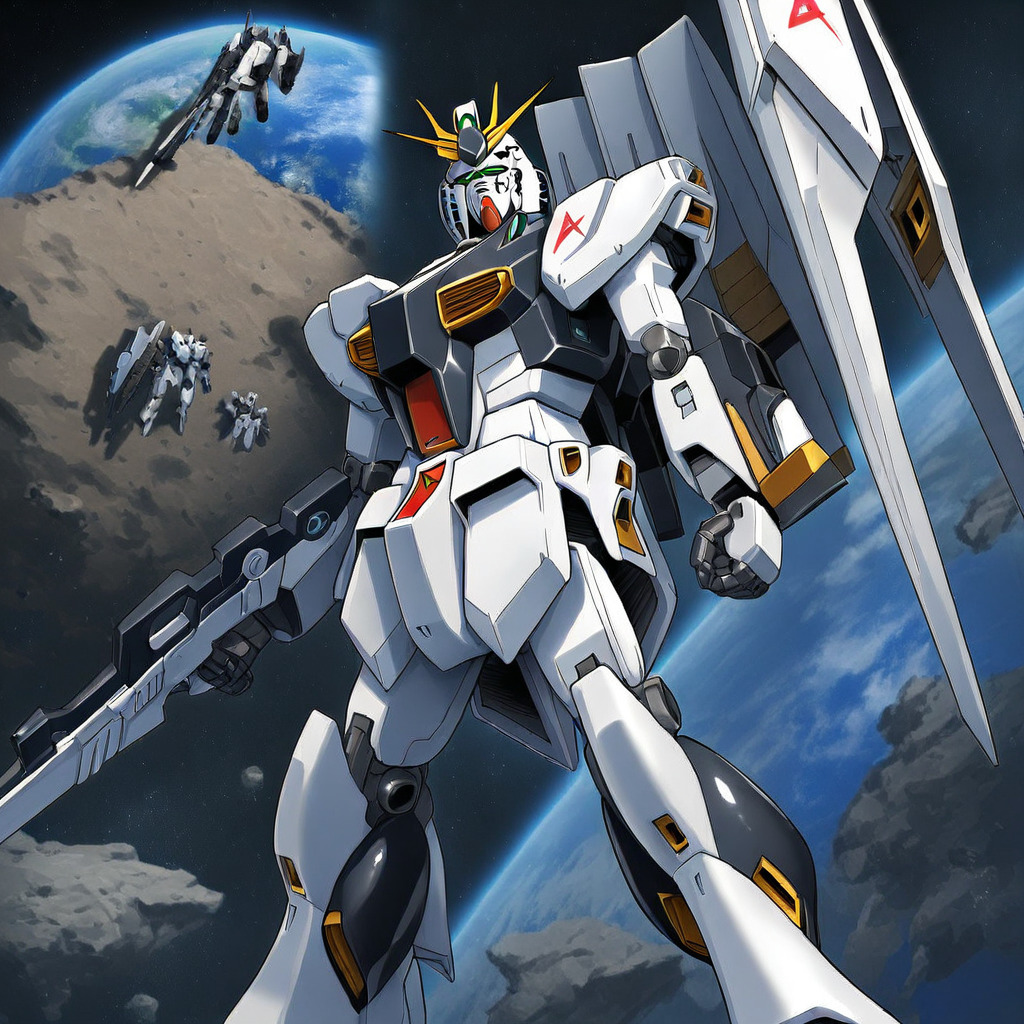} &
          \includegraphics[width=0.30\textwidth]{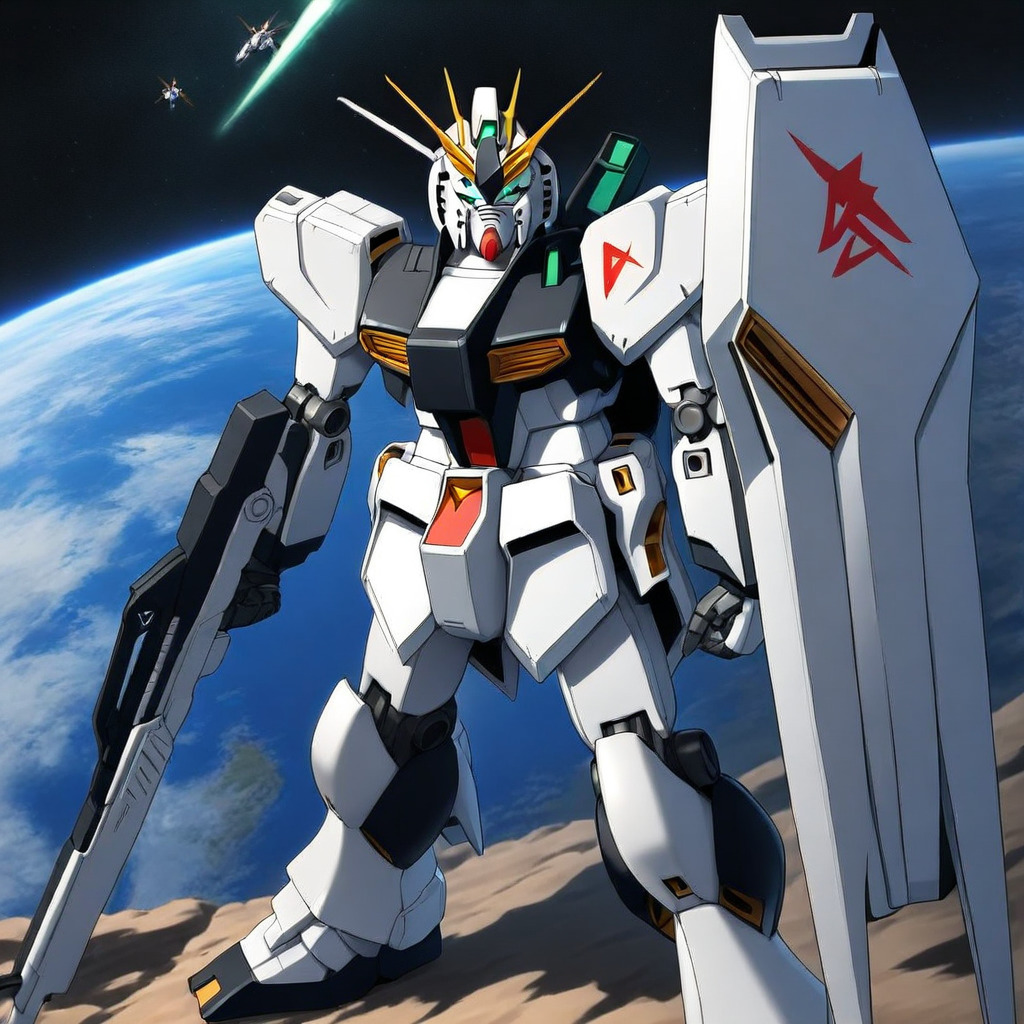} &
          \includegraphics[width=0.30\textwidth]{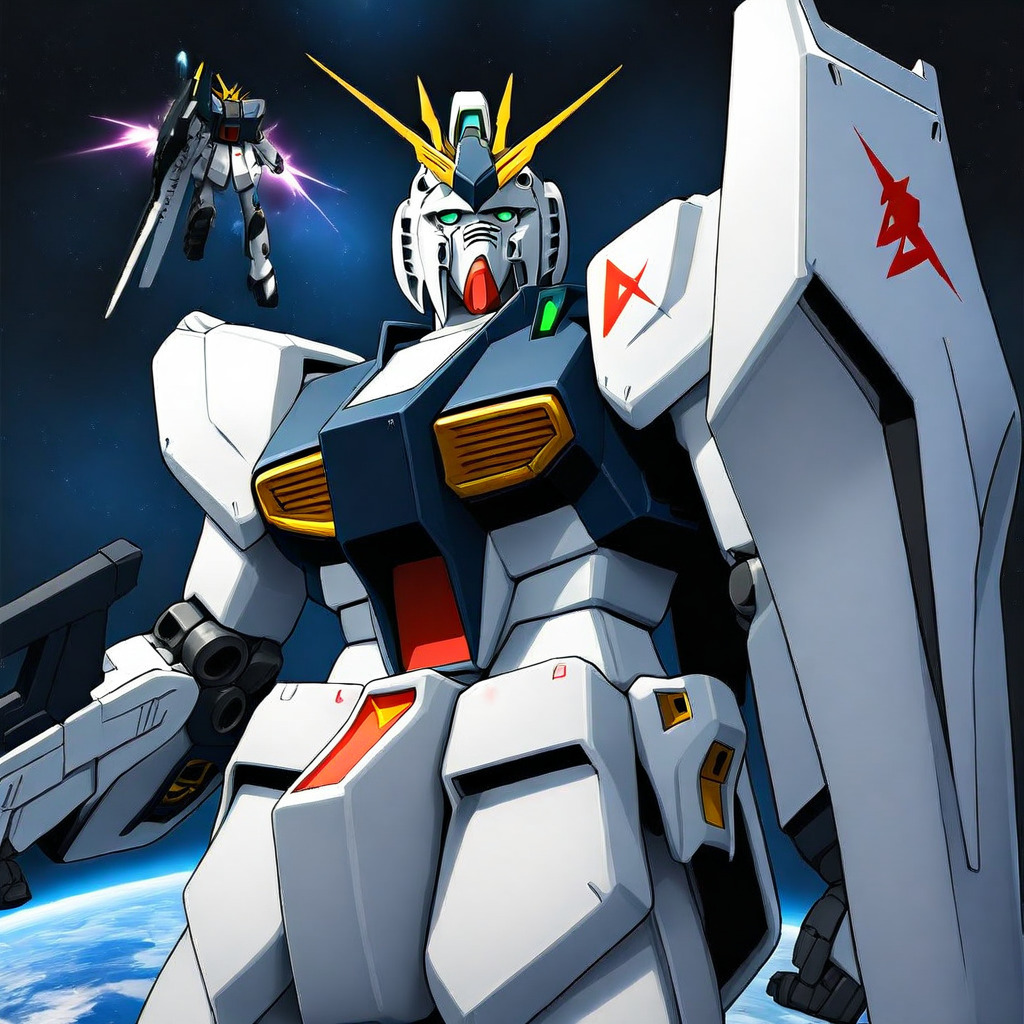} \\

          \includegraphics[width=0.30\textwidth]{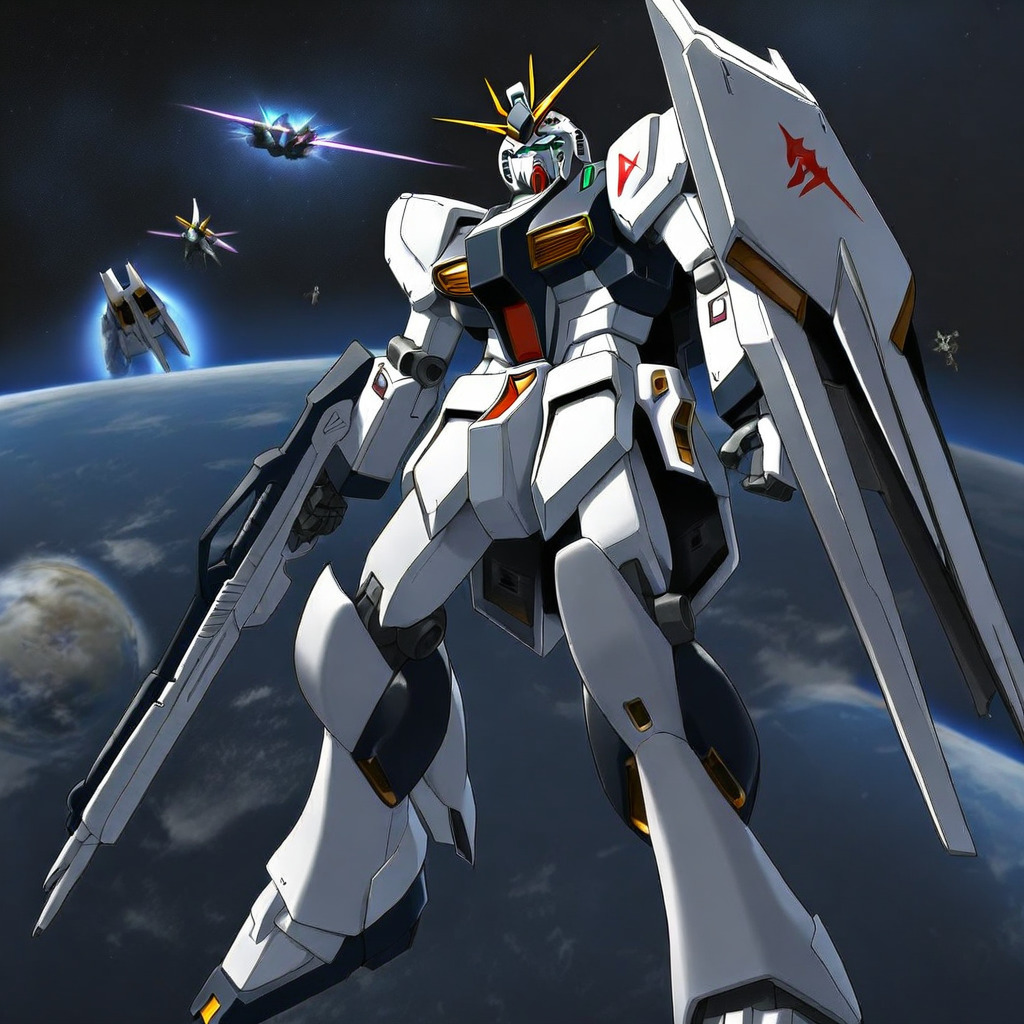} &
          \includegraphics[width=0.30\textwidth]{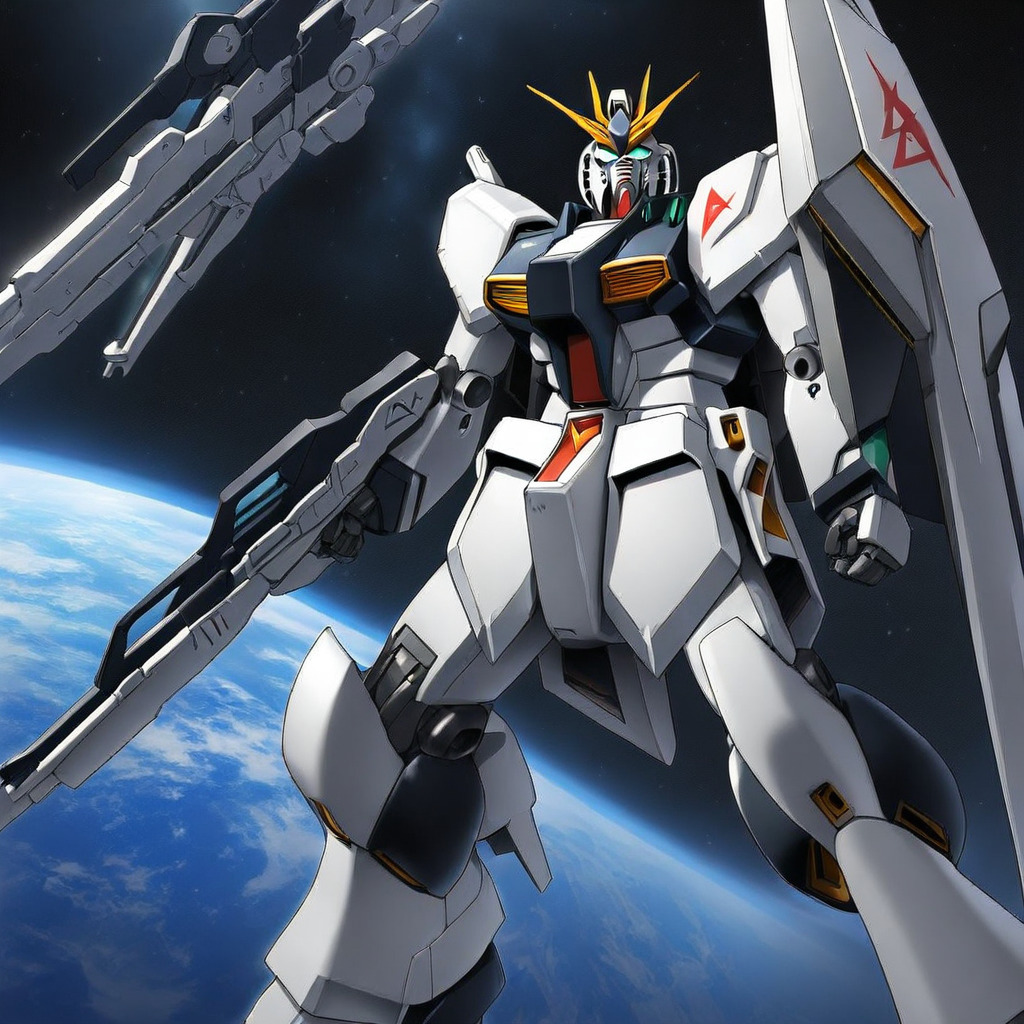} &
          \includegraphics[width=0.30\textwidth]{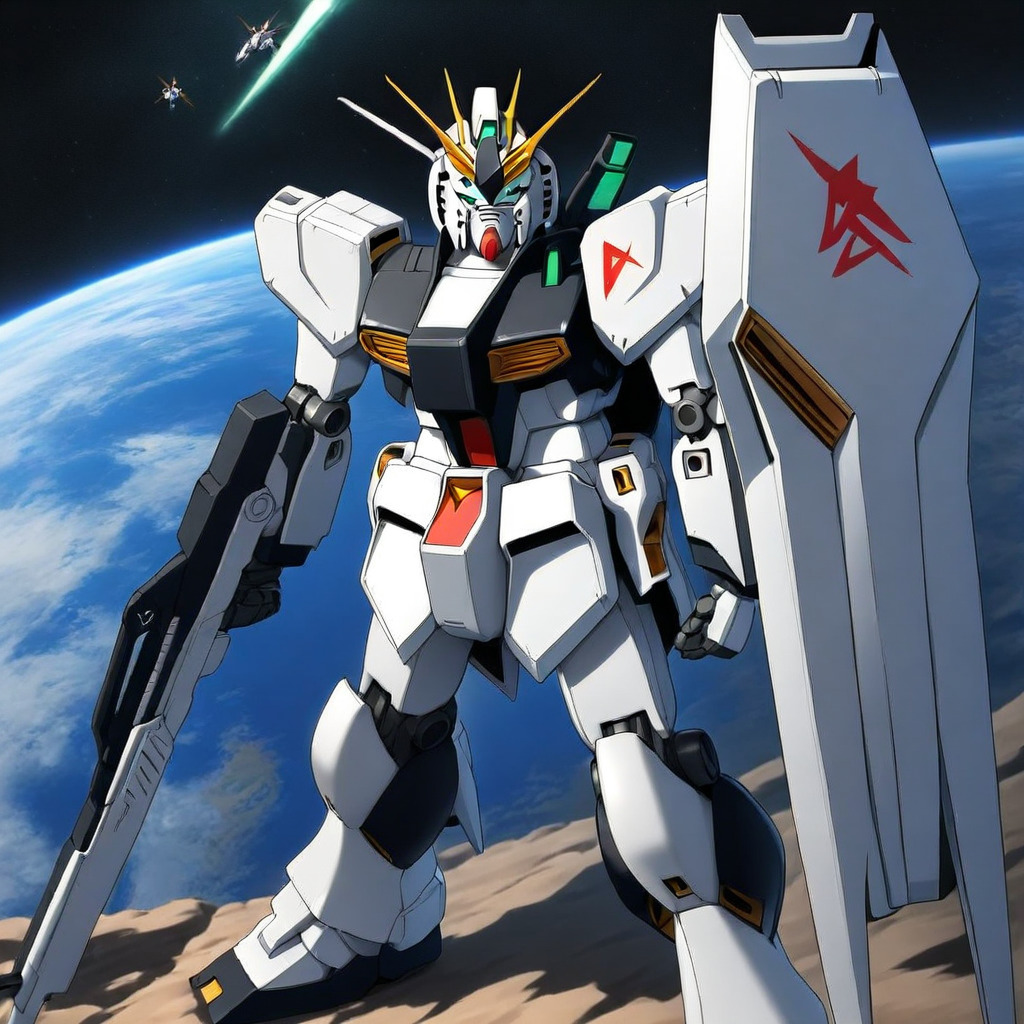} &
          \includegraphics[width=0.30\textwidth]{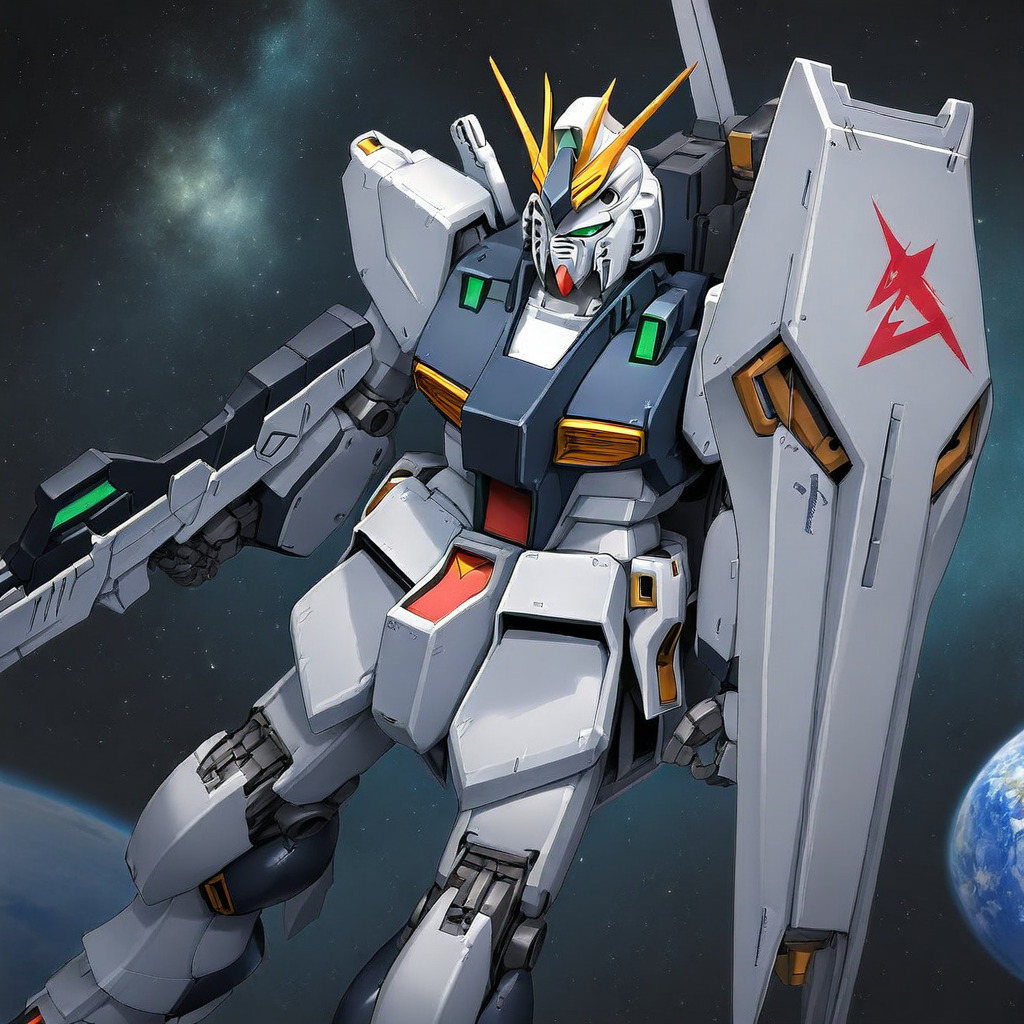} \\
          \textbf{PickScore:} 20.57 & 20.81 & \textbf{21.02} & \textit{20.95}

      \end{tabular}
      }
      \caption{Anime image comparison when integrating noise optimization methods with open-source checkpoints and LoRAs. Prompt: `nu gundam, mecha, robot, beam rifle, shield, in space, earth in background, masterpiece, best quality'.
      Source checkpoint: \cite{civitaiIllustriousV20}; LoRA: \cite{link_nu}. We \textbf{bold} and \textit{italicize} the best and second best average PickScore, respectively.} 
      \vspace{-2mm}
      \label{fig:lora_examples_illustrious_nu}
  \end{figure}

Figures~\ref{fig:lora_examples_juggernaut_woman} and \ref{fig:lora_examples_illustrious_nu} are out-of-distribution prompts from open-source LoRAs on CivitAI~\cite{Civitai}. Specifically, for each prompt, we generate $|B|=4$ images per method, and then report the mean PickScore. This gives us another facet of analysis as to exactly how each method works. Specifically, while Golden Noise continues to closely track and only slightly deviate from the image produced by the Standard Baseline random noise, NoiseAR produces a more distinct image, albeit ones that hold very little variation when $|B|>1$. In contrast, \np\ scales with $|B|$ and produces a range of higher-quality images. 

Also, we consider generation on non-uniform resolutions, such as portrait aspect ratios, i.e., $720\times1280$. In this case, \np\ is still able to score the initial noise due to the design of $\Phi_{noise}$ and we compare it to the Standard Baseline as both Golden Noise and NoiseAR are instantiated with fixed spatial resolution parameters and reshape operations which assume square-like resolutions. 

\begin{figure}[t!]
    \centering
    \setlength{\tabcolsep}{1pt}

    \begin{tabular}{@{}c@{}c@{\hspace{2pt}}c@{}c@{}}
        \multicolumn{2}{c}{Standard Baseline} &
        \multicolumn{2}{c}{\np} \\[3pt]
    
        \includegraphics[width=0.245\textwidth]{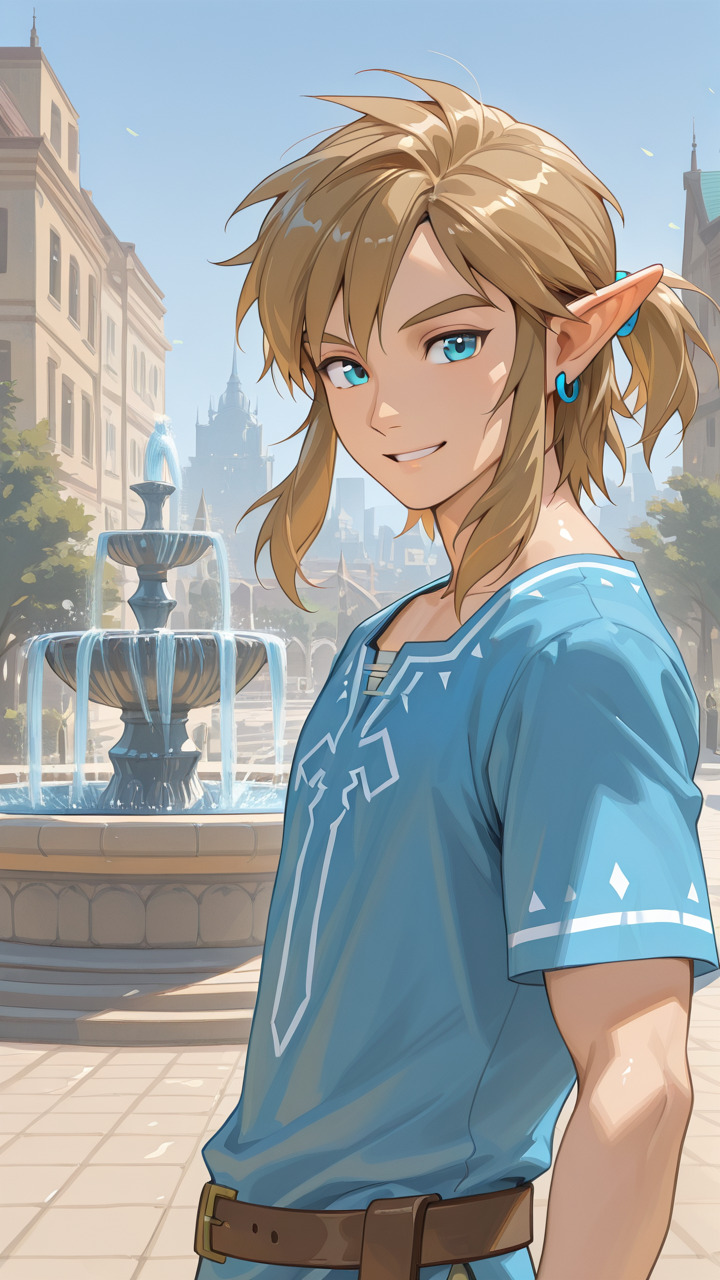} &
        \includegraphics[width=0.245\textwidth]{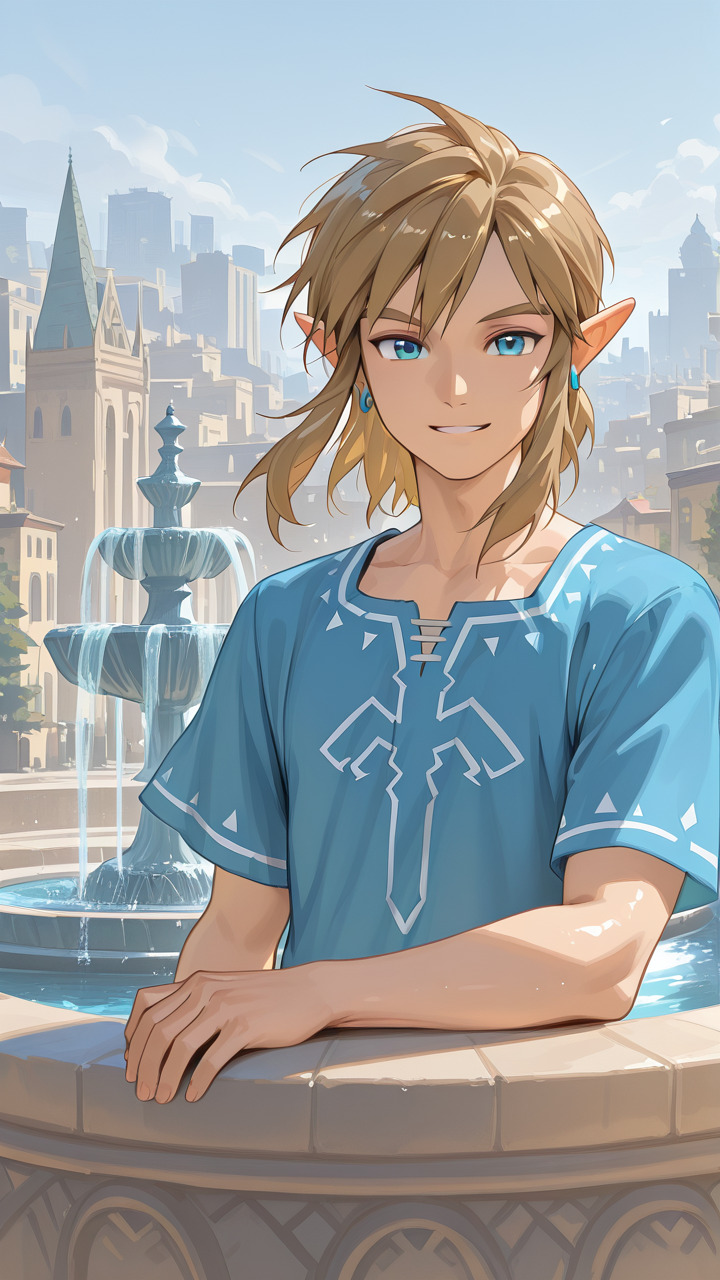} &
    
        \includegraphics[width=0.245\textwidth]{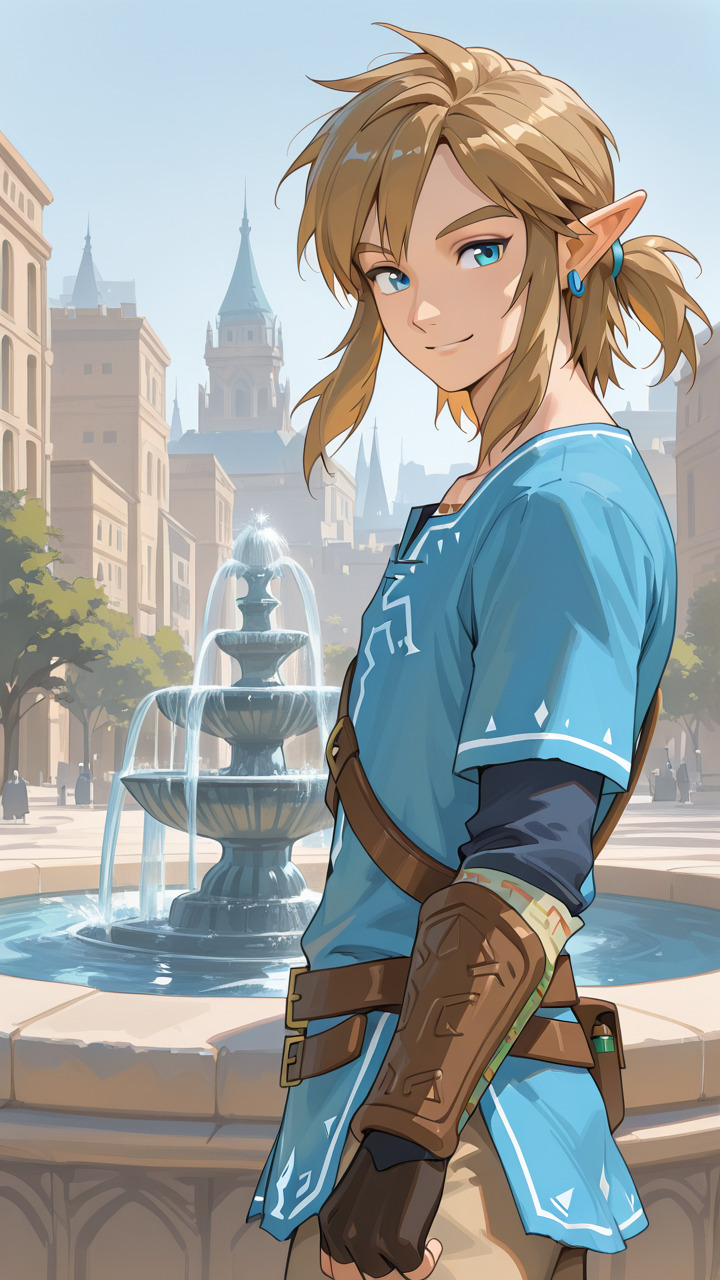} &
        \includegraphics[width=0.245\textwidth]{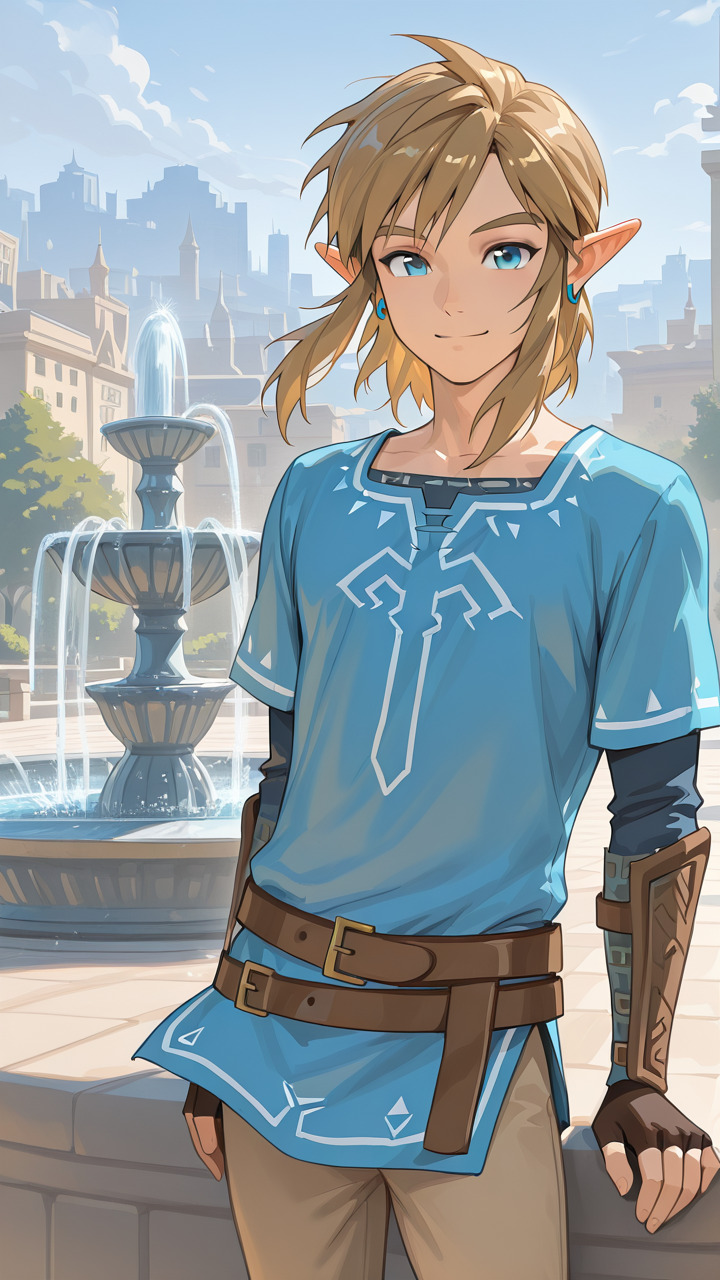} \\
    \end{tabular}

    \caption{Example $1280\times720$ images generated using open-source SDXL fine-tune~\cite{hassaku_civit} and LoRA~\cite{link_lora}. Prompt: `link, the legend of zelda, the legend of zelda: breath of the wild, 1boy, outdoors, male focus, blue eyes, blue shirt, light brown hair, shiny hair, blue earring, looking at viewer, smiling, posing, male focus, city, fountain, natural light, beautiful light,wind, pointy ears, painterly, amazing quality'}
    \label{fig:lora_examples_hassaku_link}
\end{figure}

Figure~\ref{fig:lora_examples_hassaku_link} provides a sample visualization. %
The LoRA is specialized for `Link' from 
\textit{The Legend of Zelda: Breath of the Wild} (2017). Notice how the images produced by \np\ more accurately capture Link's outfit by including his gauntlets and belts; we refer interested readers to official artwork~\cite{botw-creating-a-champion} for reference. %

\section{Additional Visualizations}
\label{sec:supp_examples}

\begin{figure}[t!]
    \centering
    \setlength{\tabcolsep}{1.2pt}
    \renewcommand{\arraystretch}{1.2}
    \footnotesize
    \scalebox{\scaleboxratio}{
      \begin{tabular}{@{}c@{\hspace{3pt}}c@{\hspace{3pt}}c@{\hspace{3pt}}c@{}}
          Standard & Golden Noise & NoiseAR & \np \\[2pt]

          \includegraphics[width=0.30\textwidth]{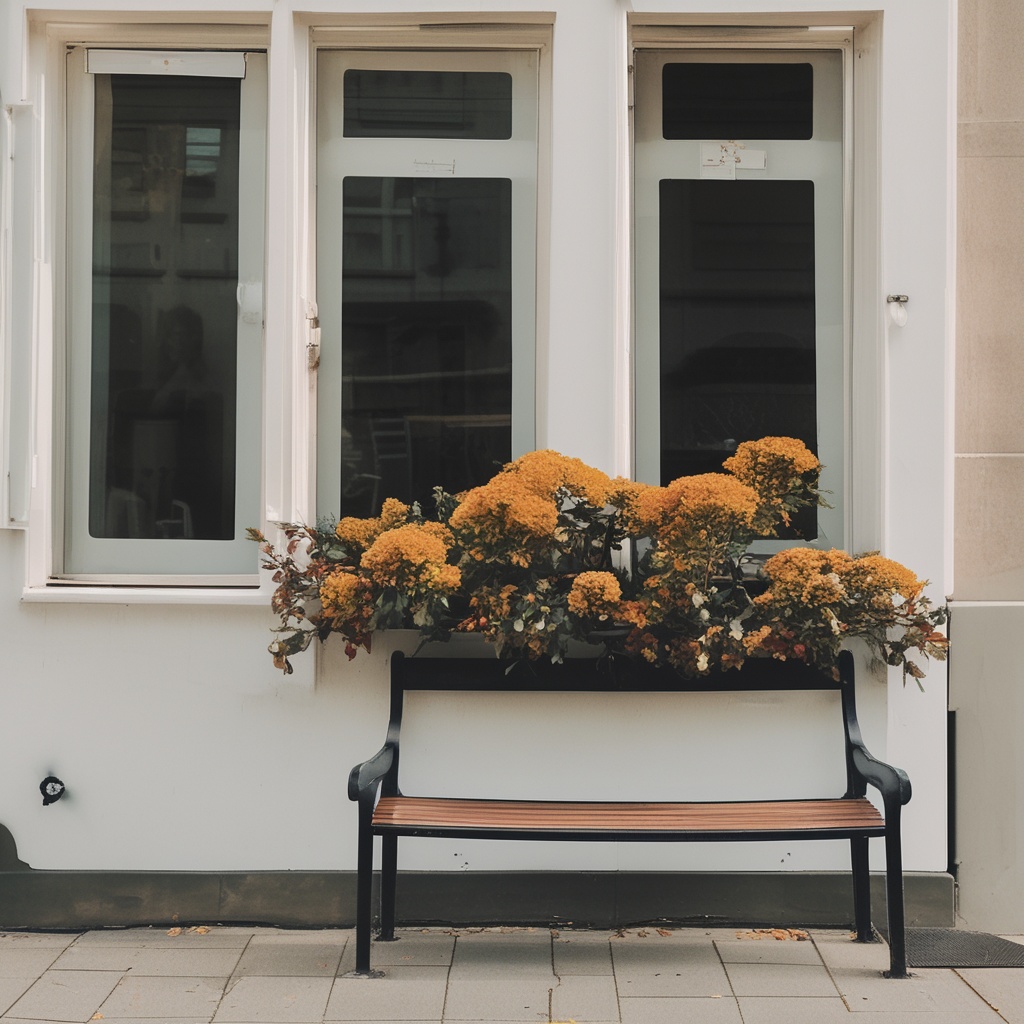} &
          \includegraphics[width=0.30\textwidth]{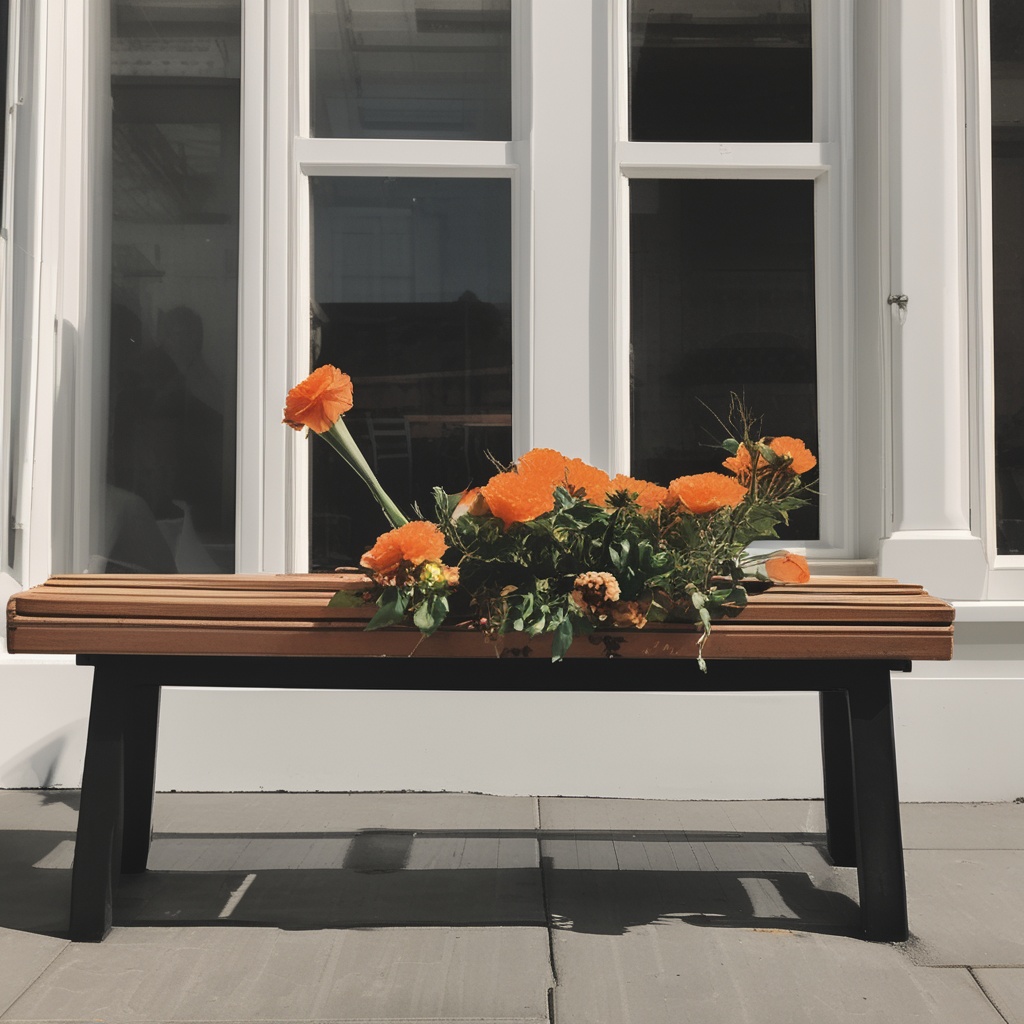} &
          \includegraphics[width=0.30\textwidth]{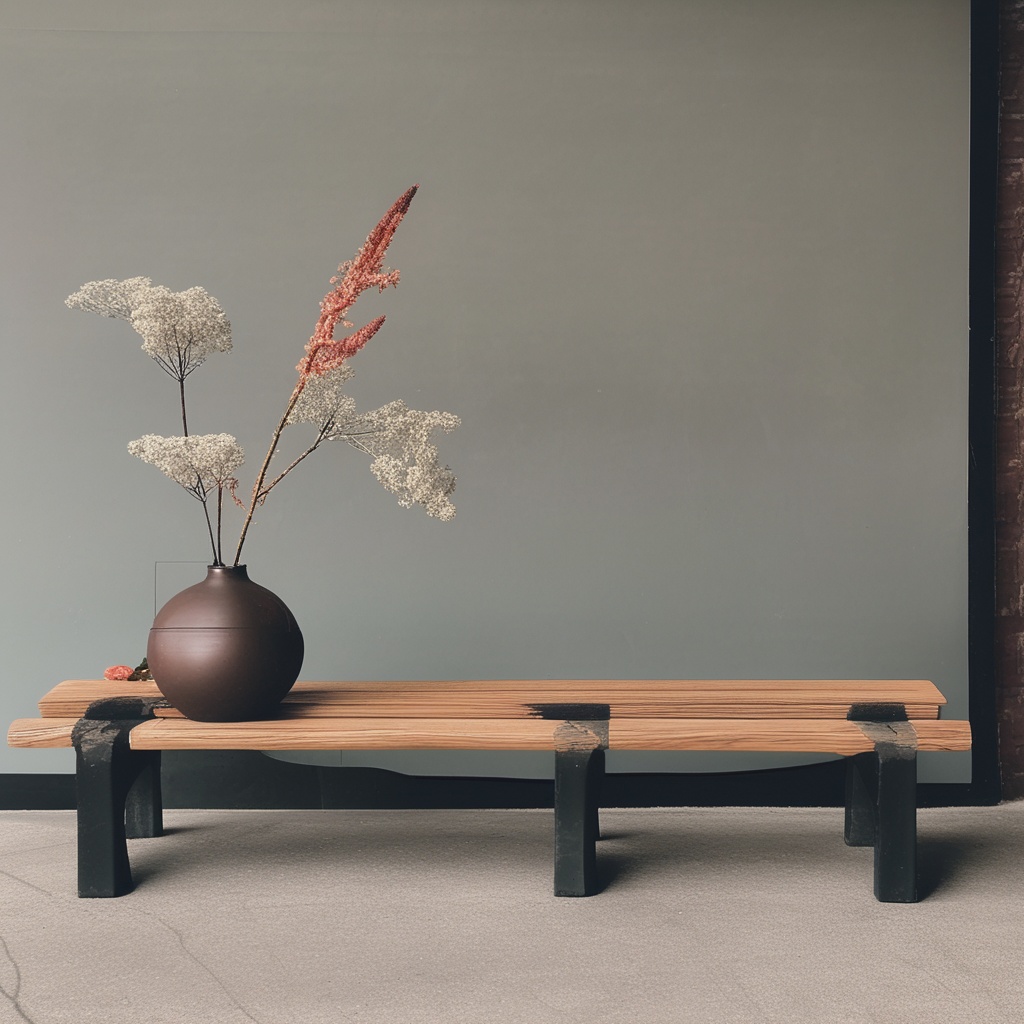} &
          \includegraphics[width=0.30\textwidth]{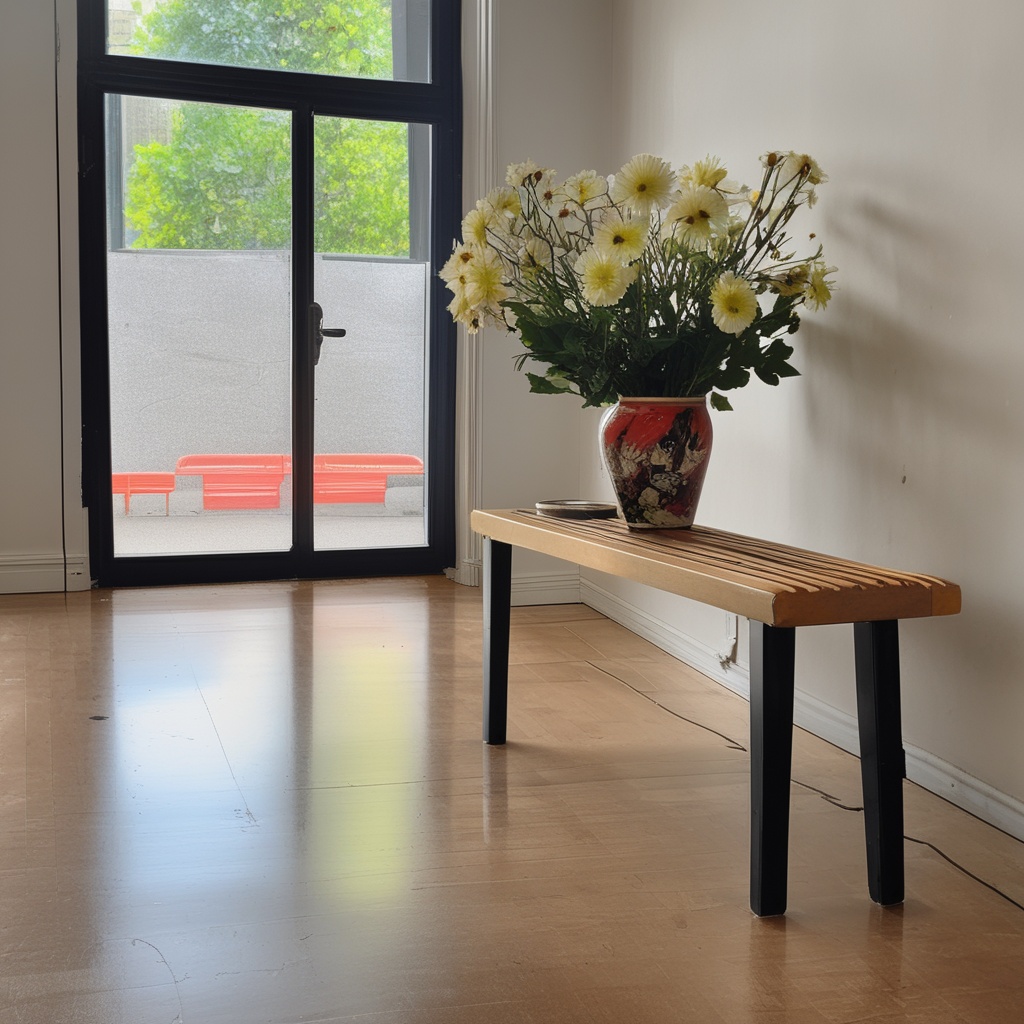} \\
          \multicolumn{4}{@{}l@{}}{SDXL $\times$ GenEval: ``a photo of a bench and a vase''} \\[3pt]

          \includegraphics[width=0.30\textwidth]{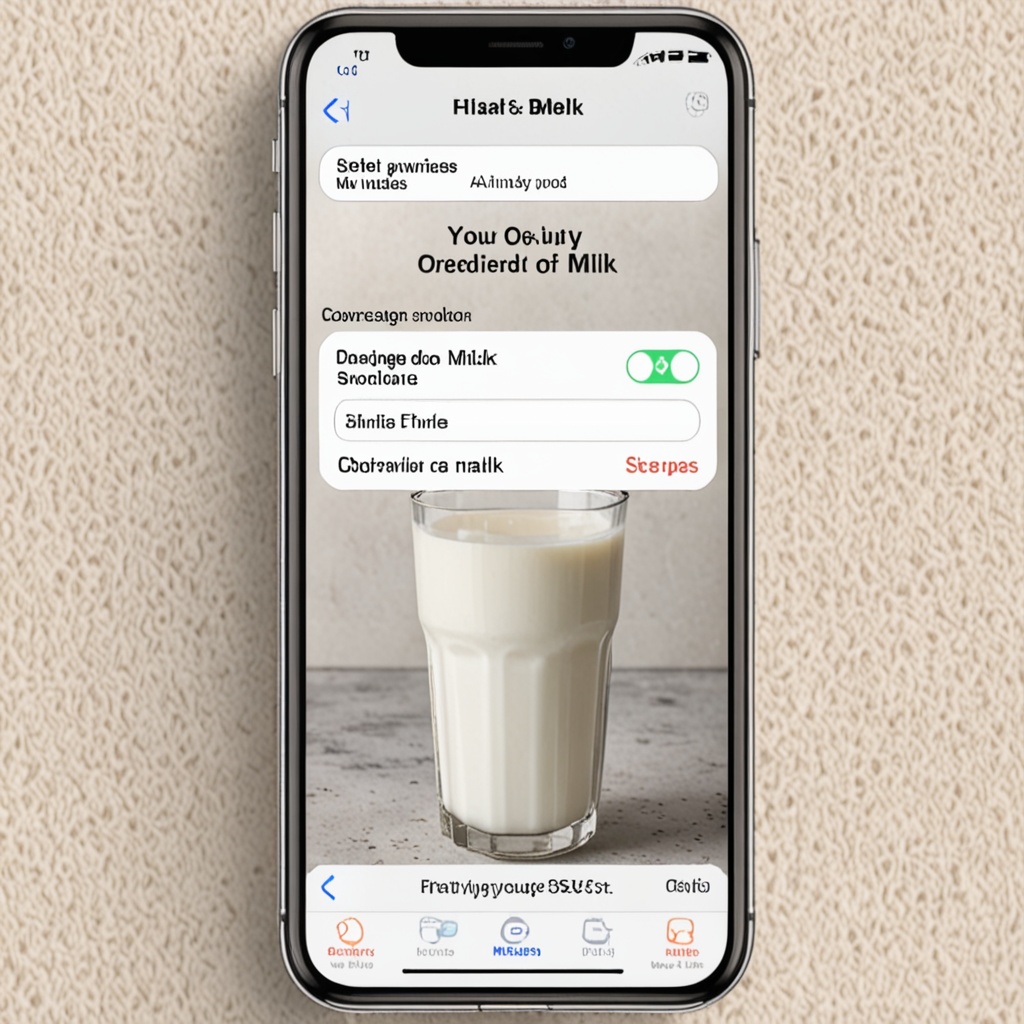} &
          \includegraphics[width=0.30\textwidth]{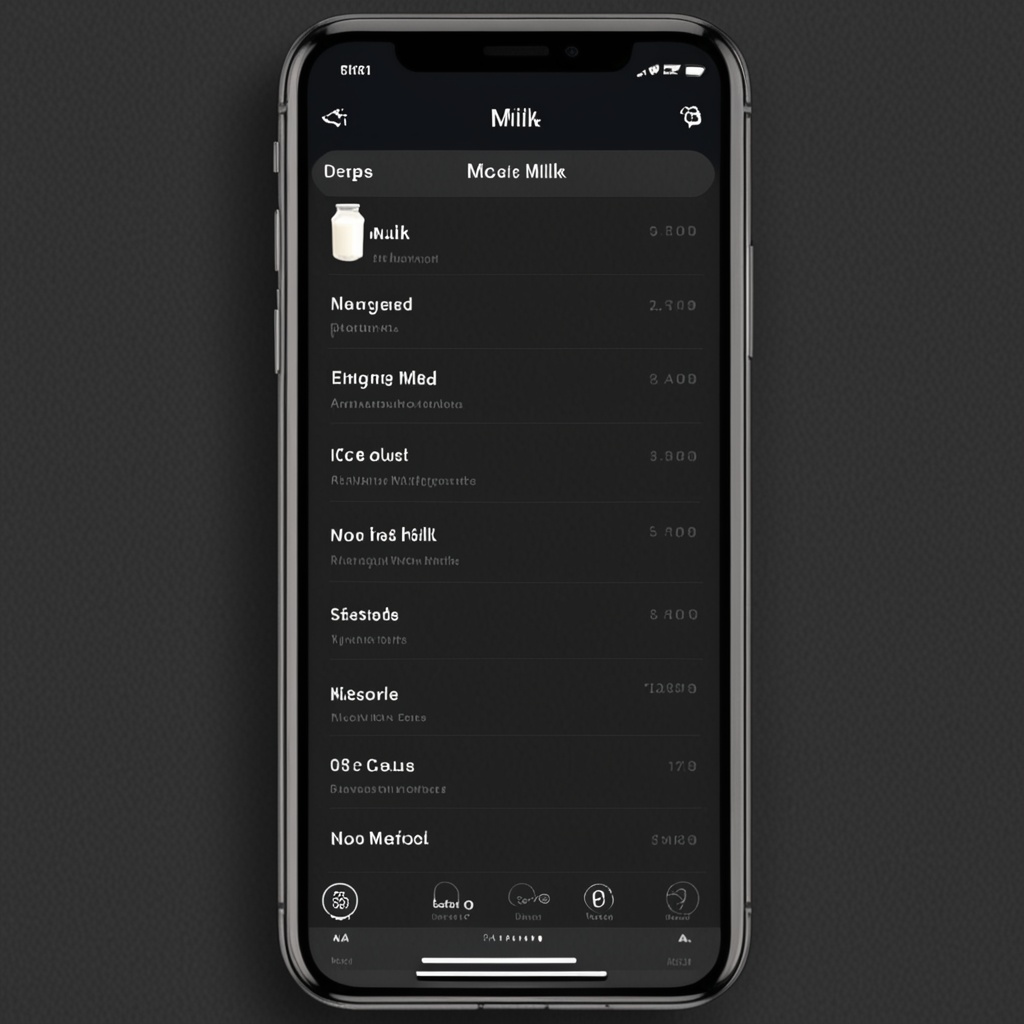} &
          \includegraphics[width=0.30\textwidth]{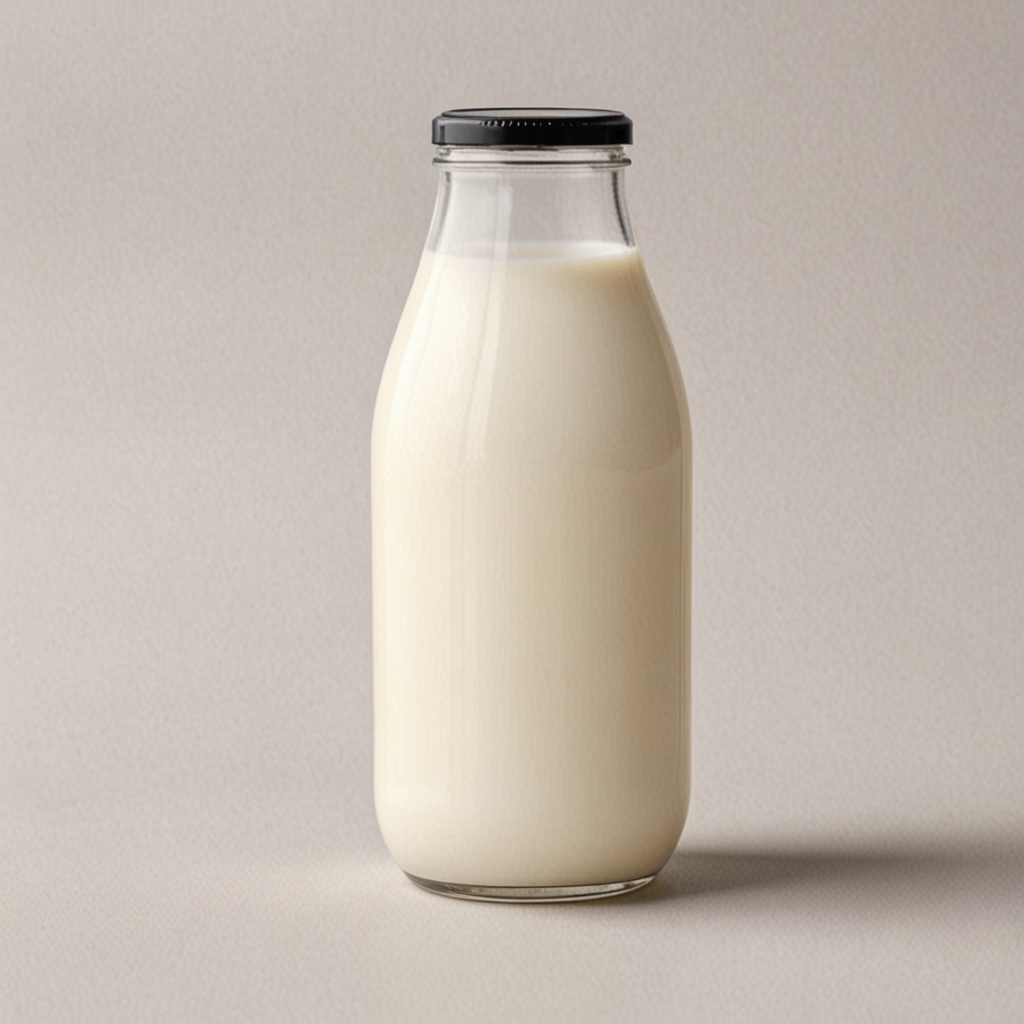} &
          \includegraphics[width=0.30\textwidth]{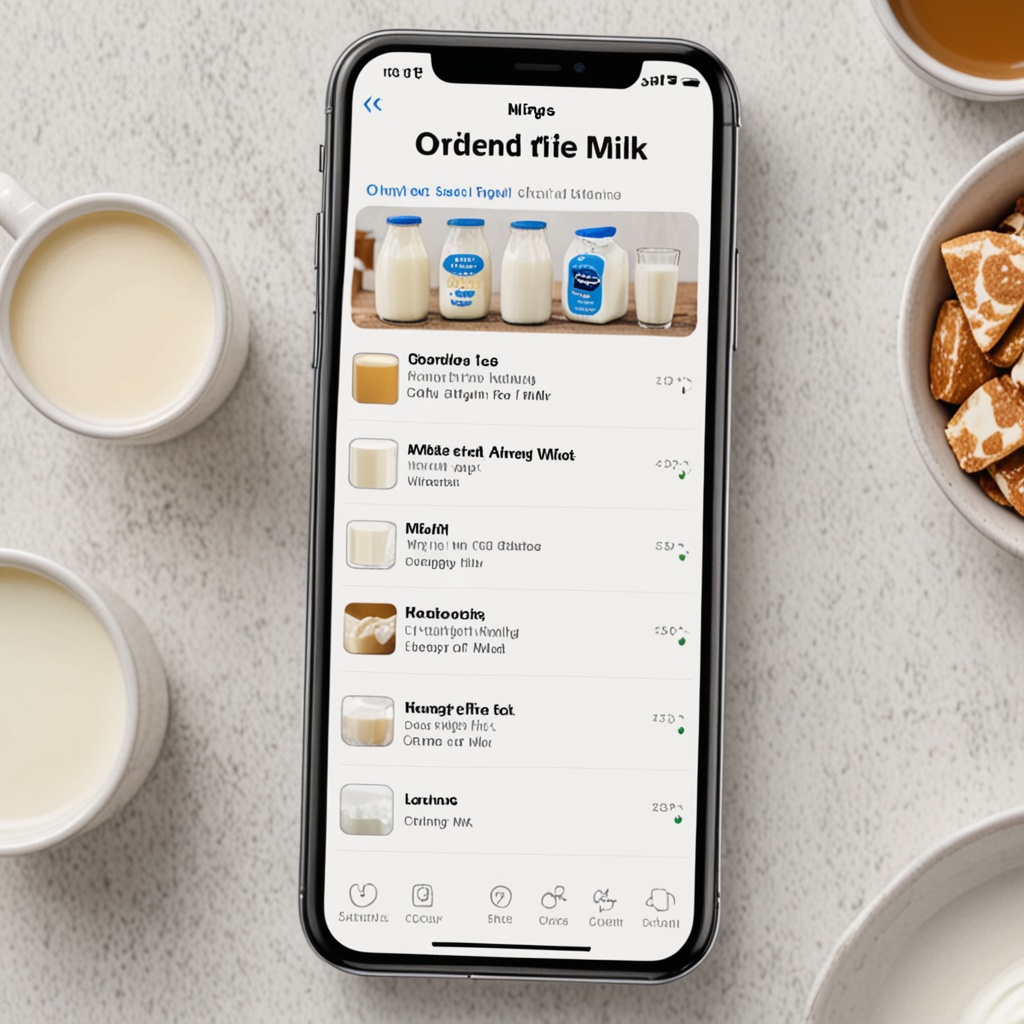} \\
          \multicolumn{4}{@{}p{0.99\textwidth}@{}}{DreamShaper $\times$ DrawBench: ``A screenshot of an iOS app for ordering different types of milk''} \\[3pt]

          \includegraphics[width=0.30\textwidth]{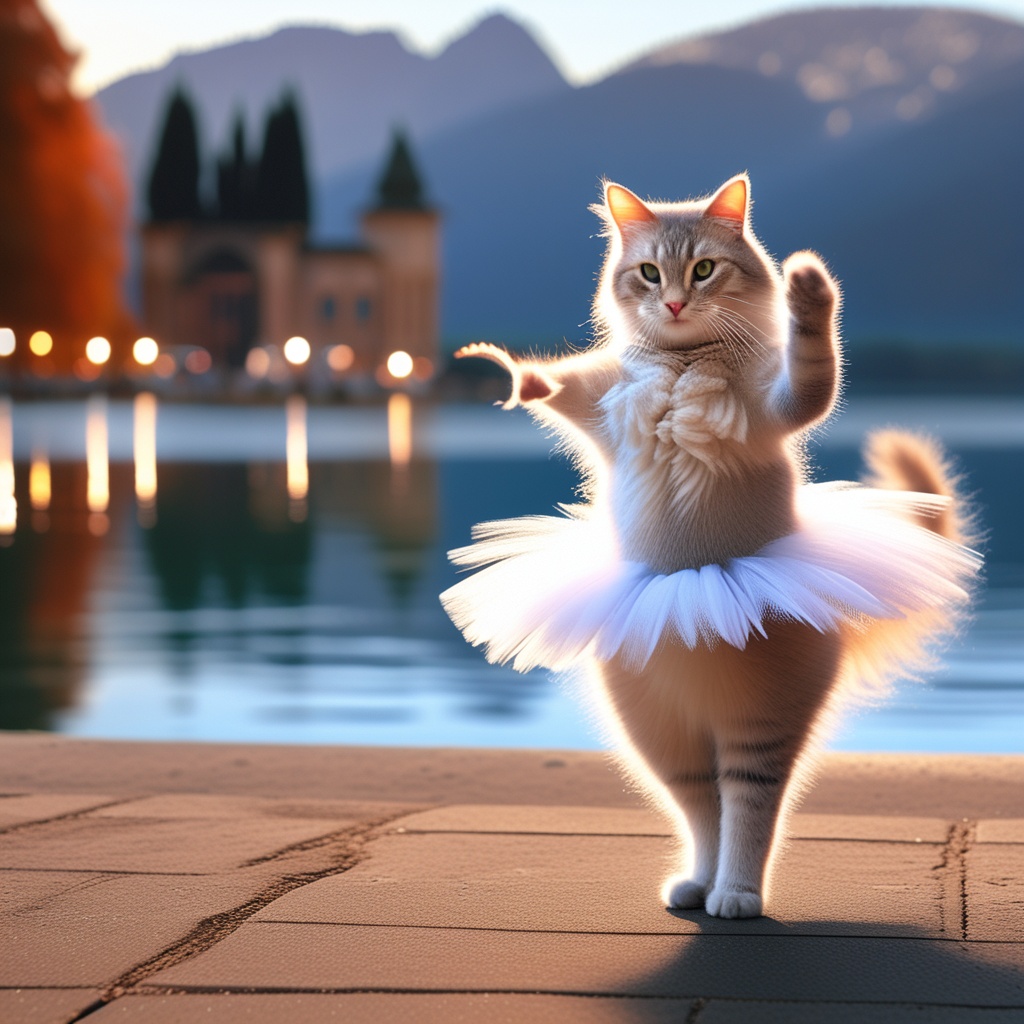} &
          \includegraphics[width=0.30\textwidth]{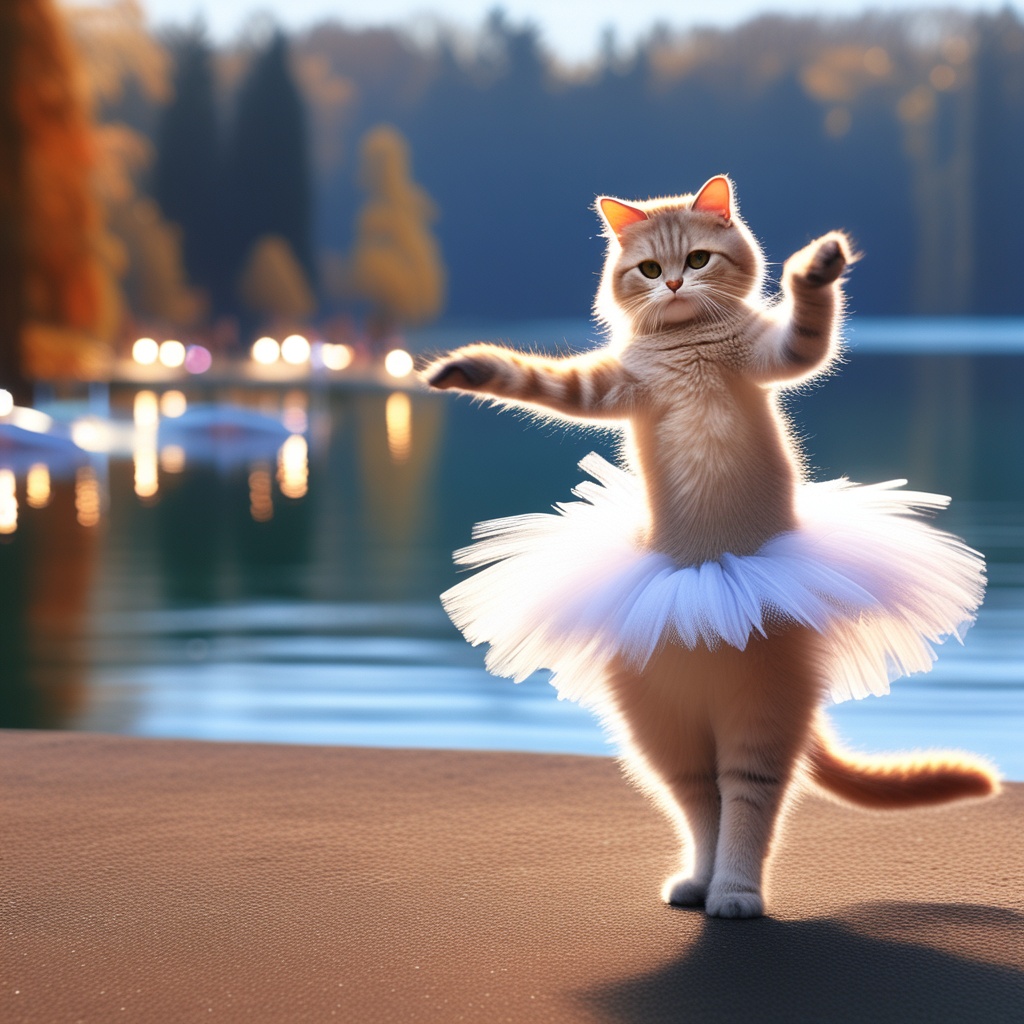} &
          \includegraphics[width=0.30\textwidth]{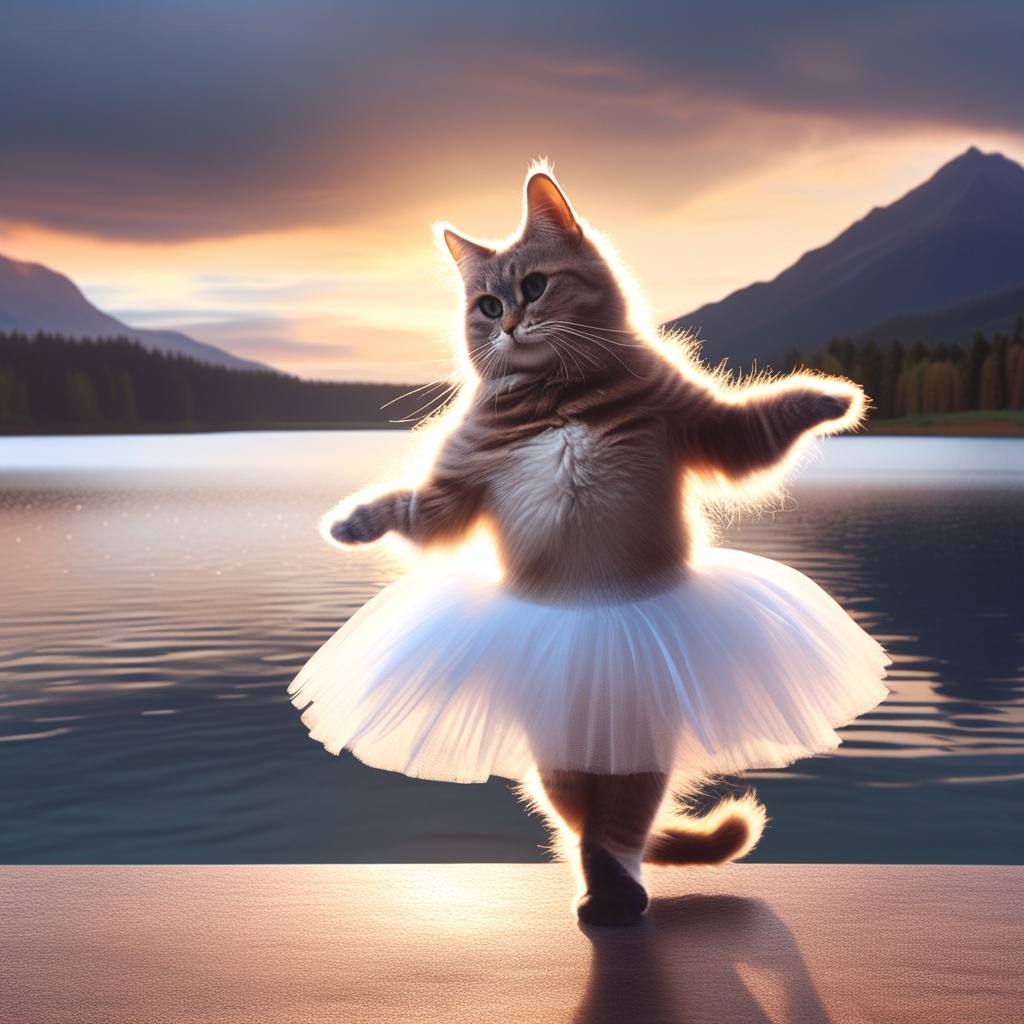} &
          \includegraphics[width=0.30\textwidth]{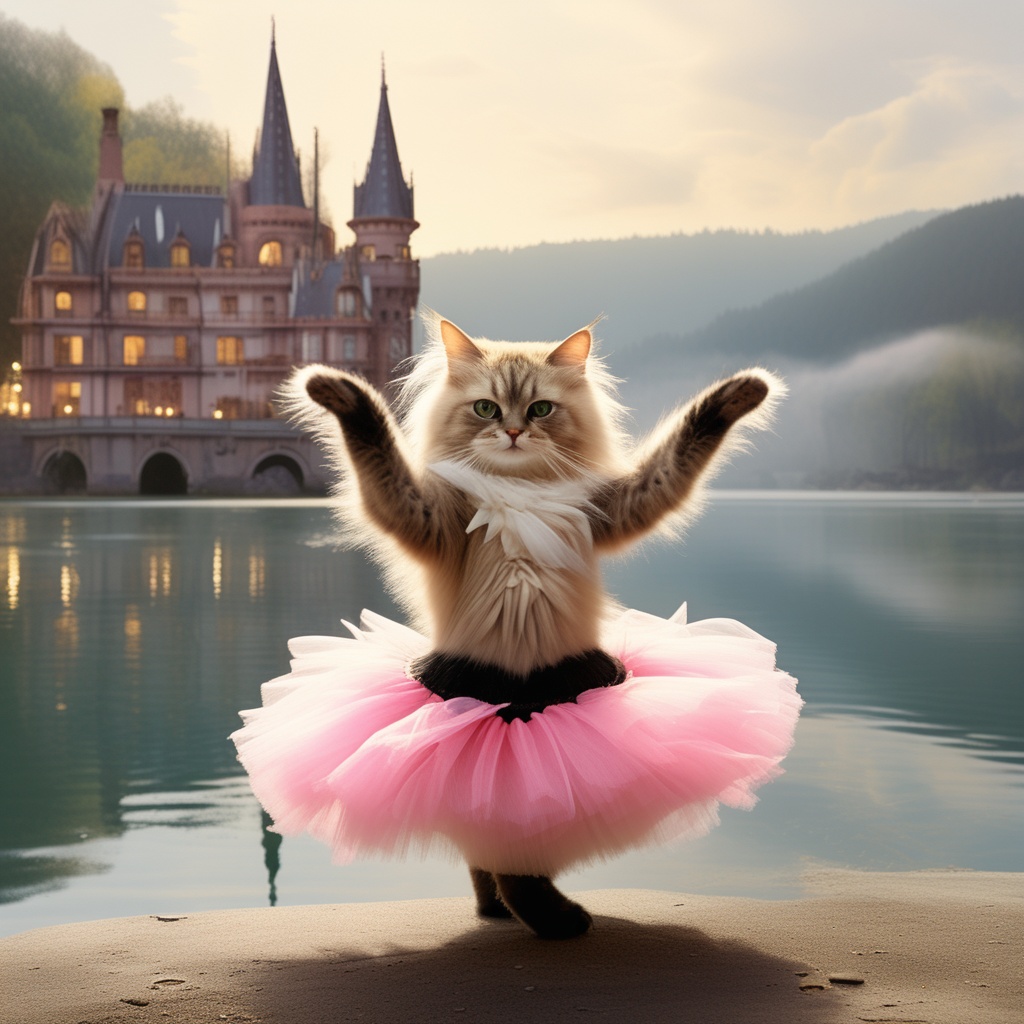} \\
          \multicolumn{4}{@{}l@{}}{Hunyuan-DiT $\times$ Pick-a-Pic: ``A cat in a tutu dancing to Swan Lake''} \\[3pt]

          \includegraphics[width=0.30\textwidth]{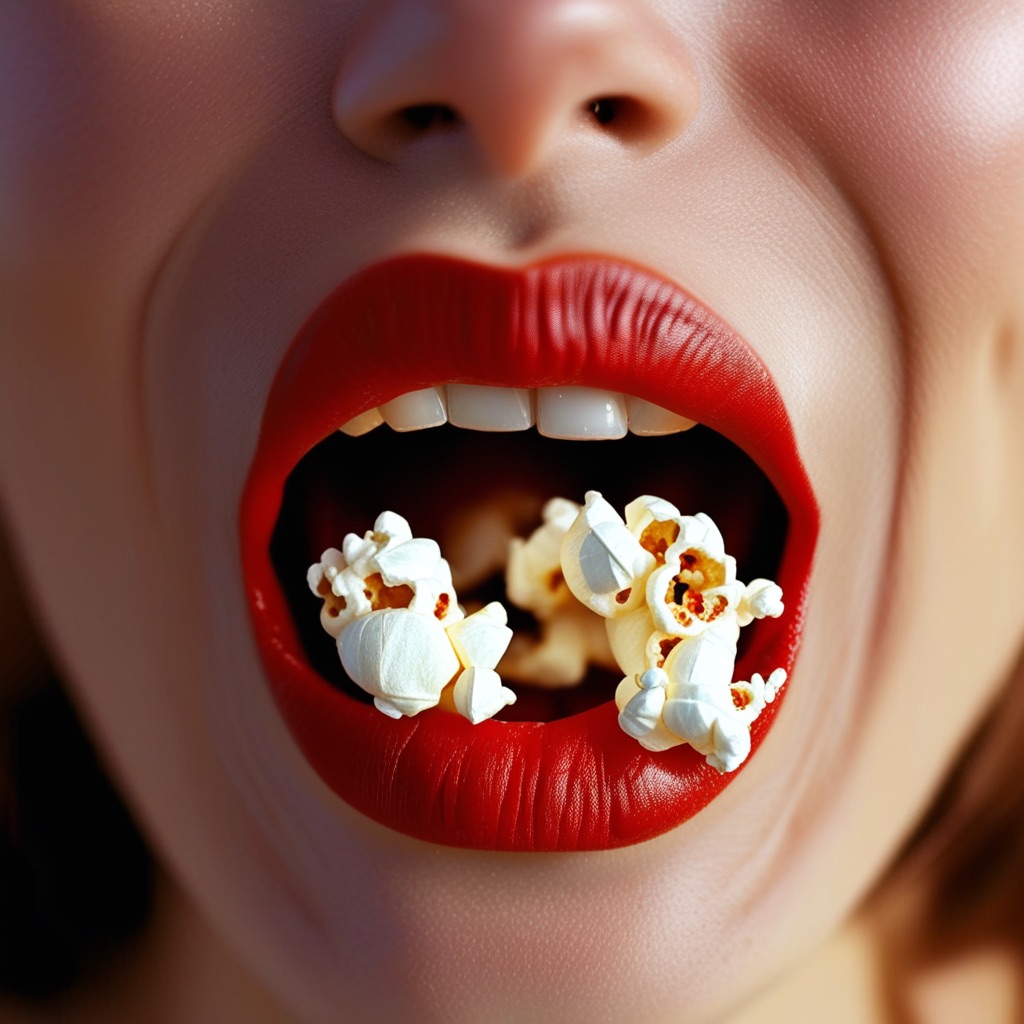} &
          \includegraphics[width=0.30\textwidth]{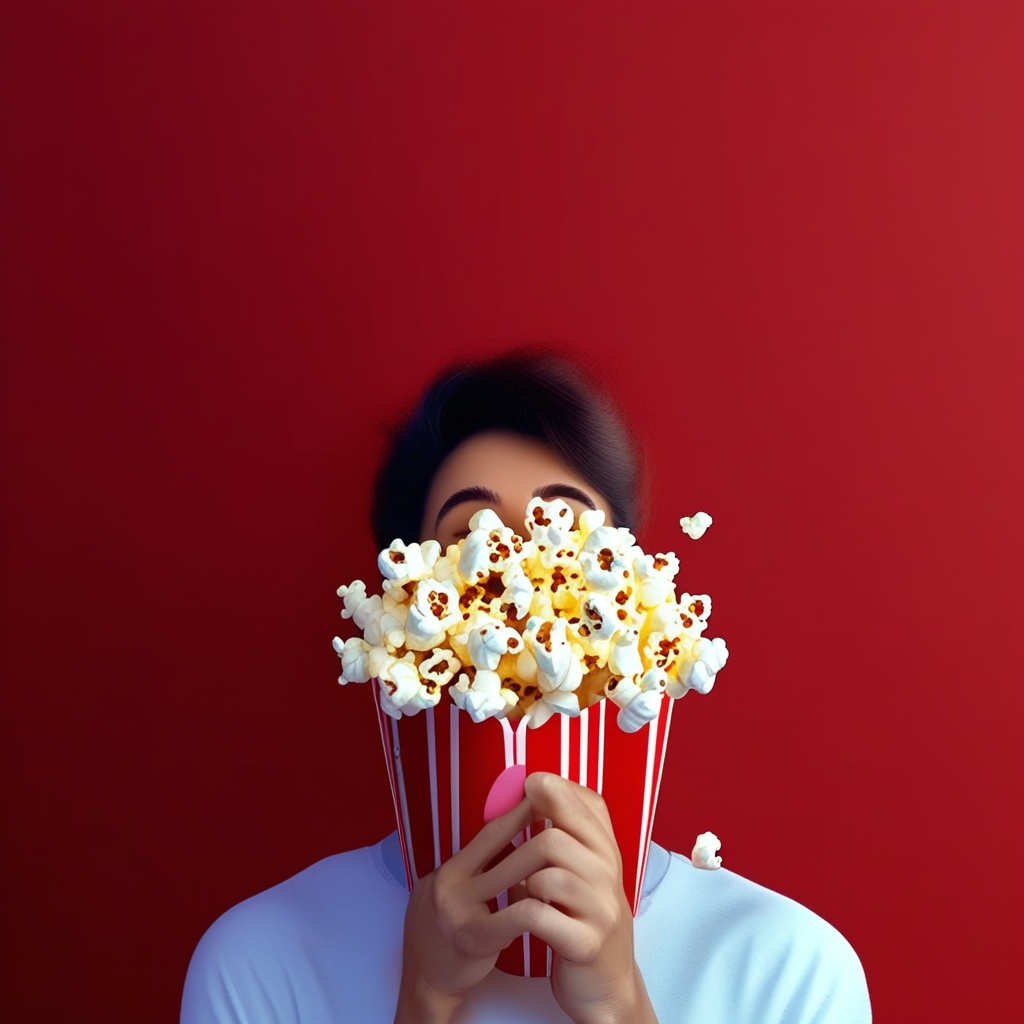} &
          \includegraphics[width=0.30\textwidth]{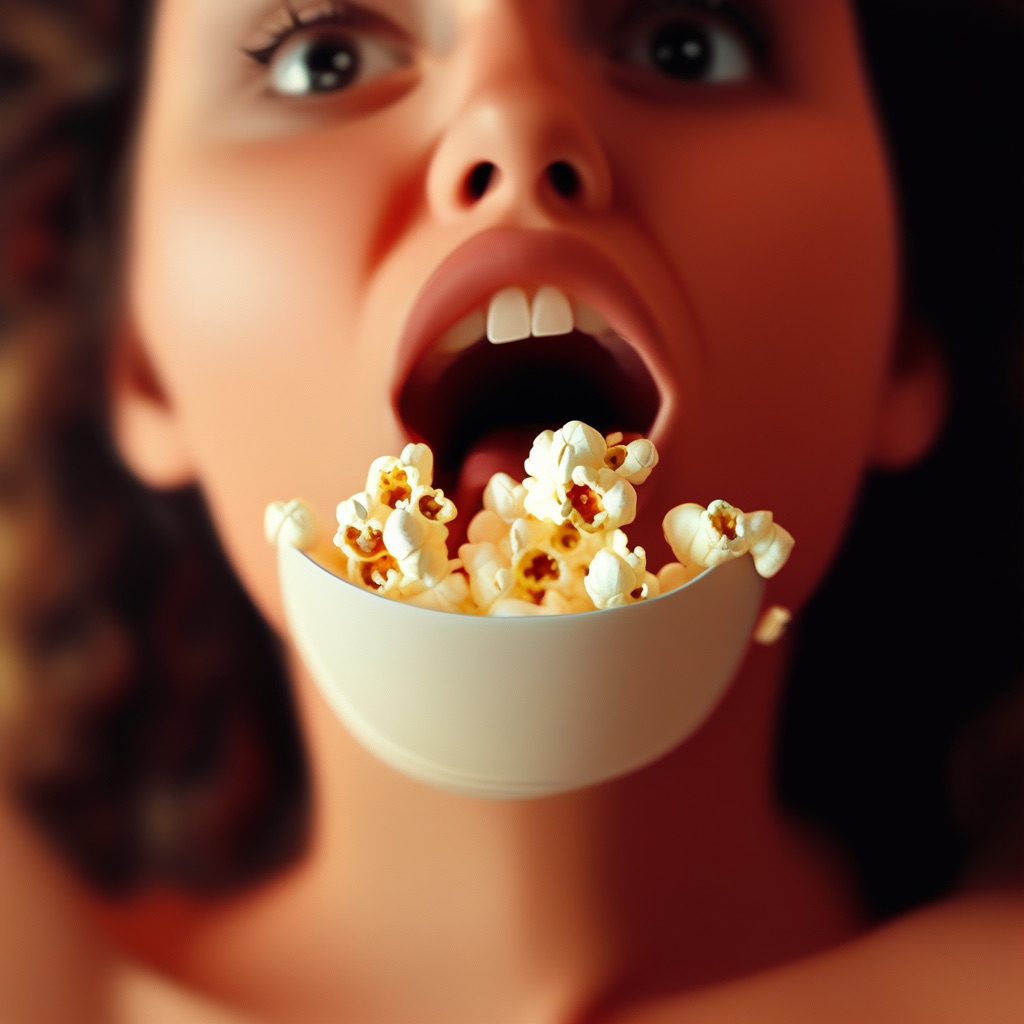} &
          \includegraphics[width=0.30\textwidth]{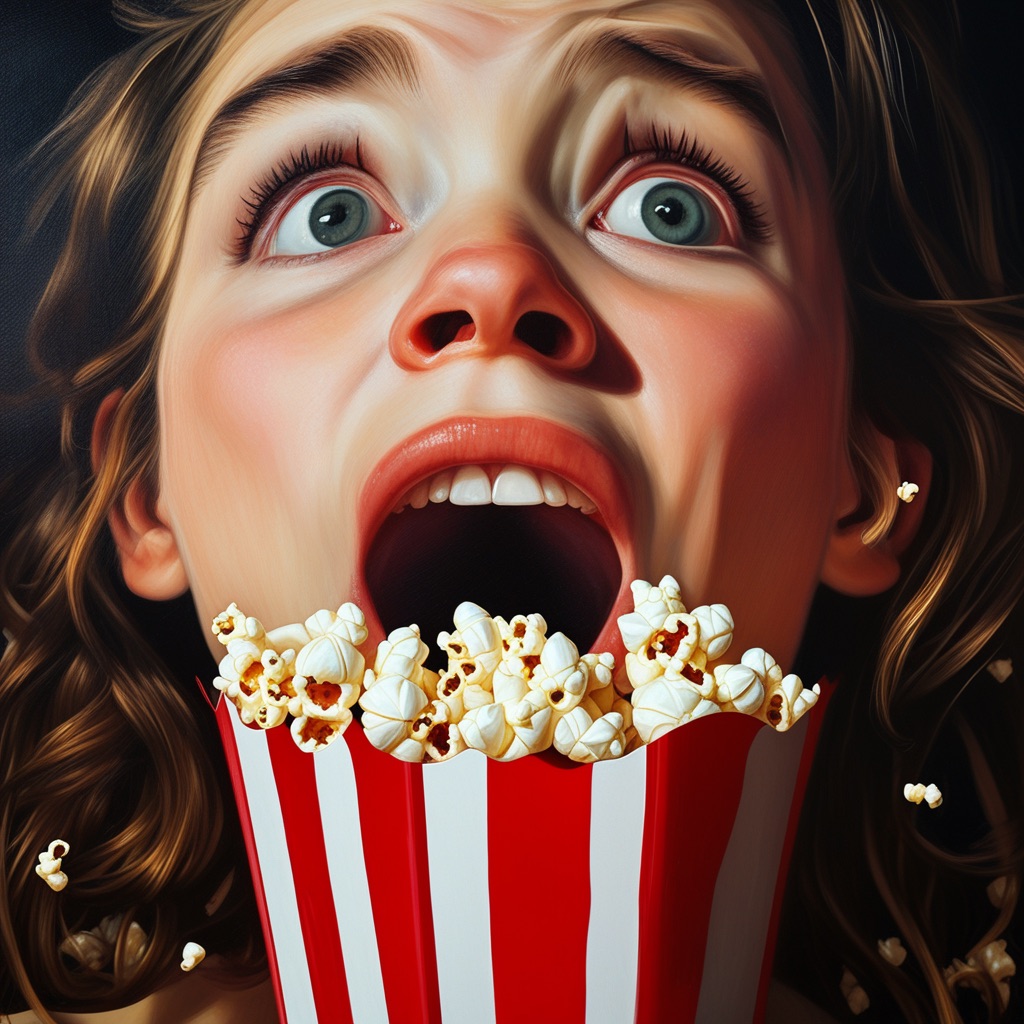} \\
          \multicolumn{4}{@{}p{0.99\textwidth}@{}}{PixArt-$\Sigma$ $\times$ HPDv2: ``Popcorn in mouth''} \\[3pt]

      \end{tabular}
      }
      \caption{Additional qualitative examples comparing \np\ to Golden Noise, NoiseAR and the standard baseline. DM and prompt details provided below each row.}
      \label{fig:qual_examples}
  \end{figure}

\begin{figure}[t!]
    \centering
    \setlength{\tabcolsep}{1.2pt}
    \renewcommand{\arraystretch}{1.2}
    \footnotesize
    \scalebox{\scaleboxratio}{
      \begin{tabular}{@{}c@{\hspace{3pt}}c@{\hspace{3pt}}c@{\hspace{3pt}}c@{}}
          Standard & Golden Noise & NoiseAR & \np \\[2pt]

          \includegraphics[width=0.30\textwidth]{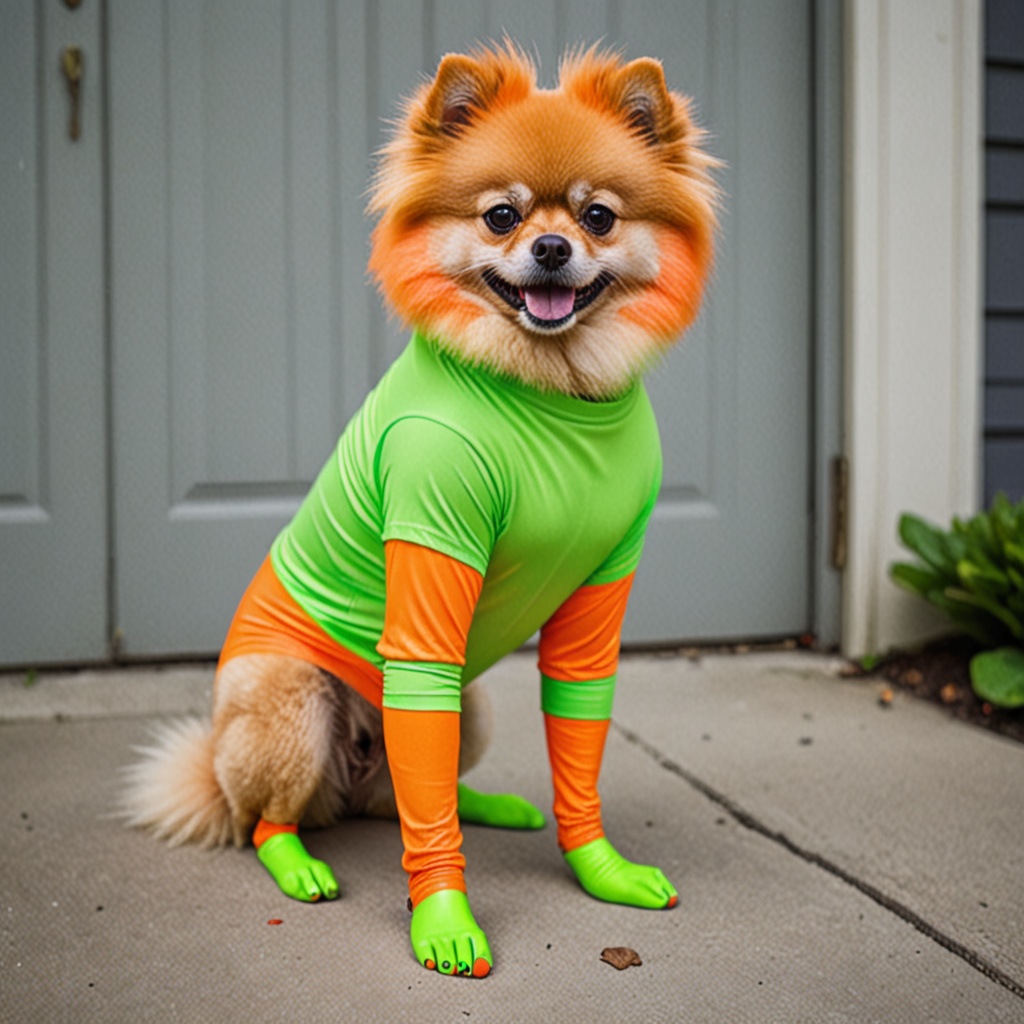} &
          \includegraphics[width=0.30\textwidth]{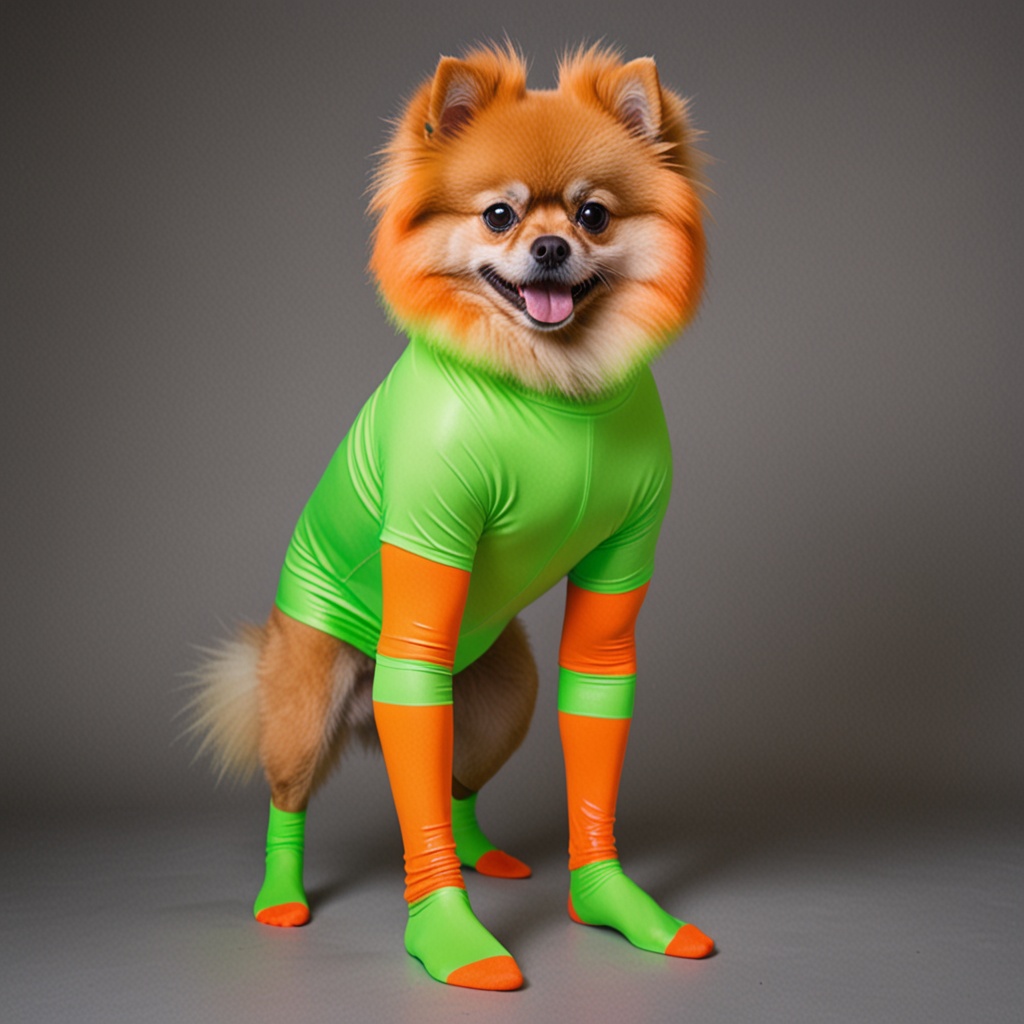} &
          \includegraphics[width=0.30\textwidth]{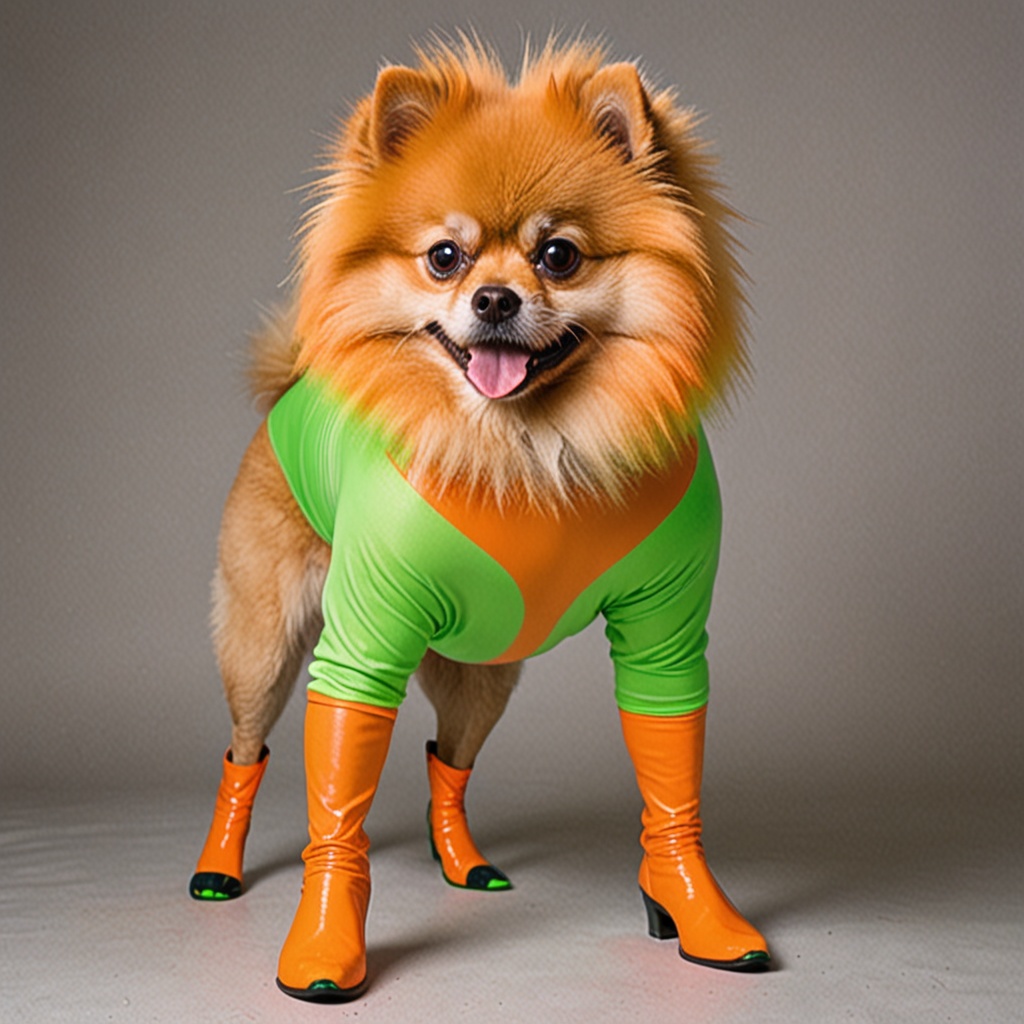} &
          \includegraphics[width=0.30\textwidth]{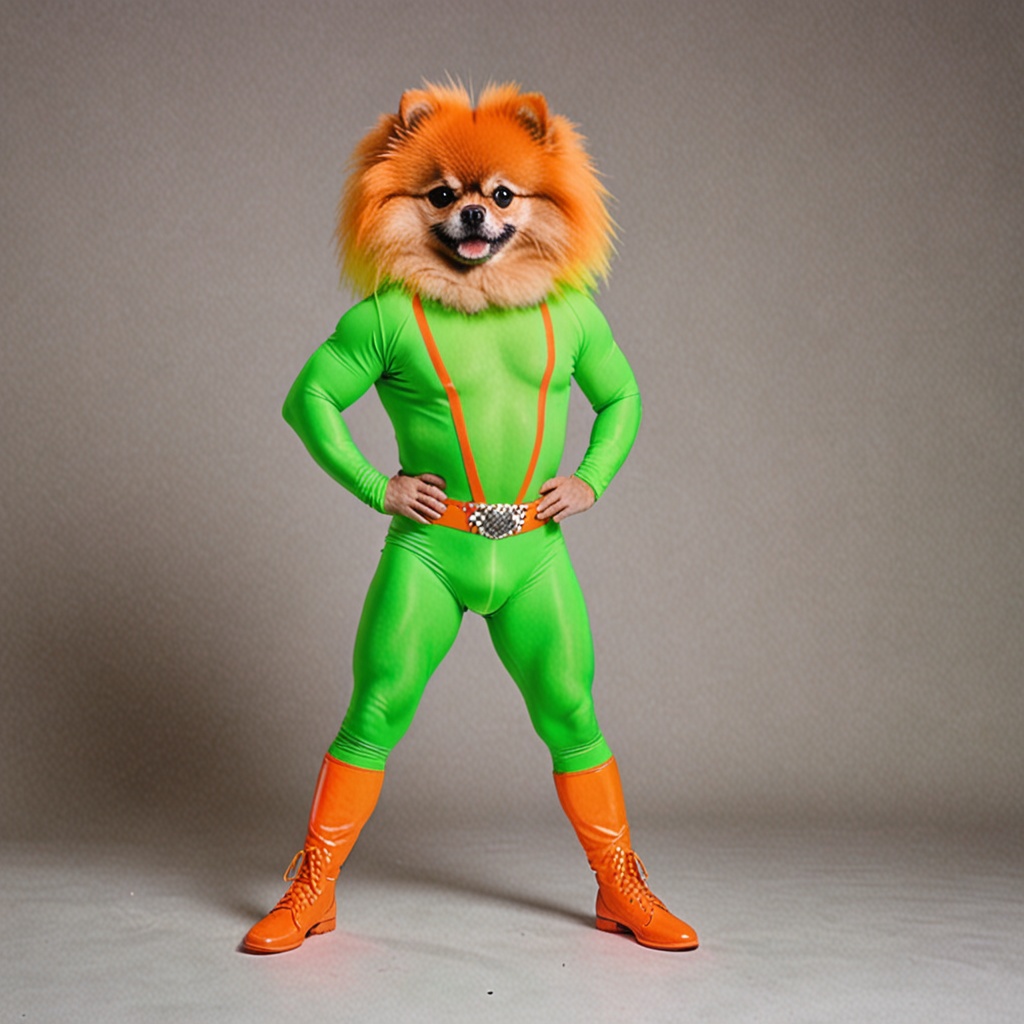} \\
          \multicolumn{4}{@{}p{0.99\textwidth}@{}}{DreamShaper $\times$ DrawBench: ``A realistic photo of a Pomeranian dressed up like a 1980s professional wrestler with neon green and neon orange face paint and bright green wrestling tights with bright orange boots.''} \\[3pt]

          \includegraphics[width=0.30\textwidth]{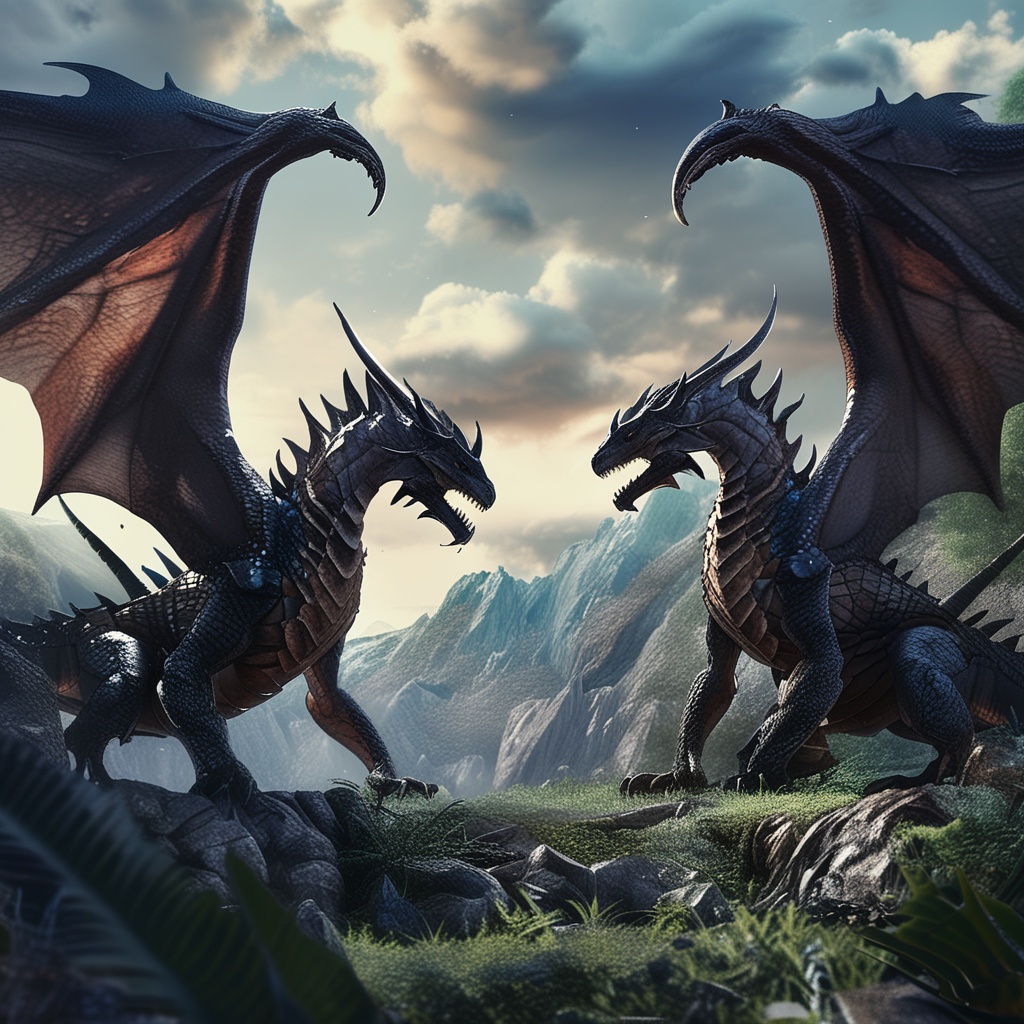} &
          \includegraphics[width=0.30\textwidth]{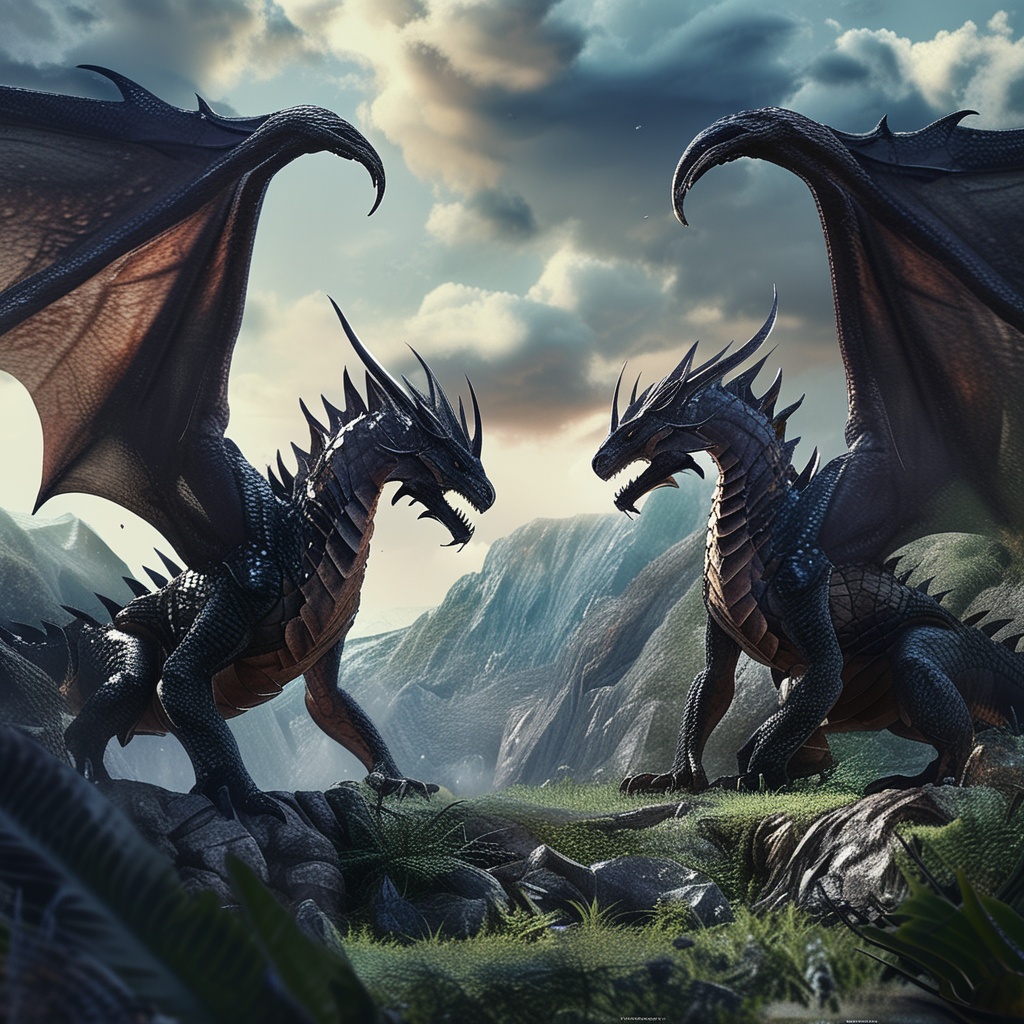} &
          \includegraphics[width=0.30\textwidth]{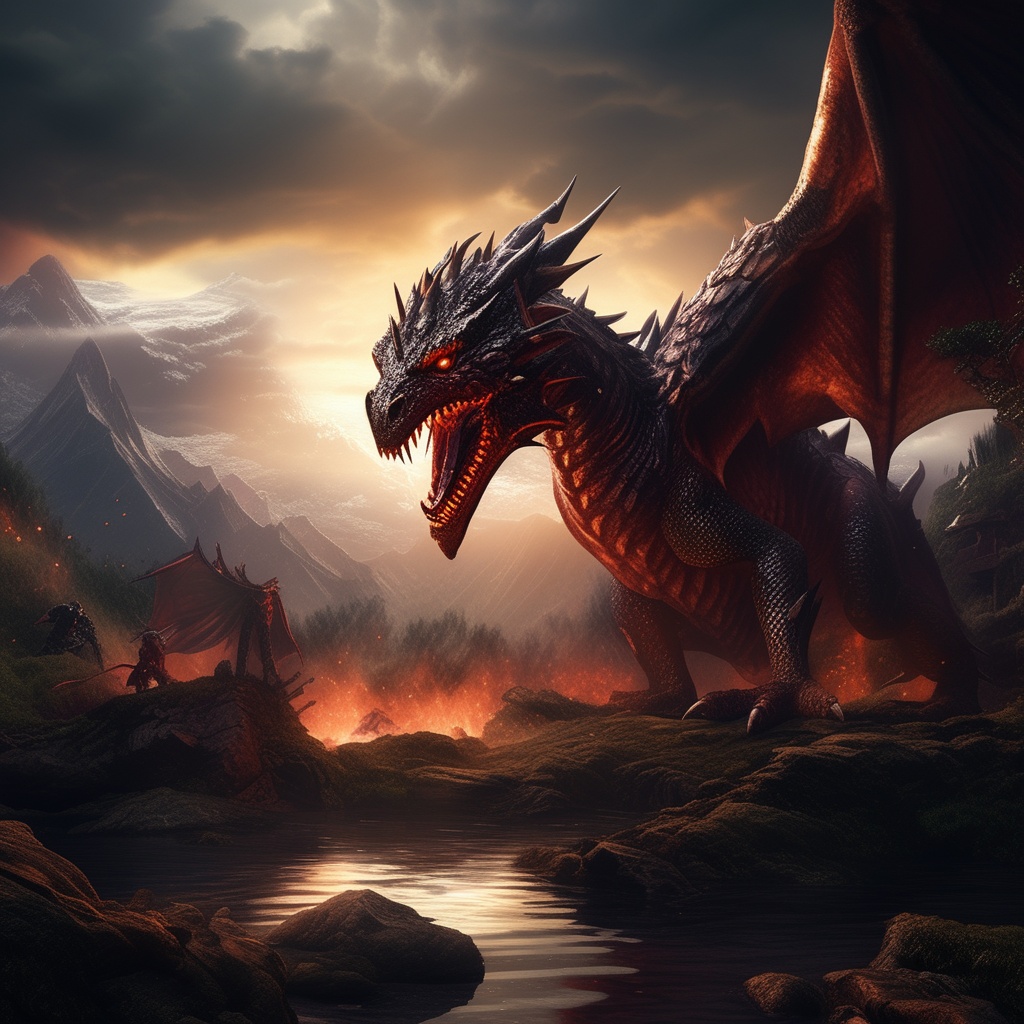} &
          \includegraphics[width=0.30\textwidth]{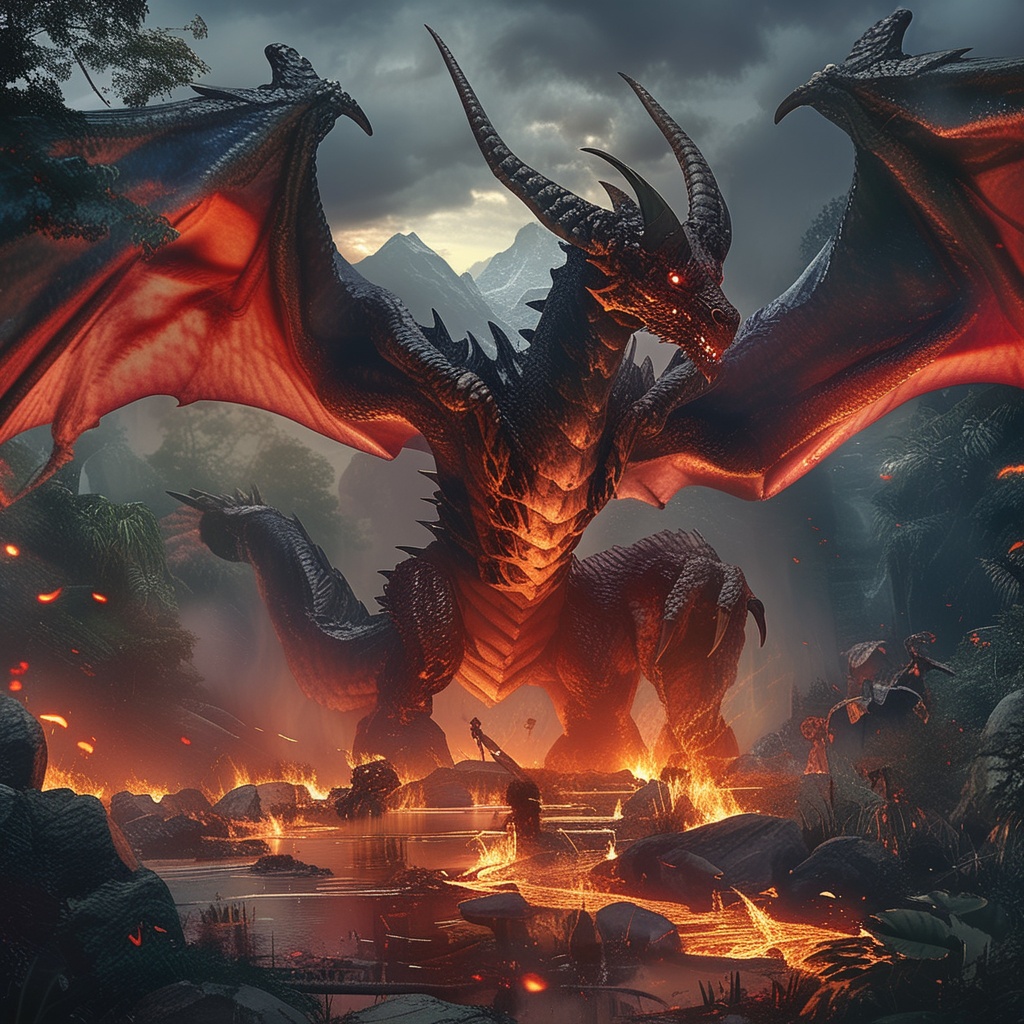} \\
          \multicolumn{4}{@{}p{0.99\textwidth}@{}}{Hunyuan-DiT $\times$ Pick-a-Pic: ``a majestic vicious dragon habitat landscape, boss fight scene, refractions, realistic, cinematic lighting, antview, natural, horrific atmosphere, sharp focus''} \\[3pt]

          \includegraphics[width=0.30\textwidth]{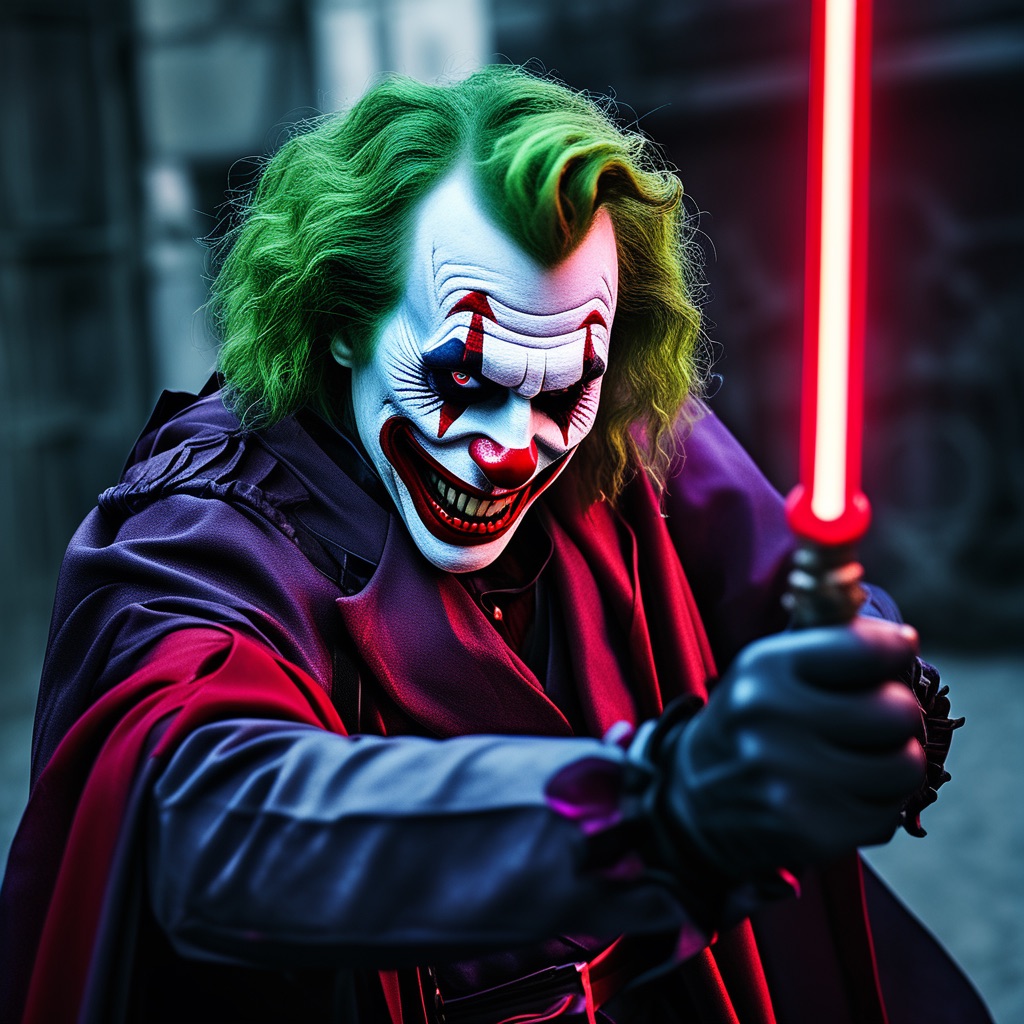} &
          \includegraphics[width=0.30\textwidth]{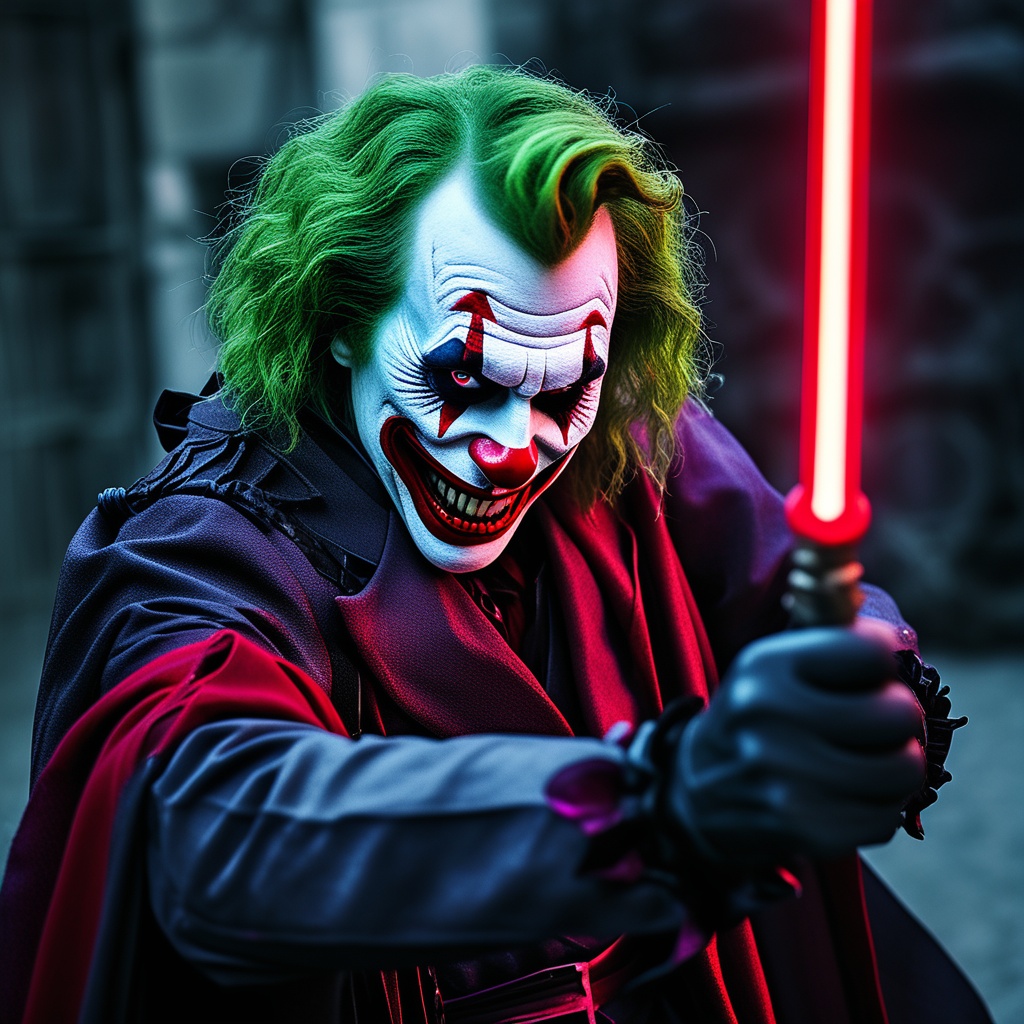} &
          \includegraphics[width=0.30\textwidth]{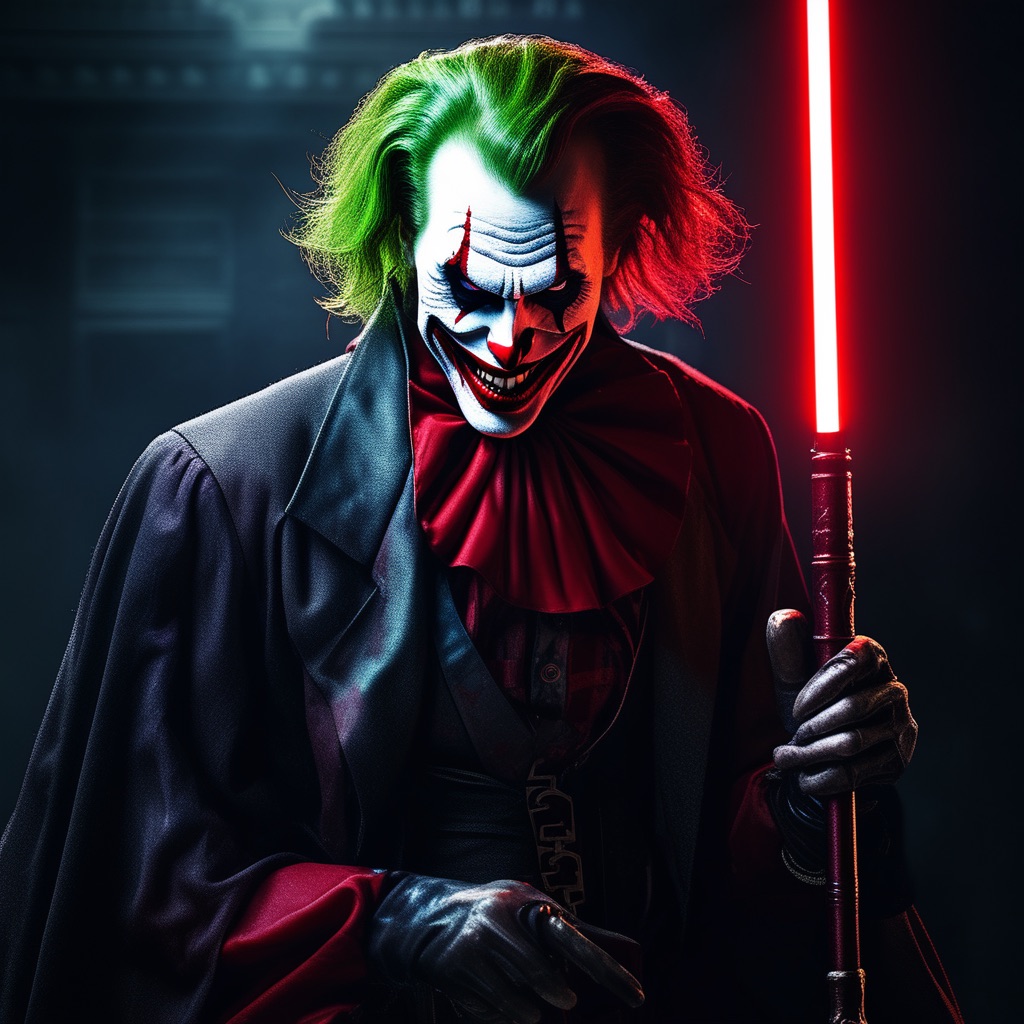} &
          \includegraphics[width=0.30\textwidth]{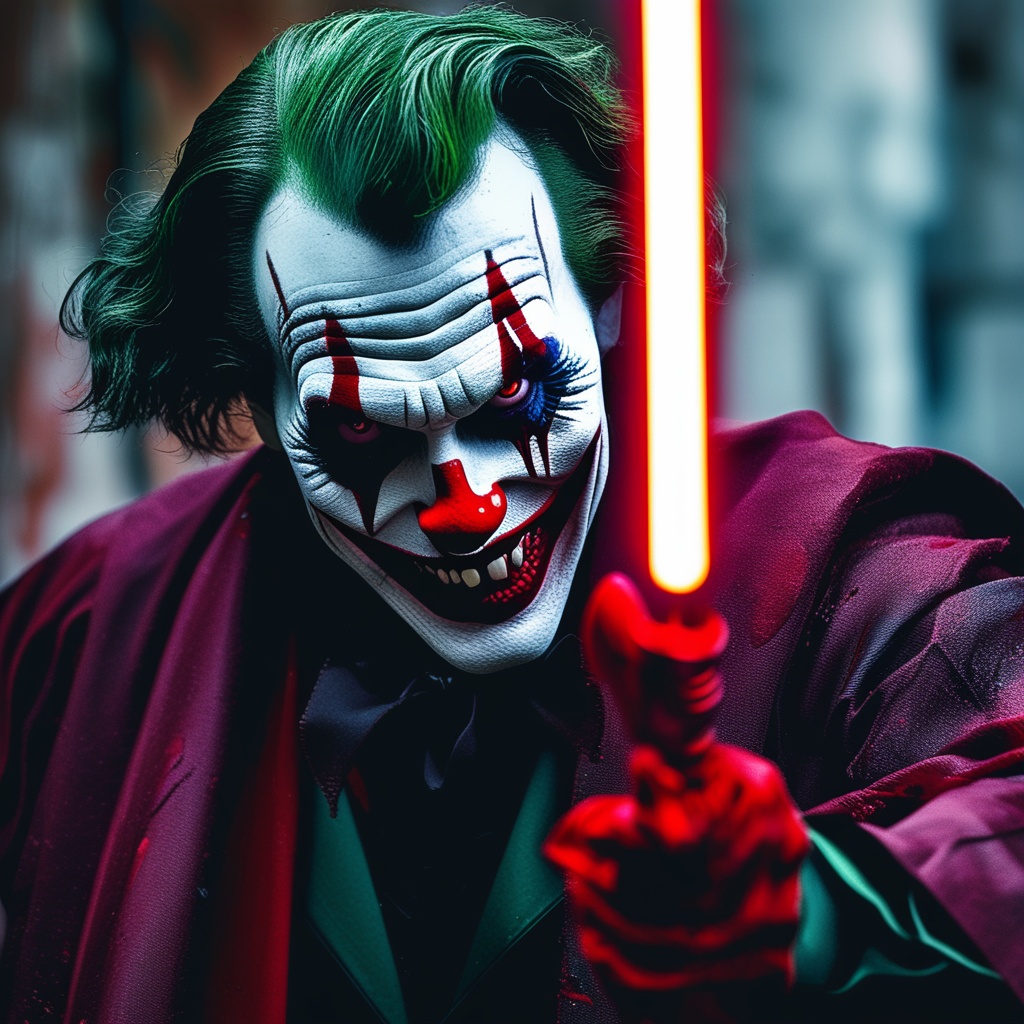} \\
          \multicolumn{4}{@{}p{0.99\textwidth}@{}}{Hunyuan-DiT $\times$ Pick-a-Pic: ``Movie Still of the Joker wielding a red Lightsaber, Darth Joker a sinister evil clown prince of crime, HD photograph''} \\[3pt]

          \includegraphics[width=0.30\textwidth]{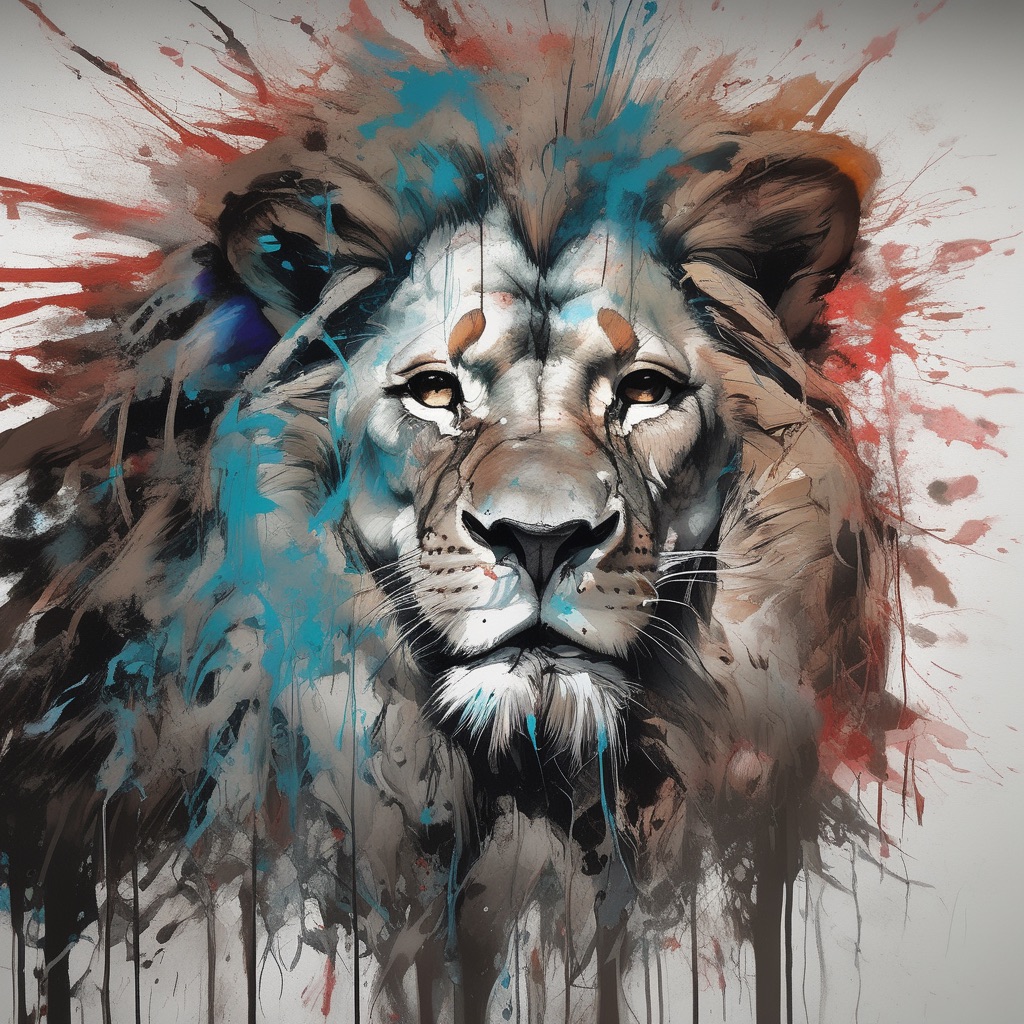} &
          \includegraphics[width=0.30\textwidth]{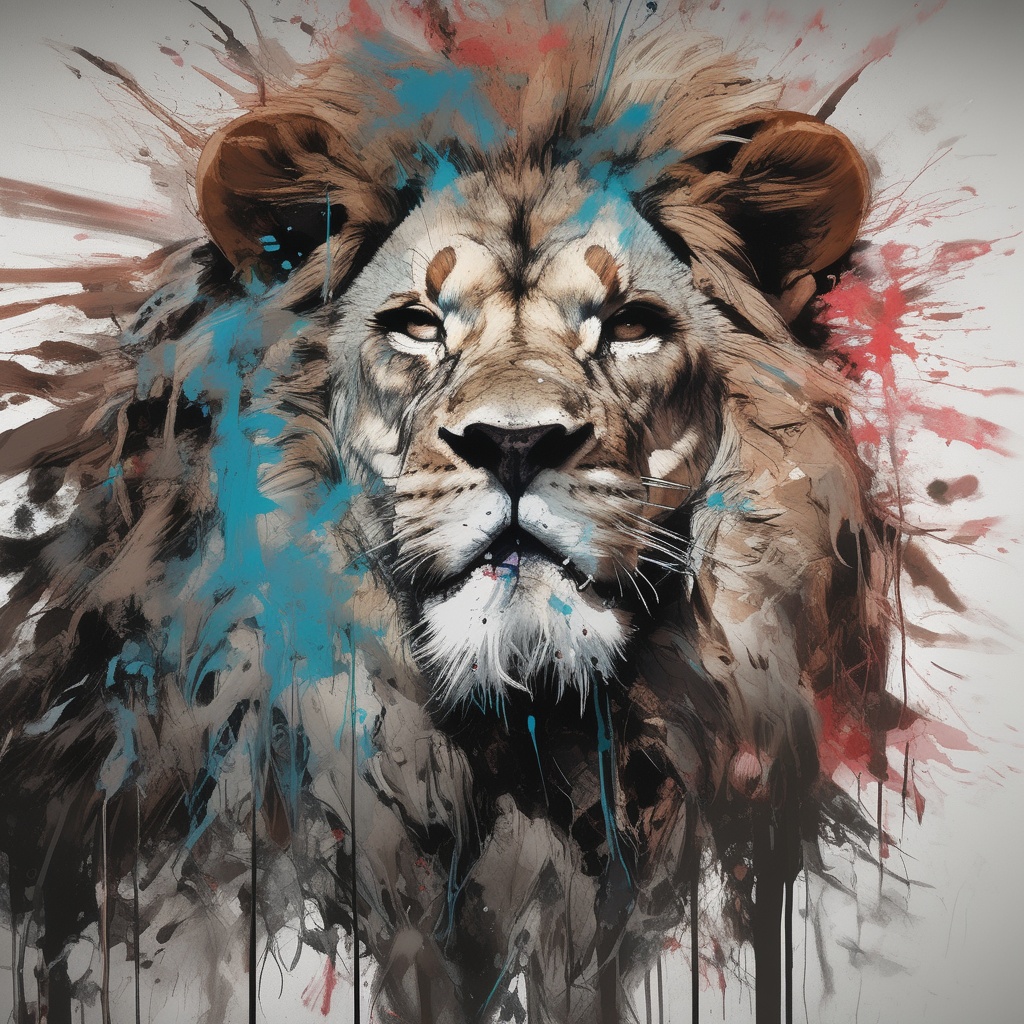} &
          \includegraphics[width=0.30\textwidth]{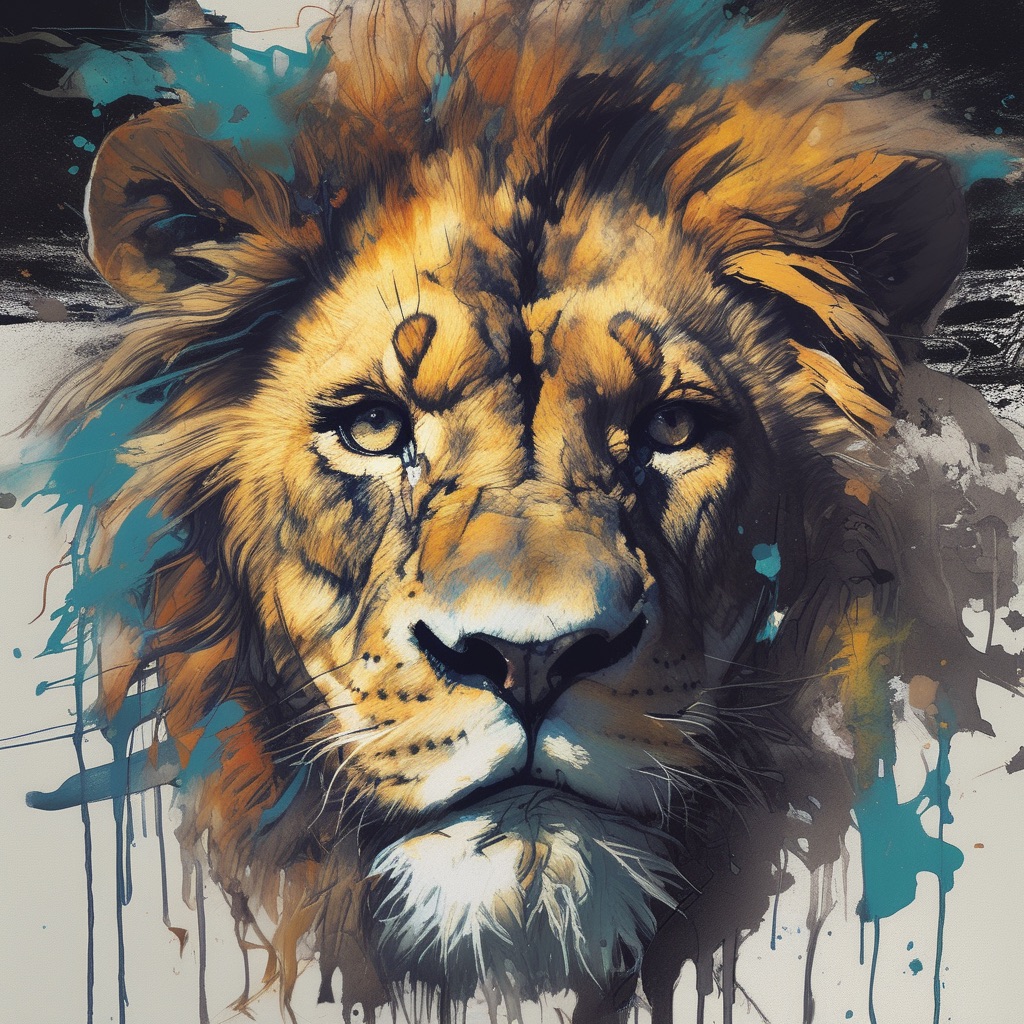} &
          \includegraphics[width=0.30\textwidth]{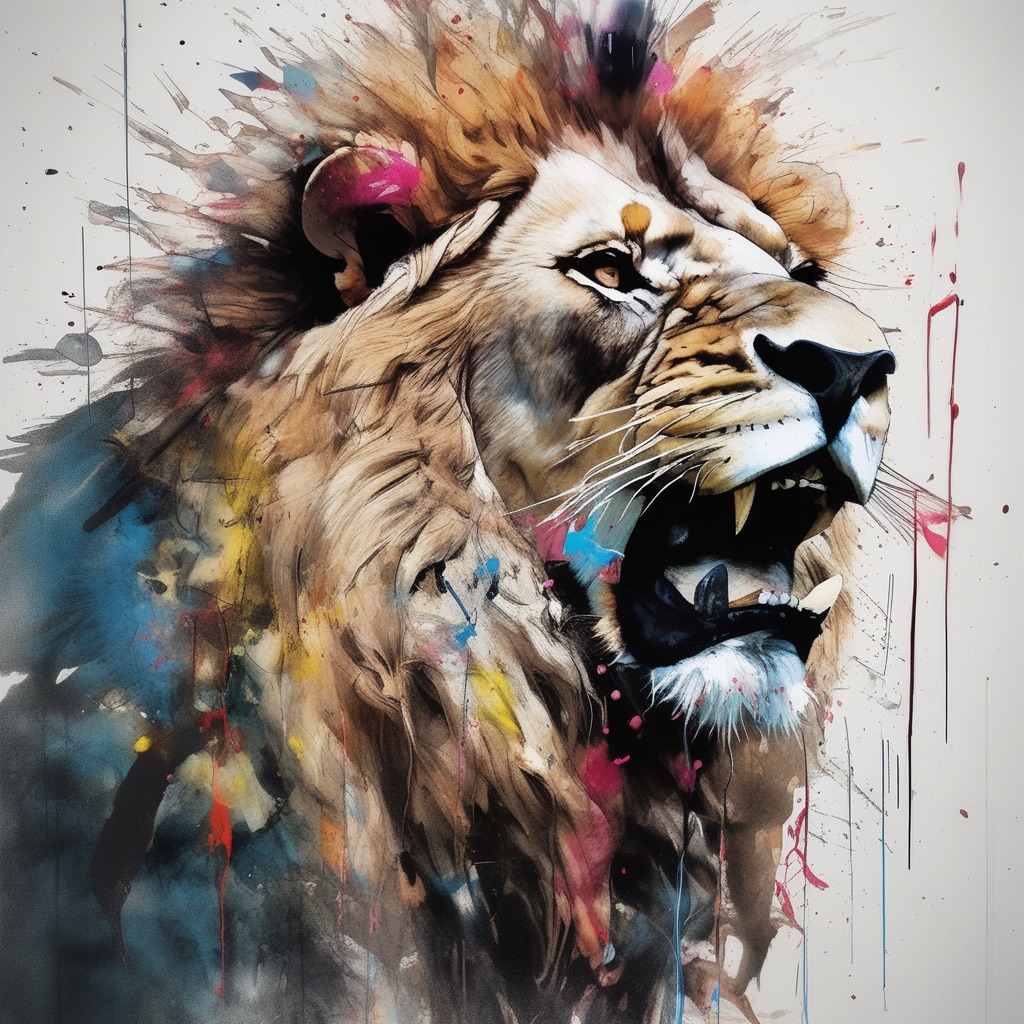} \\
          \multicolumn{4}{@{}p{0.99\textwidth}@{}}{SDXL $\times$ Pick-a-Pic: ``Detailed Ink splash portrait of a Lion: Alex maleev: Ashley Wood: Carne Griffiths: oil painting: high contrast: COLORFUL: 3D: ultra-fine details: dramatic lighting: fantastical: sharp focus: Daniel dociu: splash art: professional photography: Artur N. Kisteb: ZBrushCentral: finalRender: Unreal Engine 5: Trending on Artstation: Jeff Koons: Deep colors: deep depth of field''} \\[3pt]

      \end{tabular}
      }
      \caption{Additional qualitative image examples like Fig.~\ref{fig:qual_examples}.} %
      \vspace{-2mm}
      \label{fig:qual_examples_2}
  \end{figure}

We provide additional in-distribution prompt-image examples in Figs.~\ref{fig:qual_examples} and \ref{fig:qual_examples_2}. %
Similar to Fig.~\ref{fig:mainbody_examples} in the main body of the paper, we observe that \np\ can provide more accurate and prompt-faithful generations than the other baseline methods. Specific examples include the vase image in Fig.~\ref{fig:qual_examples}, where the standard baseline and Golden Noise struggled to adequately draw the vase, while the image generated by NoiseAR has legs which are too short. Similarly, for the second image of an `iOS app for ordering different types of milk', NoiseAR does not draw a phone or app. 

\end{document}